\documentclass[onefignum,onetabnum]{siamart171218}

\usepackage{bm}
\usepackage{float}
\usepackage{subfigure}
\usepackage{amsmath}
\usepackage{mathtools}
\usepackage{verbatim} 
\usepackage[utf8]{inputenc} % allow utf-8 input
\usepackage[T1]{fontenc}    % use 8-bit T1 fonts
\usepackage{hyperref}       % hyperlinks
\usepackage{url}            % simple URL typesetting
\usepackage{booktabs}       % professional-quality tables
\usepackage{amsfonts}       % blackboard math symbols
\usepackage{nicefrac}       % compact symbols for 1/2, etc.
\usepackage{microtype}      % microtypography
\usepackage{xcolor}         % colors
\usepackage{enumitem}
\usepackage{amssymb}
\usepackage{cite}

\usepackage{lipsum}
\usepackage{amsfonts}
\usepackage{graphicx}
\usepackage{epstopdf}
\usepackage{algorithmic}
\ifpdf
  \DeclareGraphicsExtensions{.eps,.pdf,.png,.jpg}
\else
  \DeclareGraphicsExtensions{.eps}
\fi

\newsiamremark{remark}{Remark}
\newsiamremark{hypothesis}{Hypothesis}
\crefname{hypothesis}{Hypothesis}{Hypotheses}
\newsiamthm{claim}{Claim}

\headers{Level Set Learning with Pseudo-Reversible Neural Networks}{Y.~Teng, Z.~Wang, L.~Ju,  A.~Gruber and G.~Zhang}

\title{Level set learning with pseudo-reversible neural networks for nonlinear dimension reduction in function approximation}

\author{Yuankai Teng\footnotemark[2]
 \and Zhu Wang\footnotemark[2]
 \and Lili Ju{\thanks{C\lowercase{orresponding author}}} \thanks{Department of Mathematics, University of South Carolina, Columbia, SC 29208, USA.
  (\email{yteng@email.sc.edu}, \email{wangzhu@math.sc.edu}, \email{ju@math.sc.edu})}
\and Anthony Gruber\thanks{Department of Scientific Computing, Florida State University, Tallahassee, FL 32306, USA.
  (\email{agruber@fsu.edu}).}
   \and Guannan Zhang\thanks{Computer Science and Mathematics Division, Oak Ridge National Laboratory, Oak Ridge, TN 37831, USA. (\email{zhangg@ornl.gov}).} }
\usepackage{amsopn}

\ifpdf
\hypersetup{
  pdftitle={Level set learning with pseudo-reversible neural networks for nonlinear dimension reduction in function approximation},
  pdfauthor={Y. Teng, Z. Wang, L. Ju and A. Gruber}
}
\fi

\begin{document}

\maketitle

% REQUIRED
\begin{abstract}
Due to the curse of dimensionality and the  limitation on training data, approximating high-dimensional functions is a very challenging task even for powerful deep neural networks. Inspired by the Nonlinear Level set Learning (NLL) method that uses the reversible residual network (RevNet),  in this paper  we propose a new method of Dimension Reduction via Learning Level Sets (DRiLLS) for function approximation.
Our method contains two major components: one is the pseudo-reversible neural network (PRNN) module that effectively transforms high-dimensional input variables to low-dimensional active variables, and the other is the synthesized regression module for approximating function values based on the transformed data in the low-dimensional space. The PRNN not only relaxes the invertibility constraint of the nonlinear transformation present in the NLL method due to the use of RevNet, but also adaptively weights the influence of each sample and controls the sensitivity of the function to the learned active variables. The synthesized regression uses Euclidean distance in the input space to select neighboring samples, whose projections on the space of active variables are used to perform local least-squares polynomial fitting. This helps to resolve numerical oscillation issues present in traditional local and global regressions. Extensive experimental results demonstrate that our DRiLLS method outperforms both the  NLL  and  Active Subspace methods, especially when the target function possesses critical points in the interior of its input domain.
\end{abstract}

% REQUIRED

% REQUIRED

\section{Introduction}
High-dimensional function approximation plays an important role in building predictive models for a variety of scientific and engineering problems. 
It is typical for scientists to build an accurate and fast-to-evaluate surrogate model to replace a computationally expensive physical model, in order to reduce the overall cost of a large set of model executions. 
However, when the dimension of the target function's input space becomes large, the data fitting becomes a computationally challenging task. Due to the {curse of dimensionality}, an accurate function approximation would require the number of samples in the training dataset to increase exponentially with respect to the dimension of input variables. On the other hand, given the complexity of the underlying physical model, the amount of observational data is often very limited. This causes classical approximation methods such as sparse polynomial approximation (e.g. sparse grids) to fail on high-dimensional problems outside of some special situations.
One way to alleviate the challenge is to reduce the input dimension of the target function by finding intrinsically low-dimensional structures. 

The existing methods for dimension reduction in function approximation can be divided into two main categories. The first one is to exploit the dependence between input variables to build low-dimensional manifolds in the input space. 
For example, principal component analysis \cite{abdi2010principal} is widely used, due to its simplicity, to compress the input space to a low-dimension manifold. Isometric feature mapping \cite{tenenbaum2000global} is an effective method to compute a globally nonlinear low-dimensional embedding of high-dimensional data. Its modification known as locally linear embedding \cite{roweis2000nonlinear,donoho2003hessian} provides solutions to more general cases.
However, in practice there are often no dependences between input variables to exploit, so that the dimension of the input space cannot be effectively reduced by methods reliant on this assumption.
This represents a challenging research question for function approximation, namely, {\em how to effectively reduce the dimension of a function with independent input variables}. 

To answer this question, the second category of dimension reduction methods  aims at reducing the input dimension by 
exploiting the relationship between the input and the output, i.e., learning the geometry of a function's level sets. This includes methods such as  sufficient dimension reduction (SDR) \cite{cook2005sufficient,adragni2009sufficient,li2018sufficient,pinkus_2015}, 
the active subspace (AS) method \cite{constantine2014active,constantine2015active}, and
neural network based methods \cite{zhang2019learning, tripathy2018deep,karumuri2020simulator,Bigoni2021nonlinear}.
This type of method first identifies a linear/nonlinear transformation that maps the input variables to a handful of active variables (or coordinates), then projects the observational data onto the subspace spanned by the active variables, and finally performs the data fitting in the low-dimensional subspace to determine the  function approximation.

The SDR method \cite{cook2005sufficient,adragni2009sufficient,li2018sufficient} provides a general framework for finding reduced variables in statistical regression. Given the predictor (input) $\bm{x}\in \mathbb{R}^d$ and its associated scalar response (output) $y$, the SDR seeks $\bm{R}: \mathbb{R}^d\rightarrow \mathbb{R}^p$ such that $\mathbb{E}(y|\bm{x}) = \mathbb{E}(y|\bm{R}(\bm{x}))$. Various algorithms have been developed  to determine $\bm{R}$, including the sliced inverse regression \cite{li1991sliced,pan2017sliced,10.2307/2290564}, sliced average variance estimation \cite{dennis2000save}, and principal Hessian directions \cite{li1992principal} in which the population moment matrices of the inverse regression are approximated based on the given regression data. These methods can be extended to the nonlinear setting by introducing the kernel approaches as done in \cite{10.2307/26362824, 10.1214/12-AOS1071, doi:10.1198/106186008X345161,Yeh2009NonlinearDR} . 

The AS method \cite{constantine2014active,constantine2015active} is a popular dimension reduction approach that seeks a set of directions
in the input space, named active components, affecting the function value most significantly on average. Given the values of the function 
 $f(\bm{x})$ and its gradient $\nabla f(\bm{x})$ at  a set of sample points, this method first evaluates the uncentered covariance matrix of the gradient $\bm{C} = \mathbb{E}[\nabla f (\nabla f)^\top]$. The eigenvectors associated with the leading eigenvalues of $\bm{C}$, denoted by $\bm{W}_A$, are selected to define active components $\bm{z}_A =\bm{W}_A^\top \bm{x}$, which is a {linear transformation} of the input $\bm{x}$. The subspace spanned by the set of active components $\bm{z}_A$ describes a low-dimensional linear subspace embedded in the original input space that captures most of the variation in model output. A regression surface $\tilde{f}(\bm{z}_A)$ is then constructed based on the data projected onto the active subspace $\{\bm{z}_A, f(\bm{x})\}$, i.e., $f(\bm{x})\approx \tilde{f}(\bm{W}_A^\top \bm{x})$.

Recently, neural network based approaches \cite{zhang2019learning, tripathy2018deep,karumuri2020simulator,Bigoni2021nonlinear} were developed to extract low-dimensional structures from high-dimensional functions. For instance, 
a feature map was built in \cite{Bigoni2021nonlinear} by aligning its Jacobian with the target function's gradient field, and the function was approximated by solving a gradient-enhanced least-squares problem. 
The Nonlinear Level set Learning (NLL) method \cite{zhang2019learning} finds a bijective {nonlinear transformation} that maps an input point $\bm{x}$ to a new point $\bm{z}$ which is of the same dimension as $\bm{x}$, more specifically, $\bm{z} = \bm{r}(\bm{x})$ with $\bm{r}$ modeled by a reversible residual neural network (RevNet) \cite{gomez2017reversible,DBLP:conf/aaai/ChangMHRBH18}.  In this approach, the transformed variables $\bm{z}$ are expected to be split into two sets: a set of active variables (coordinates)
$\bm{z}_A = \{z_k\}_{k\in A}$  and a set of inactive variables $\bm{z}_I = \{z_k\}_{k\in I}$, so that the function value $f(\bm{r}^{-1}(\bm{z}))$ is insensitive to perturbations in $\bm{z}_I$. That is, if $z_k\in\bm{z}_I$, a small perturbation in $z_k$ within the neighborhood of $\bm{z}$ would lead almost no change in function value. Based on this fact, the NLL method employs a loss function that encourages the gradient vector field $\nabla f(\bm{x})$ to be orthogonal to the derivative of $\bm{r}^{-1}(\bm{z})$ with respect to each inactive variable $z_k\in\bm{z}_I$. Therefore, after a successful training, the NLL method can provide a manifold that captures the low-dimensional structure of the function's level sets.  Similar to the AS method, once $\bm{z}_A$ is determined, a regression surface can be built using the data projected onto the subspace of active variables, $\{\bm{z}_A, f(\bm{x})\}$. It has been shown in \cite{zhang2019learning} that NLL outperforms AS when the level sets of the function have nontrivial curvature.
An improved algorithm for the NLL method was studied in \cite{gruber2021mnll}. However, there still are even simple cases in which the NLL fails to effectively extract low-dimensional manifolds as shown later in this paper.

In this paper, we introduce a new {\em Dimension Reduction via Learning Level Sets} (DRiLLS) method for function approximation that improves upon existing level set learning methods in the following aspects: 
(1) To enhance the model's capability, we propose a novel pseudo-reversible neural network (PRNN) to model the nonlinear transformation for extracting active variables.
(2) The learning process is driven by geometric features of the unknown function, which is reflected in a loss function consisting of three terms: the pseudo-reversibility loss, the active direction fitting loss, and the bounded derivative loss.
(3)  A novel synthesized regression on the manifold spanned by the learned active variables is also proposed, which helps to resolve numerical oscillation issues and provides accuracy benefits over traditional local and global regressions.
Extensive numerical experiments demonstrate that the proposed DRiLLS method leads to significant improvements on high-dimensional function approximations with limited or sparse data.   

The rest of paper is organized as follows. In \Cref{sec:ll}
the setting of the function approximation problem is introduced and the DRiLLS method is proposed and discussed. More specifically,  the PRNN module is described in \Cref{sec:PRNN} and the  synthesized regression module in \Cref{sec:newreg}. We then numerically investigate the performance of our DRiLLS method in \Cref{sec:tests}, including ablation studies in \Cref{sec:abl}, high-dimensional function approximations with limited/sparse data in \Cref{sec:mor} and a PDE-related application in \Cref{pdeapp}. Finally, some concluding remarks are drawn in \Cref{sec:conclusion}.

\section{The proposed DRiLLS method}\label{sec:ll} 

We consider a scalar {target function}, which is continuously differentiable on a bounded Lipschitz domain $\Omega$ in $\mathbb{R}^d$:
\begin{equation}\label{problem_1}
y=f(\bm{x}), \quad \bm{x}=(x_1, x_2, \ldots, x_d) \in \Omega.
\end{equation}
The input variables $x_1, x_2, \ldots, x_d$ are assumed to be independent from each other, which implies that the input space itself does not possess a low-dimensional structure. 
The goal is to find an approximation $\hat{f}(\bm{x})$ of the target function, given the information of $f$ and  $\nabla f$ on a set of training samples in $\Omega$. 
We denote the training dataset by 
$$\Xi \coloneqq \left\{\left(\bm{x}^{(n)}, f(\bm{x}^{(n)}), \nabla f(\bm{x}^{(n)})\right) : n=1, \dots, N \right\},$$
which contains  the input, the output and the gradient information at the samples.
When the number of dimensions $d$ is large, taking a handful of random selections in each coordinate would result in a huge amount of data, which is infeasible in many application scenarios. Therefore, the sample dataset is usually sparse for high-dimensional problems.   

The NLL method has achieved successes in high-dimensional function approximation on sparse data for real-world applications such as composite material design problems \cite{zhang2019learning}, however, it has difficulties in learning level sets of certain functions. In particular, NLL struggles on functions with critical points contained  in the interior of  the domain  $\Omega$, such as the functions $x_1^2+x_2^2$ or $x_1^2- x_2^2$ on $\Omega = [-1, 1]^2$ to be discussed later in \Cref{sec:abl}.
One reason for such drawback is that the RevNet employed by NLL  enforces  invertibility as a hard constraint, which limits the capability of the RevNet in learning the structure of functions whose level sets are not homeomorphic to hyperplanes in the input space. Another reason is that the rate of change in the target function  with respect to the inactivate variables is always zero at any interior critical points, as the gradient of the function vanishes there. Hence, the training process tends to ignore samples lying in a small neighborhood of the critical points since they do not contribute much to the training loss. 

\begin{figure}[!htbp]
  \centerline{
  \includegraphics[width=5.1in]{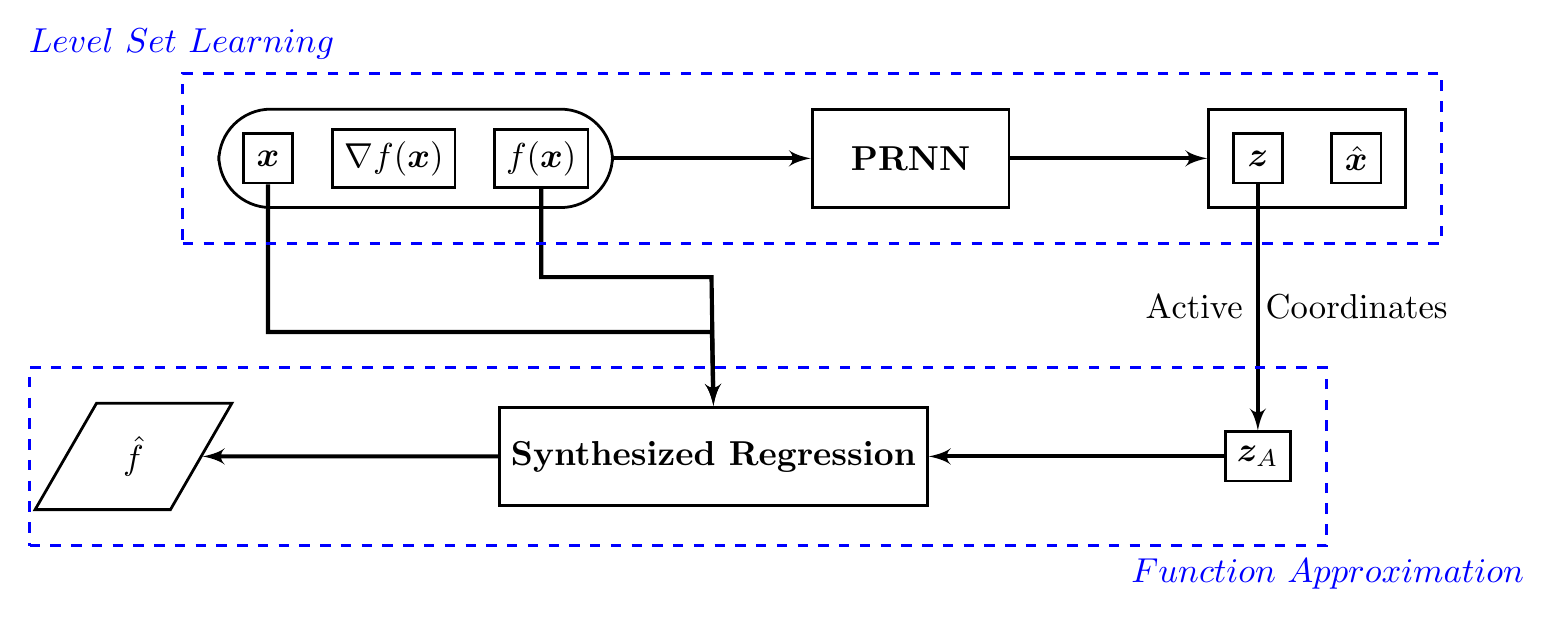}}
     \vspace{-0.2cm}
    \caption{The overall structure of the proposed DRiLLS method, which consists of two major components: the PRNN module and the synthesized regression module.}
  \label{fig:whole_struc}
\end{figure}

To overcome these issues and improve the performance of level set learning based function approximation,  the proposed DRiLLS method consists of two major components: (1) the PRNN module that identifies active variables and reduces the dimension of input space, and (2) the synthesized regression module 
that selects neighboring sample points according to their Euclidean distances in the original input space and performs a local least-squares fitting based on the learned active variables to approximate the target function.
A schematic diagram of the proposed method is shown in \Cref{fig:whole_struc}. 

\subsection{The pseudo-reversible neural network}\label{sec:PRNN}
\begin{figure}[!htbp]
  \centerline{
  \includegraphics[width=1.05\textwidth]{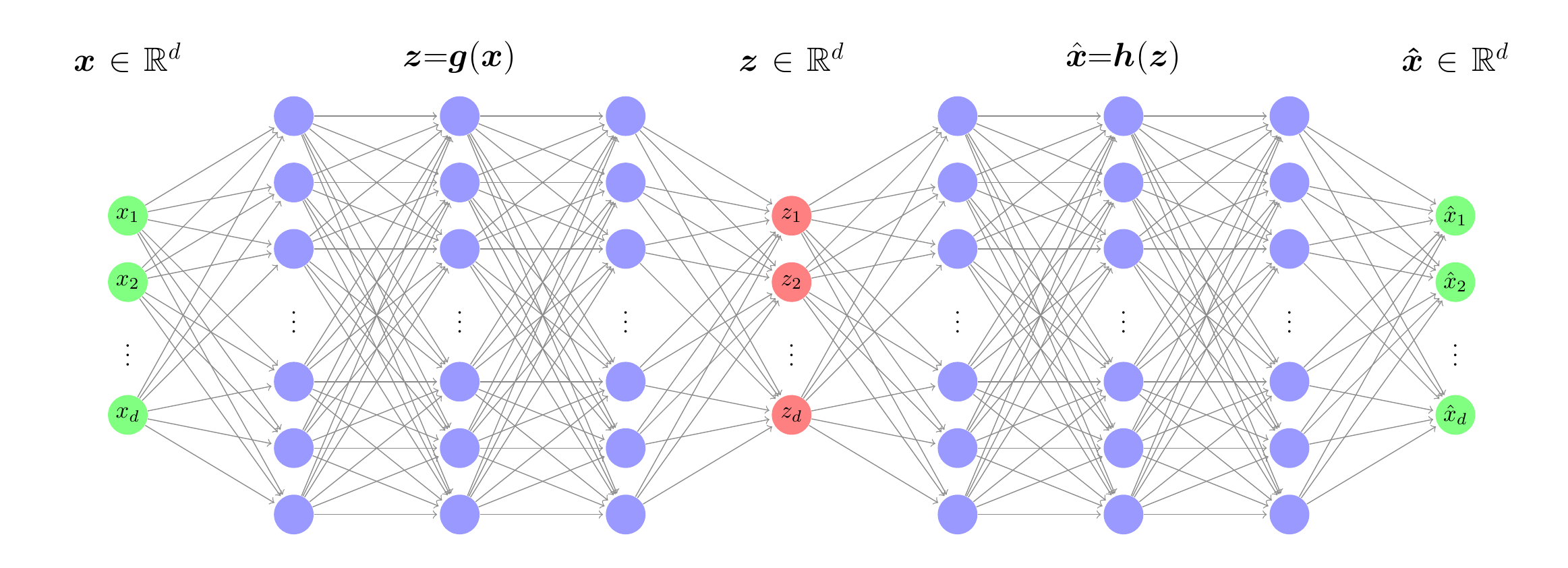}}
  \vspace{-0.2cm}
  \caption{The pseudo-reversible neural network (PRNN) consists of two FCNNs representing $\bm{g}$ and $\bm{h}$ respectively, that  possess the same number of hidden layers (3 layers for illustration) and the same number of neurons at each layer.}
  \label{fig:prnn_struc}
\end{figure}

To construct the PRNN, we first define a nonlinear mapping from the input $\bm{x}$ to a new point $\bm{z}$ of the same dimension. 
In contrast to the RevNet used by the NLL method, the invertibility of this transformation is relaxed by defining another mapping from $\bm{z}$ to $\hat{\bm x}$ and encouraging $\hat{\bm x}$ to be close to $\bm{x}$ in distance. Thus, the reversibility is imposed as a soft constraint on the PRNN model. Specifically,
the two {nonlinear transformations} are denoted by 
\begin{equation}\label{e14}
\bm z = \bm g(\bm x; \Theta_g) \quad \text{and}\quad \hat{\bm x} = \bm h(\bm z; \Theta_h), 
\end{equation}
respectively, where $\bm{g}$, $\bm{h}: \mathbb{R}^d \rightarrow \mathbb{R}^d$ with $\Theta_g$ and $\Theta_h$ being  their learnable parameters. 
Since $\bm{g}$ is not exactly invertible by definition, $\bm h$ can be viewed as a pseudo-inverse function to $\bm {g}$. 
Both $\bm{g}$ and $\bm{h}$ are represented by a fully connected neural network (FCNN), as displayed in \Cref{fig:prnn_struc}. 
The PRNN network structure is reminiscent of an autoencoder \cite{goodfellow2016deeplearning}, but the dimension of latent space (i.e., the dimension of $\bm{z}$) remains the same as the dimension of $\bm x$.  While there are no theoretical restrictions on the structure of $\bm{g}$ and $\bm{h}$, the experiments in \Cref{sec:tests} use the same FCNN architecture for both mappings.

\subsubsection{The loss function} 
The learnable parameters $\Theta_g$ and $\Theta_h$ are updated synchronously during the training process by minimizing the following total loss function:
\begin{equation}\label{totalloss}
\mathcal{L}=\mathcal{L}_{1}+\lambda_1 \mathcal{L}_{2}+\lambda_2 \mathcal{L}_{3}.
\end{equation}
Here, $\mathcal{L}_{1}$ is the {\em pseudo-reversibility loss} which measures the difference between $\bm{x}$ and the PRNN output $\hat{\bm{x}}$, $\mathcal{L}_{2}$ is the  {\em active direction fitting loss} which which enforces the tangency between $\frac{\partial\bm{x}}{\partial\bm{z}_I}$ and the level sets of $f$, and $\mathcal{L}_{3}$ is the {\em bounded derivative loss} which  regularizes the sensitivity of $f$ with respect to the active variables $\bm z_A$. The weights $\lambda_1$ and $\lambda_2$ are hyper-parameters for balancing the three loss terms. Below each term of $\mathcal{L}$ is discussed in detail.

\paragraph{The pseudo-reversibility loss} 
In order to train $\hat{\bm{x}} = \bm{h}(\bm{z})$ to be a pseudo inverse of $\bm{z} = \bm{g}(\bm{x})$,
the pseudo-reversibility condition is simply enforced in the $L^2$ sense: 
\begin{equation}\label{eq:L1}
\mathcal{L}_{1} = \frac{1}{N} \sum_{n=1}^N ||\bm{x}^{(n)}-\bm{h} \circ \bm{g}(\bm{x}^{(n)})||_2^2,
\end{equation}
which is the same as the standard loss used to train autoencoders.

\paragraph{The active direction fitting loss} This loss is defined based on the fact that if the $k$-th output $z_k$ of $\bm g(\bm x)$ is inactive, a small perturbation of $z_k$ in a neighborhood of $\bm{z}$ would change  the target function $f$ along a direction tangent to its level sets.
Specifically, we define the Jacobian matrix of the nonlinear transformation $\bm{h}$ as:
$$\bm{J}_{\bm{h}} (\bm{z})= [\bm{J}_1 (\bm{z}), \bm{J}_2 (\bm{z}), \dots, \bm{J}_d (\bm{z})]$$
with $$\bm{J}_i (\bm{z}) \coloneqq \left[\frac{\partial \hat{x}_1}{\partial z_i}(\bm{z}), \dots, \frac{\partial \hat{x}_d}{\partial z_i}(\bm{z})\right]^{\top}.$$
In the ideal case, if $z_k$ is completely inactive, then
the gradient vector field $\nabla f(\bm{x})$ is orthogonal to $\bm{J}_k (\bm{z})$, that is
$\langle \bm{J}_k (\bm{z}), \nabla f(\bm{x}) \rangle=0$ with $\langle \cdot , \cdot \rangle$ denoting the inner product. 
Thus the {\em active direction fitting loss} is defined to encourage the orthogonality, i.e., 
\begin{equation}\label{eq:L2}
\mathcal{L}_{2}=\frac{1}{N}  \sum_{n=1}^N \gamma_n\sum_{i=1}^d \left[\omega_i \left\langle {\bm J_i(\bm z^{(n)})},  \nabla f(\bm x^{(n)}) \right\rangle \right]^2,
\end{equation}
where the scaling factors
$$\gamma_n=1+\alpha e^{-\|\nabla f(\bm{x}^{(n)})\|},\quad n=1,2\cdots,N$$
contain the hyper-parameter $\alpha\geq 0$ and $\omega_1, \omega_2, \ldots, \omega_d \in \{0,1\}$ are weight hyper-parameters determining 
how strictly the orthogonality condition is enforced for each of the $d$ variables. A typical choice is 
\begin{equation}\label{actfor}
\bm{\omega}=(\overbrace{0,\dots,0}^{k^*},\overbrace{1, \dots, 1}^{d-k^*}),
\end{equation}
 where $k^*$ 
denotes  the dimension of the active variables/coordinates.
An ideal case would be $k^*=1$, which implies that there exists only one active variable $\bm{z}_A = \{z_1\}$ and the intrinsic dimension of $f \circ \bm{h} (\bm{z})$ is exactly one when $\mathcal{L}_2=0$. 
The scaling factor $\gamma_i$ distinguishes $\mathcal{L}_2$ from the one used in \cite{zhang2019learning}, and
its value changes according to the magnitude of the gradient: it approaches $1+\alpha$ if $\|\nabla f(\bm{x}^{(n)})\|$ gets close to $0$ and stays close to $1$ otherwise. 
Therefore, it serves as a rescaling factor designed to overcome the situation where the contributions of samples near interior critical points are ignored by the optimization due to their small gradients.

\paragraph{The bounded derivative loss}
Existing methods such as NLL do not place any restrictions on the active variables, because the used RevNet imposes sufficient regularization on those variables. On the other hand, using PRNNs without regularization in $\bm{z}_A$ may cause the network to learn an active subspace which changes too fast, producing undesirable oscillations in the target function.
To address this issue, we introduce a regularization term into the loss as
\begin{equation}\label{eq:L3}
\mathcal{L}_{3}=\frac{1}{N}  \sum_{n=1}^N  \text{sigmoid} \left(\frac{1}{\sigma}{\Big(\Big\|\frac{\partial f\circ \bm{h}}{\partial \bm{z}_A}(\bm{z}^{(n)})\Big\|-1}\Big)\right),
\end{equation}
where $\sigma$ is a positive rescaling hyper-parameter. The purpose is to regularize the magnitude of $\frac{\partial f\circ \bm{h}}{\partial \bm{z}_A}(\bm{z}^{(n)})$ to be not much greater than one. In the practical implementation, we further approximate 
$\frac{\partial f\circ \bm{h}(\bm{z}^{(n)})}{\partial \bm{z}_A} $  with  
$(\nabla f(\bm{x}^{(n)}))^\intercal  \frac{\partial \bm{h}}{\partial \bm{z}_A}(\bm{z}^{(n)})$ by considering the pseudo-reversibility of the PRNN.

\subsection{The synthesized regression}\label{sec:newreg}

\begin{figure}[htbp]
\centerline{
\includegraphics[width=5in]{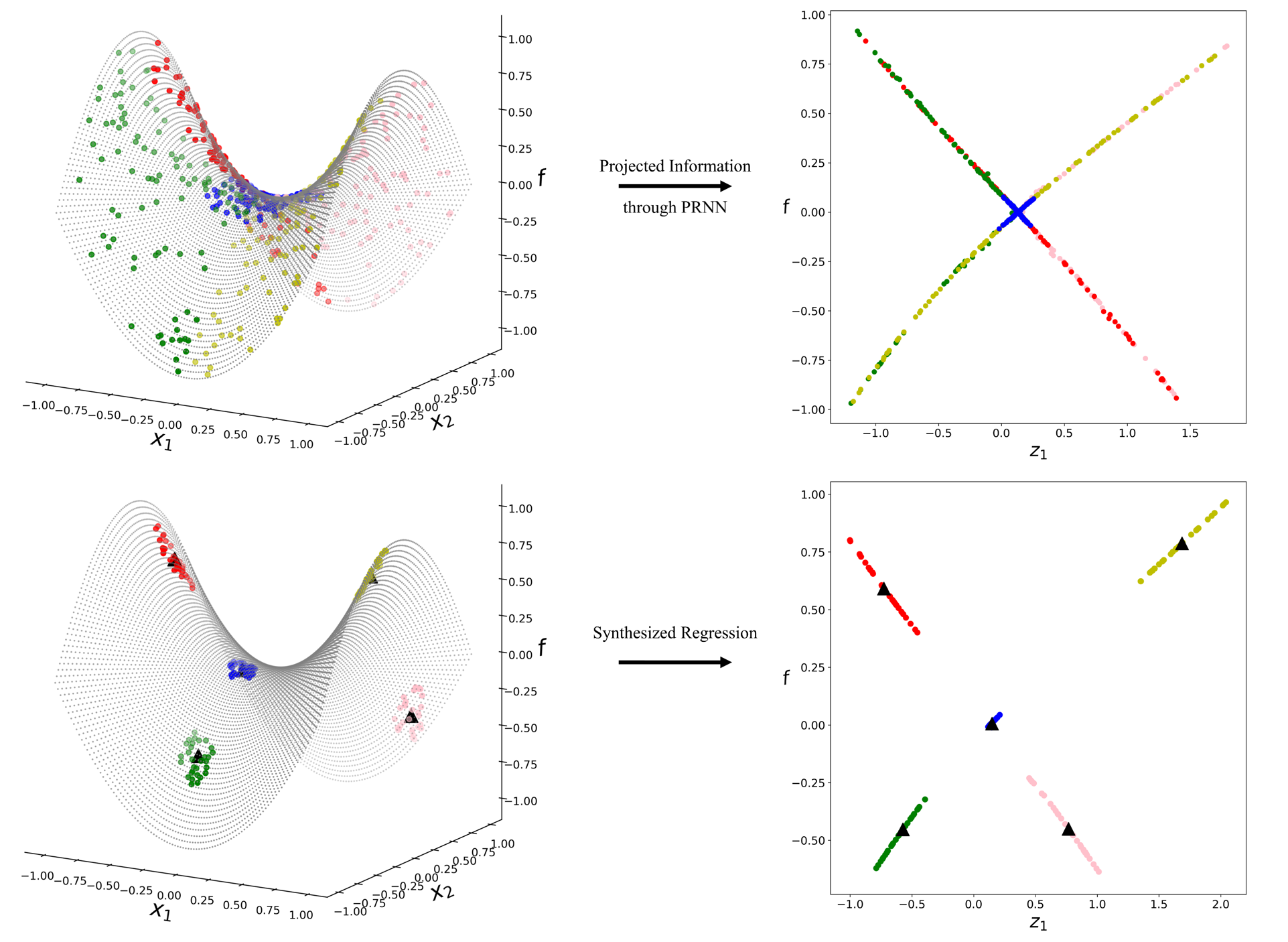}}
  \caption{Top row: An example illustration of  the sample points and their active variables learned by the PRNN, 
where we take   $f(x_1, x_2) = x_1^2 - x_2^2$ and  set 
$z_1$ as the active variable and $z_2$ the inactive variable.   
As $f$ is very oscillatory with respect to $z_1$ and even may not be a function of $z_1$, this case is  not suitable with general local or global regression approaches
based solely on the projected information in the space of active variables, $\{ (z^{(n)}_1,f(\bm{x}^{(n)})\}_{n=1}^{N}$. Bottom row: The proposed synthesized regression first selects the local neighbor sample points for each of the five new inputs (i.e., the five $\blacktriangle$-shaped points whose function values are to be predicted) from the original input space, then  performs respective least-squares polynomial data fitting.}
  \label{fig:reg}
\end{figure}

 The active variables (coordinates) $\bm{z}_A$ is naturally identified based on the pre-setting values of the weights $\bm{\omega}$. 
 Once the PRNN training is completed, the sample points $\{\bm x^{(n)}, n=1, \ldots, N\}$ then can be nonlinearly projected through PRNN to a much lower dimensional space spanned by $\bm{z}_A$. 
Ideally, approximating the high-dimensional function $f(\bm x)$ often can be achieved by approximating the low-dimensional function 
\begin{equation}
    {\widetilde f}(\bm z_A) := f\circ \bm g^{-1}(\bm z) \;\text{ with } \;\bm z_A \in \mathbb{R}^{k*},
\end{equation}
where $k^* \ll d$. 
Many existing methods could be used, including classic polynomial interpolations, least-squares polynomial fitting \cite{GNZ2020SIAMRev}, and regression by deep neural networks. 
However, because the control on $\bm{g}$ is quite loose through the PRNN,  $\widetilde f$ could be very oscillatory 
with respect to $\bm z_A$ or even make $\widetilde f$ fail to form a function. For example, there could exist  two sample points $\bm{x}$ and $\bm{y}$, which are separated in the input space with different values $f(\bm{x})$ and $f(\bm{y})$ but mapped close together in the transformed space, i.e., $(\bm{g}(\bm{x}))_A \approx (\bm{g}(\bm{y}))_A$. This is often the case for functions with interior critical points. The top row of Figure   \ref{fig:reg}  presents an example illustration of  such case, where we take   $f(x_1, x_2) = x_1^2 - x_2^2$ and  set 
$z_1$ as the active variable and $z_2$ the inactive variable.  Consequently, general global or local regression approaches based solely on the projected information in the space of active variables are not able to effectively  handle this case due to large numerical oscillations. 

We develop a synthesized regression method to address this type of numerical oscillation problem.  The method
uses  local least-squares polynomial fitting in the space of active variables, but  selects neighboring sample points based on the Euclidean distance in the original input space to help to keep track of original neighborhood relationships. Our synthesized regression algorithm can be described as follows:
\vspace{0.2cm}
\begin{enumerate}[leftmargin=20pt]\itemsep0.2cm
\item Given an unseen input sample $\bm{x}^*$, 
we select a set of $N_{f}$  points closest to $\bm{x}^*$ from the set of all 
training samples,  denoted by $\{\bm{x}_{*}^{(m)}\}_{m=1}^{N_{f}}$. 
\item The $N_f$ samples are fed into the trained PRNN to generate the samples of the active variables $\{\bm{z}_{*,A}^{(m)}=(\bm{g}(\bm{x}_*^{(m)}))_A\}_{m=1}^{N_{f}}$. 
\item We perform least-squares polynomial fitting using the data $\{(\bm{z}_{*,A}^{(m)}, f(\bm{x}_{*}^{(m)})\}_{m=1}^{N_{f}}$ that is a subset of the training set. The approximation of $f(\bm x^*)$, denoted by $\hat{f}(\bm{x}^*)$, is defined by the value of the resulting polynomial at $(\bm{g}(\bm{x}^*))_A$.
\end{enumerate}
\vspace{0.2cm}

Note that when the graph of $\widetilde f$ in $\bm{z}_A$ has several branches, the first two steps in the proposed synthesized regression encourages localization of the polynomial data fitting to only one of the branches. Indeed, the selected 
neighbors $\{\bm{x}_{*}^{(m)}\}_{m=1}^{N_{f}}$ to $\bm{x}^*$ usually stay on the same branch or intersecting region without much oscillations as shown in the bottom row of Figure   \ref{fig:reg}.   

\section{Experimental results}\label{sec:tests}

The goal of this section is two-fold: the first is to test the influence of each ingredient of the proposed DRiLLS method on its overall performance, and the second is to investigate the numerical performance of the method in approximating high-dimensional functions. In particular, an ablation study is implemented in \Cref{sec:abl}, including PRNN vs. RevNet and the effect of $\alpha$ in \Cref{sec:rev}, the effect of bounded derivative loss 
in \Cref{sec:bdl}, and the synthesized regression vs. some existing regression methods in \Cref{sec:loc}.  
Then, through extensive comparisons with the AS and the NLL methods, we  demonstrate the effectiveness and accuracy of the proposed DRiLLS method under  limited/sparse data. Particularly, high-dimensional example functions are considered in \Cref{sec:mor} and a PDE-related application is given in \Cref{pdeapp}.

The training dataset of size $N$ is randomly generated using the Latin hypercube sampling (LHS) method \cite{tang1993lhs}. 
To measure the approximation accuracy, we use the 
normalized root-mean-square error ($\mathtt{NRMSE}$) and the relative $l_1$ error ($\mathtt{RL_1}$) over a test set of $M$ randomly selected input points from the domain:
\begin{equation}\label{errdef}
\mathtt{NRMSE}=\frac{1}{\sqrt{M}} \left(\frac{\|\bm{f}-\bm{\hat{f}}\|_2}{\max \bm{f} - \min \bm{f}}\right), 
\quad 
\mathtt{RL_1} = \frac{\|\bm{f}-\bm{\hat{f}}\|_1}{\|\bm f\|_1},
\end{equation}
where $\bm{f} = (f(\bm{x}_{test}^{(1)}), \ldots, f(\bm{x}_{test}^{(M)}))$ are the exact function values and
and $\bm{\hat{f}} = (\hat{f}(\bm{x}_{test}^{(1)}), \ldots, \hat{f}(\bm{x}_{test}^{(M)}))$ are the approximated values. 
In the experiments, we set $M=1000$ for low-dimensional problems ($d\leq 3$) and $M=10000$ for high-dimensional problems ($d> 3$).
This procedure is replicated for 10 times and the average values are reported as the final $\mathtt{NRMSE}$ and $\mathtt{RL_1}$ errors for function approximation. 

% common network structure
Our DRiLLS method is implemented using PyTorch. 
% common hyperparameters
If not specified otherwise, we choose the following {\em default model setting}: 
$\bm{g}$ and $\bm{h}$ in the PRNN are constructed by FCNNs that contain 4 hidden layers with $10d$ hidden neurons per layer, respectively; $\mathsf{Tanh}$ is used as the activation function; the
hyper-parameters $\lambda_1=1$, $\lambda_2=1$, $\sigma= 0.01$ and $\alpha = 50$ are selected in the loss function;  
% common regression setting
$N_f= 30$ and cubic polynomial are used for the local least-squares fitting in the synthesized regression.
% common optimization setting
For the  {\em training} of PRNN, we use a combination of the Adam optimizer \cite{kingma2015adam} and the L-BFGS optimizer \cite{nocedal1980updating}. The Adam iteration \cite{kingma2015adam} is first applied with the initial learning rate 0.001, and the learning rate decays every $5000$ steps by a factor of $0.7$ for up to $60000$ steps. Then, the L-BFGS iteration is applied for a maximum of $200$ steps to accelerate the convergence. The training process is 
immediately stopped when training error reduces to $5\times 10^{-5}$. 
Both the AS and the NLL methods used for comparison  are implemented in \text{ATHENA}\footnote{ATHENA codes available at https://github.com/mathLab/ATHENA.} \cite{romor2020athena}, which is a Python package for parameter space dimension reduction in the context of numerical analysis. 
All the experiments reported in this work are  performed on an Ubuntu 20.04.2 LTS desktop with a 3.6GHz AMD Ryzen 7 3700X CPU, 32GB DDR4 memory and NVIDIA RTX 2080Ti GPU.

\subsection{Ablation studies}\label{sec:abl}
We first numerically investigate the effect of major components in the proposed DRiLLS method, including the PRNN, the loss functions, the hyper-parameters and the synthesized regression. 
Several functions of two dimensions are considered.  
Since the dimension $d$ is $2$, it is natural to take $\bm{\omega}=(0,1)$, i.e., $k^*=1$ in \eqref{actfor} with $z_1$ being the active variable and 
 $z_2$  the inactive one in the transformed space of $\bm{z}$.  
For the same reason, two hidden layers are used for each of the FCNNs representing $\mathbf{g}$ and $\mathbf{h}$, different from the default settings.
From  the tests reported in \Cref{sec:rev,sec:bdl}, we observe that the Adam optimization during PRNN training terminated within 20000 steps in all cases, while the tests 
in \Cref{sec:loc}  required up to 60000 steps to meet the stopping criterion due to more complicated geometric structures in the target function.

To visually evaluate the function approximation, we present two types of plots: 
The {\em quiver plot} shows the gradient field of $f$ (blue arrows) and the vector field corresponding to the second Jacobian column $\bm{J}_2$ (red arrows) on a $15\times 15$ uniform grid, where increased orthogonality between the red and blue arrows indicates increased accuracy in the network mapping;
 The {\em regression plot} draws the approximated function values (red circles) over 400 randomly generated points in the domain  together with the associated exact function values (blue stars), where good performance is indicated by a thin regression curve and a large degree of overlap between the blue stars and the red circles (exact and approximate function values).

\subsubsection{PRNN vs. RevNet and the effect of $\alpha$}\label{sec:rev}
One of the main differences between the proposed PRNN and the RevNet is their treatments of {\em reversibility}: the former imposes it as a soft constraint while the latter imposes a hard constraint (realized by a special network structure). Furthermore, the special structure of the RevNet  requires an equal separation of the inputs into two groups. Thus, if the input space has an odd dimension, it has to be padded with an auxiliary variable (e.g., a column of zero). On the other hand, the PRNN represents a larger class of functions than the RevNet \cite{DBLP:conf/aaai/ChangMHRBH18}, so that a better nonlinear transformation can be found when there is no need for explicit invertibility. To compare these two neural network structures, the following two functions are considered for testing: 
\begin{align}
f_1(\bm{x}) = x_1^2 + x_2^2 \quad \text{and} \quad f_2(\bm{x}) &= \frac{5}{8} x_1^2 + \frac{5}{8} x_2^2 - \frac{3}{4} x_1 x_2,  
\end{align}
where the domain of $\bm{x}$ is either $\Omega^2_{A}=[0,1]^2$ or $\Omega^2_{B}=[-1,1]^2$. Note that both $f_1$ and $f_2$ reach their minimum at the origin, which is located in the interior of $\Omega^2_{B}$ but only on the boundary of $\Omega^2_{A}$. 
Since we focus on the influence of reversibility in this subsection, we temporarily set $\lambda_1 = 1$ and $\lambda_2=0$. 
The corresponding total loss  $\mathcal{L}$ for our DRiLLS method defined by \eqref{totalloss} then becomes
\begin{align*}
\mathcal{L}_{\mathtt{PRNN}}\coloneqq \mathcal{L}_{1}+ \mathcal{L}_{2} \quad \text{and} \quad 
\mathcal{L}_{\mathtt{RevNet}}\coloneqq \mathcal{L}_{2},
\end{align*}
respectively, because $\mathcal{L}_1$ is automatically zero in the case of RevNet. 
In the following tests, the RevNet uses 10 RevNet Blocks with 2 neurons each, as the input space has the dimension two, and a step size $0.25$ (see \cite{zhang2019learning} for details about the used RevNet structure). 
 We choose the size of the sample dataset for training to be $N=500$  and both the PRNN and the RevNet are trained  using the same dataset.

\begin{figure}[!htbp]
\hspace{-0.3cm}
\subfigure{
\begin{minipage}[t]{0.33\linewidth}
\centering
  \texttt{PRNN} with $\alpha=0$
\includegraphics[width=1.65in]{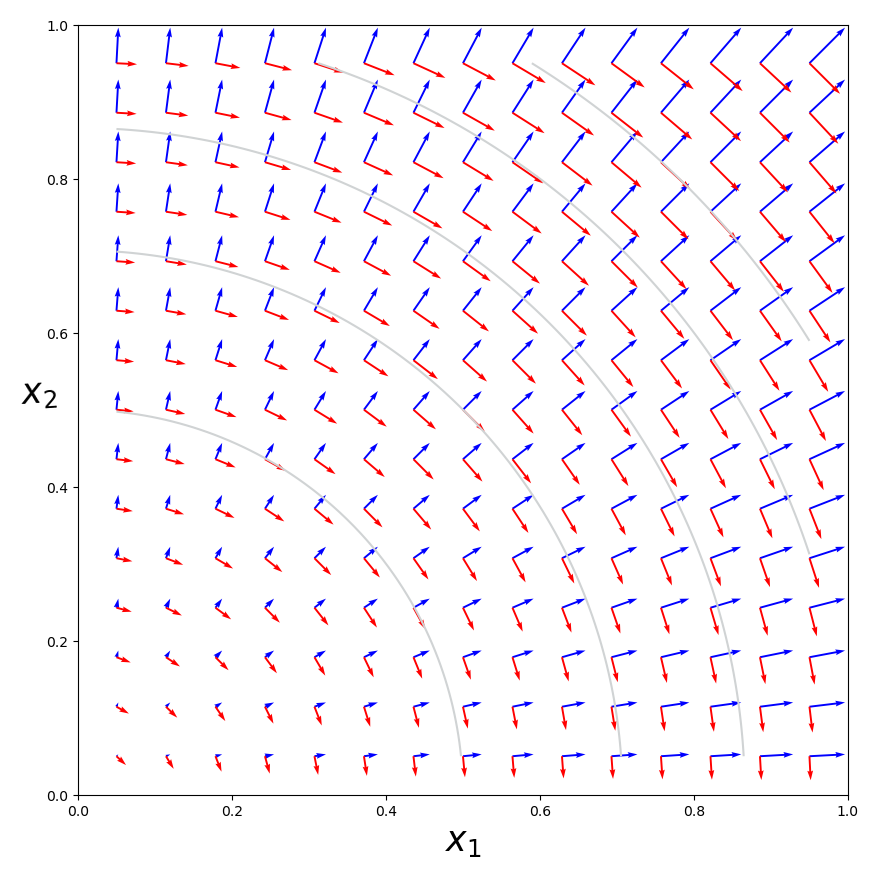}
\includegraphics[width=1.6in]{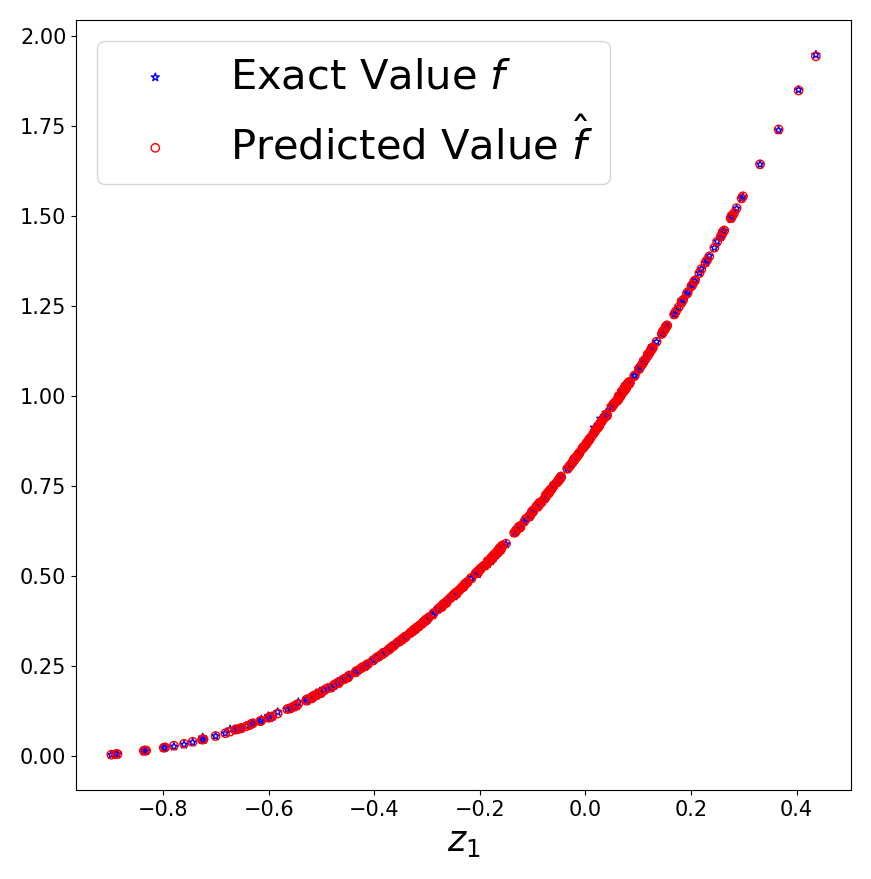}
  \texttt{RevNet} with $\alpha=0$
\includegraphics[width=1.65in]{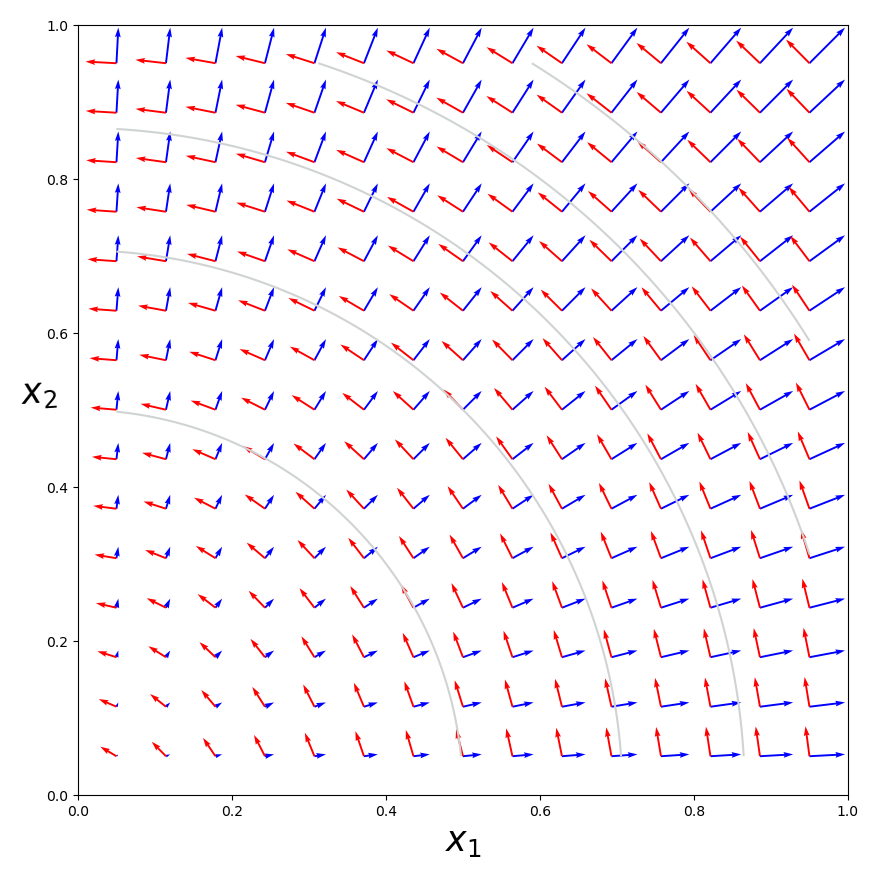}
\includegraphics[width=1.6in]{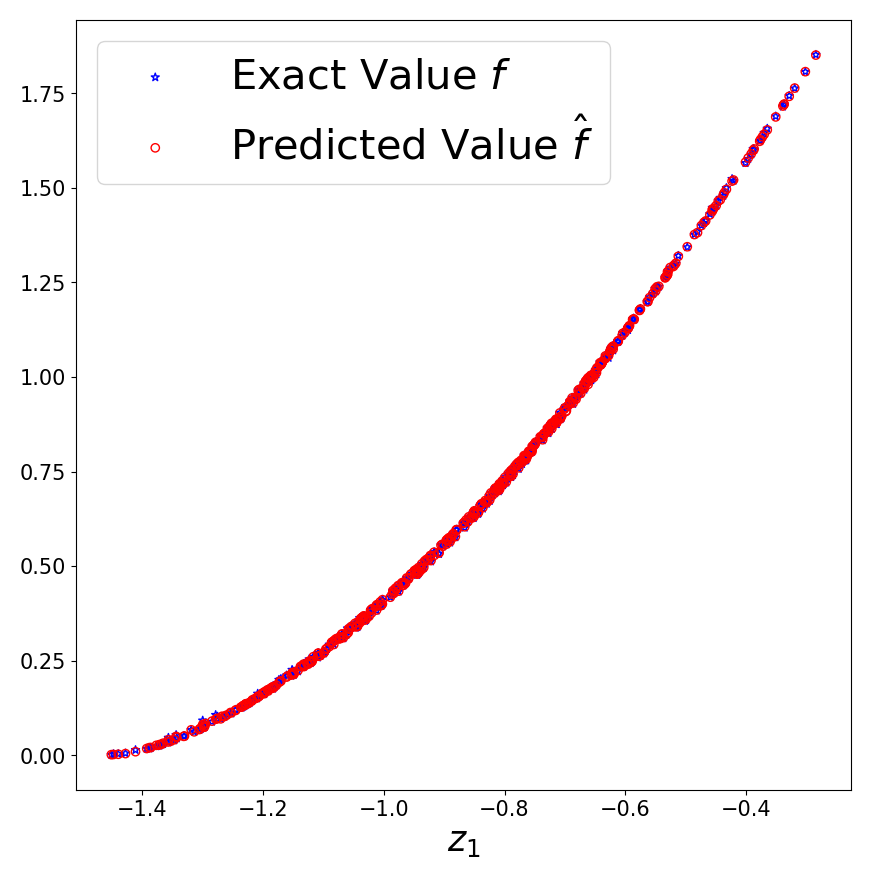}
%\caption{fig1}
\end{minipage}%
}%
\subfigure{
\begin{minipage}[t]{0.33\linewidth}
\centering
  \texttt{PRNN} with $\alpha=25$
\includegraphics[width=1.65in]{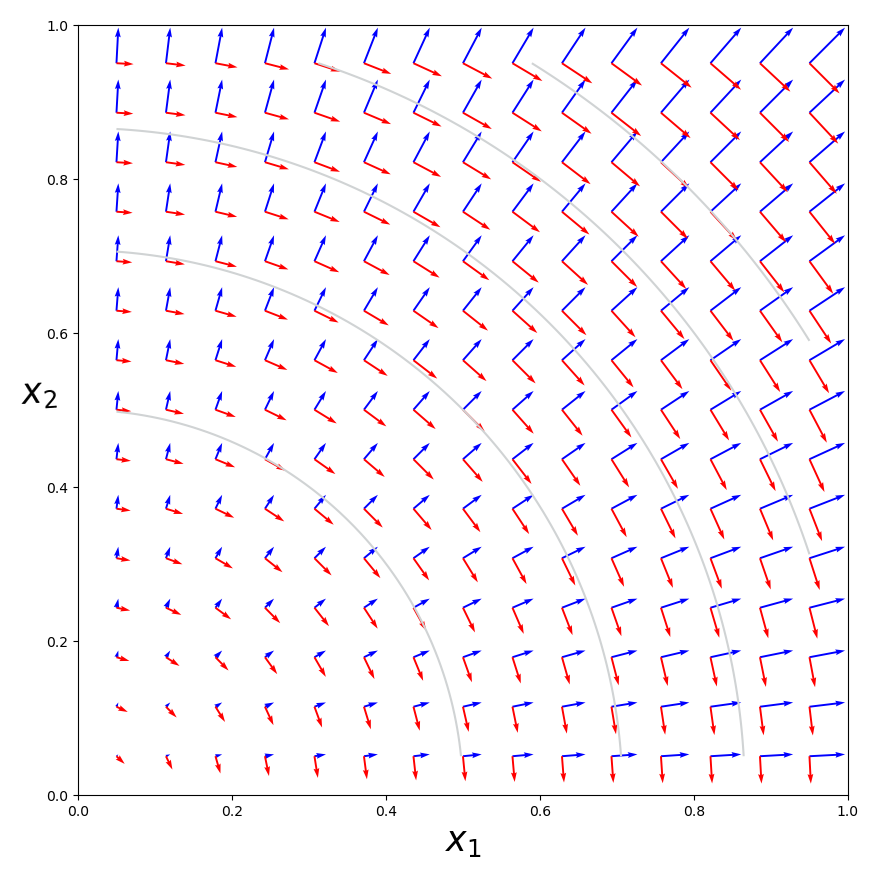}
\includegraphics[width=1.6in]{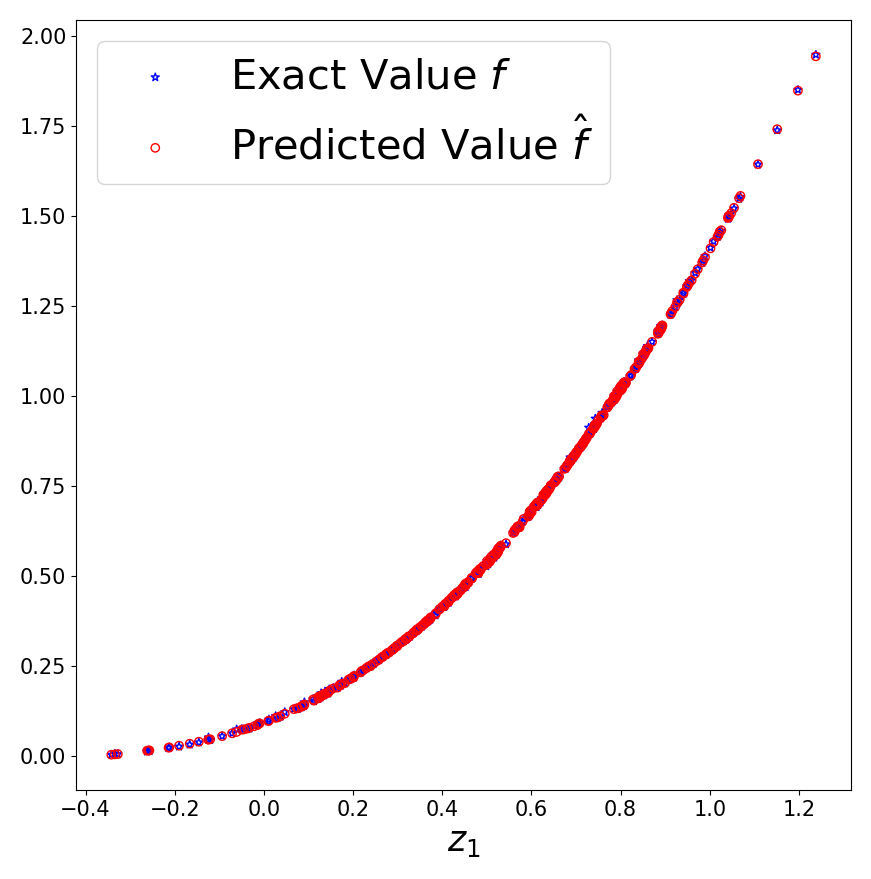}
  \texttt{RevNet} with $\alpha=25$
\includegraphics[width=1.65in]{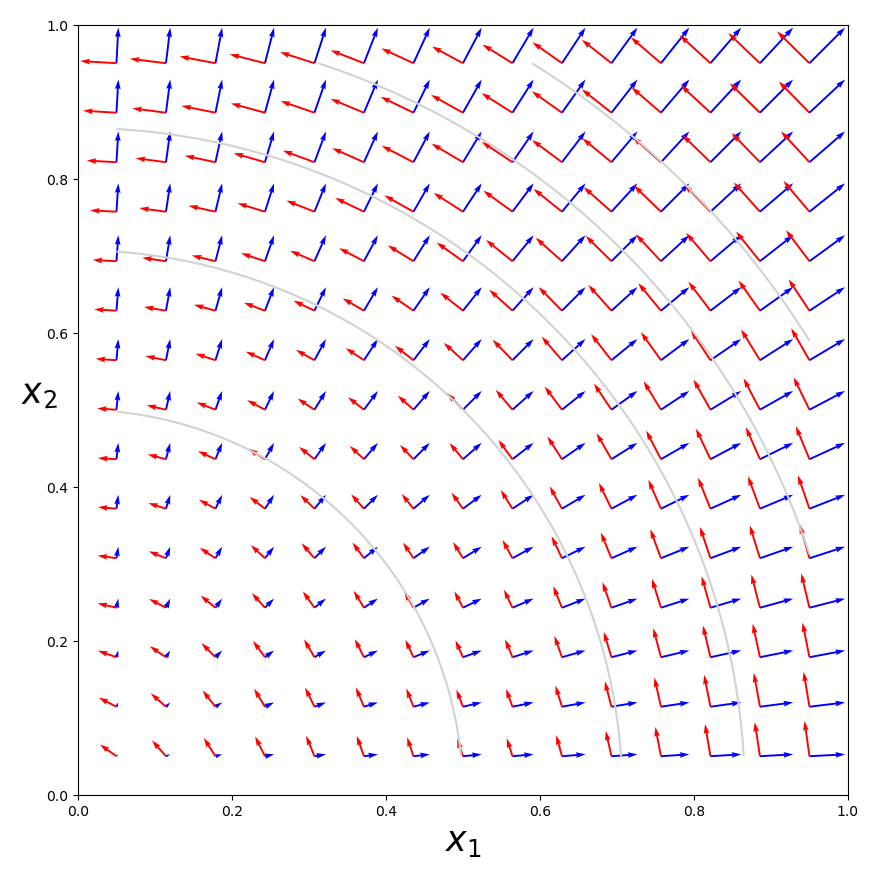}
\includegraphics[width=1.6in]{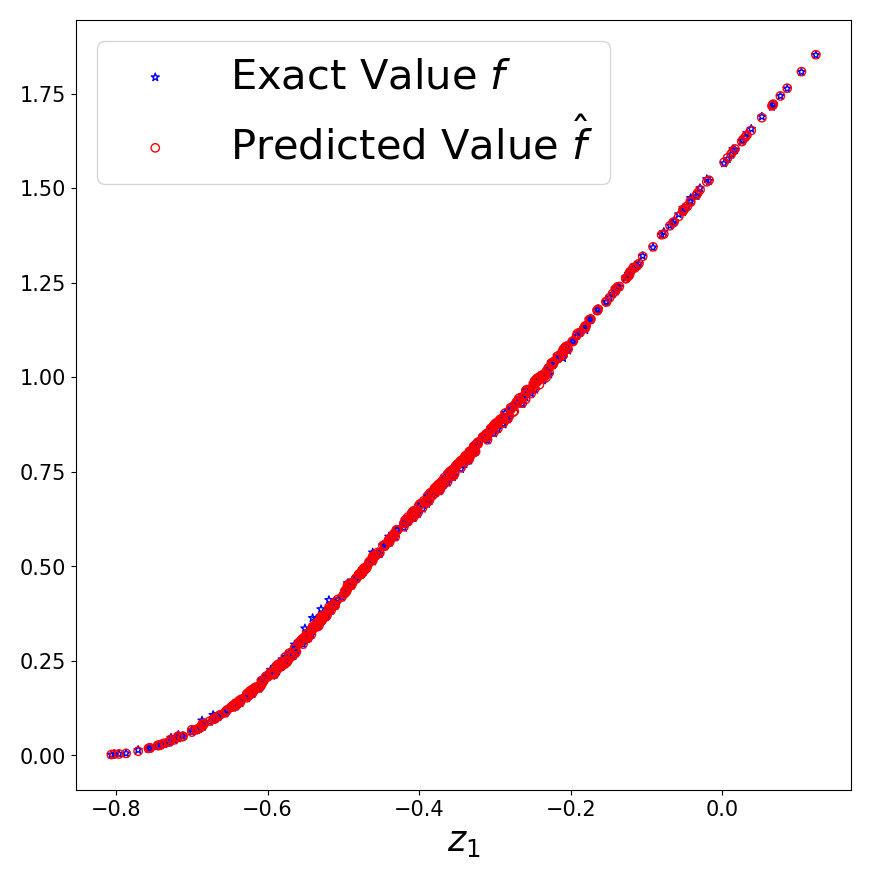}
%\caption{fig1}
\end{minipage}%
}%
\subfigure{
\begin{minipage}[t]{0.33\linewidth}
\centering
  \texttt{PRNN} with $\alpha=50$
\includegraphics[width=1.65in]{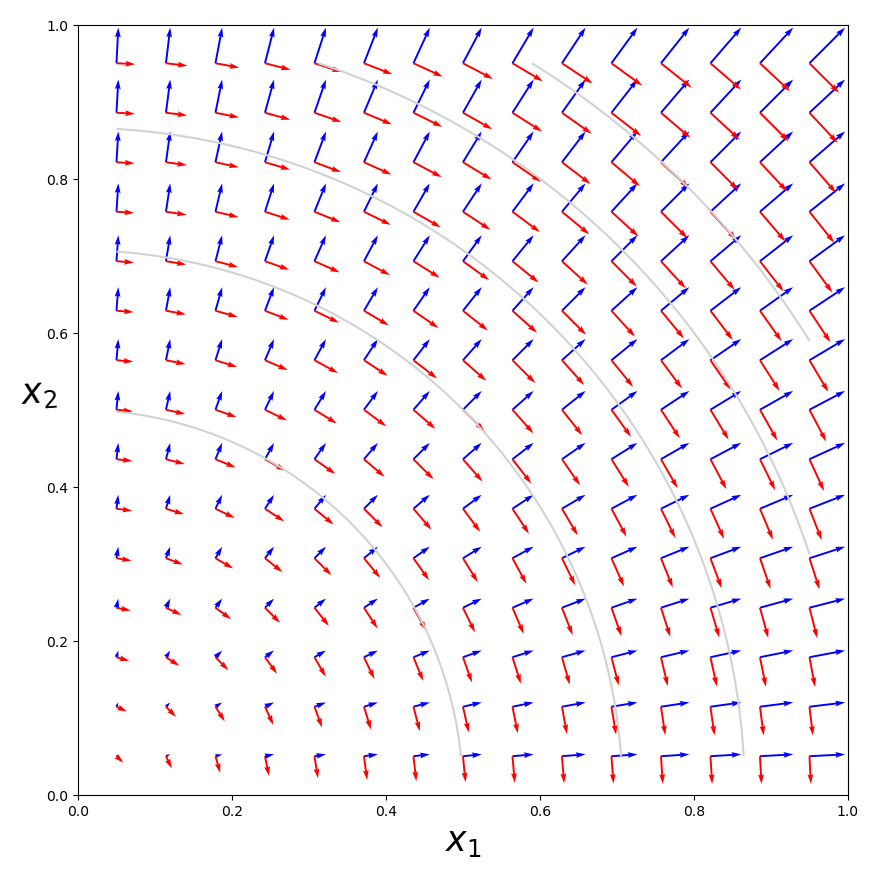}
\includegraphics[width=1.6in]{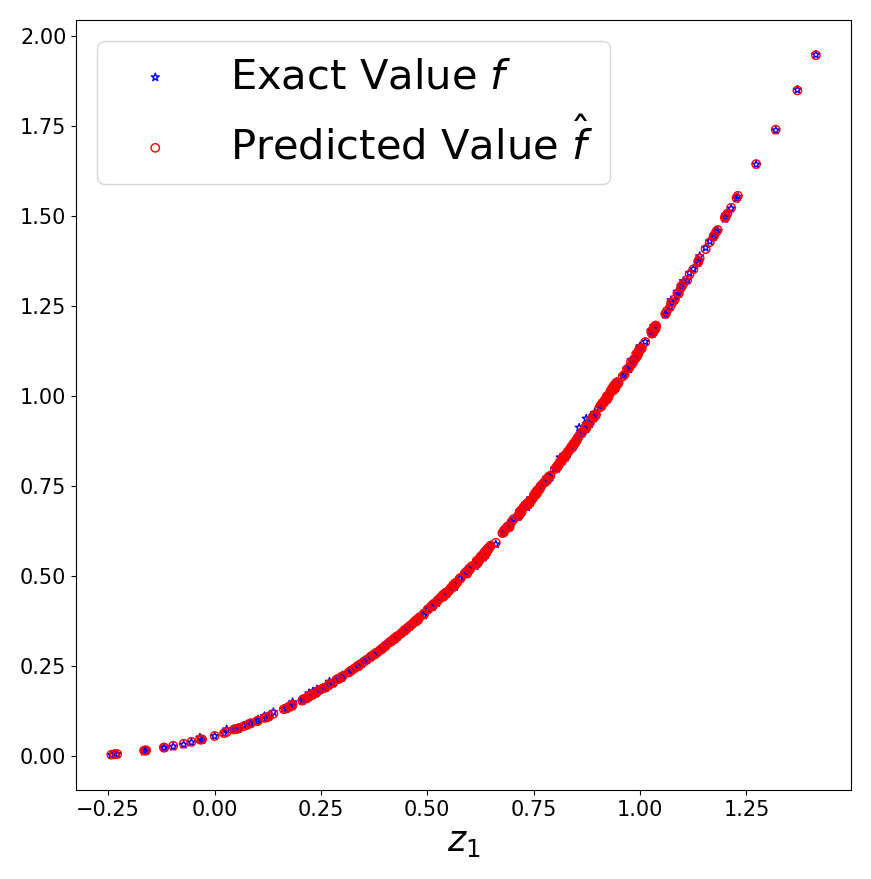}
  \texttt{RevNet} with $\alpha=50$
\includegraphics[width=1.65in]{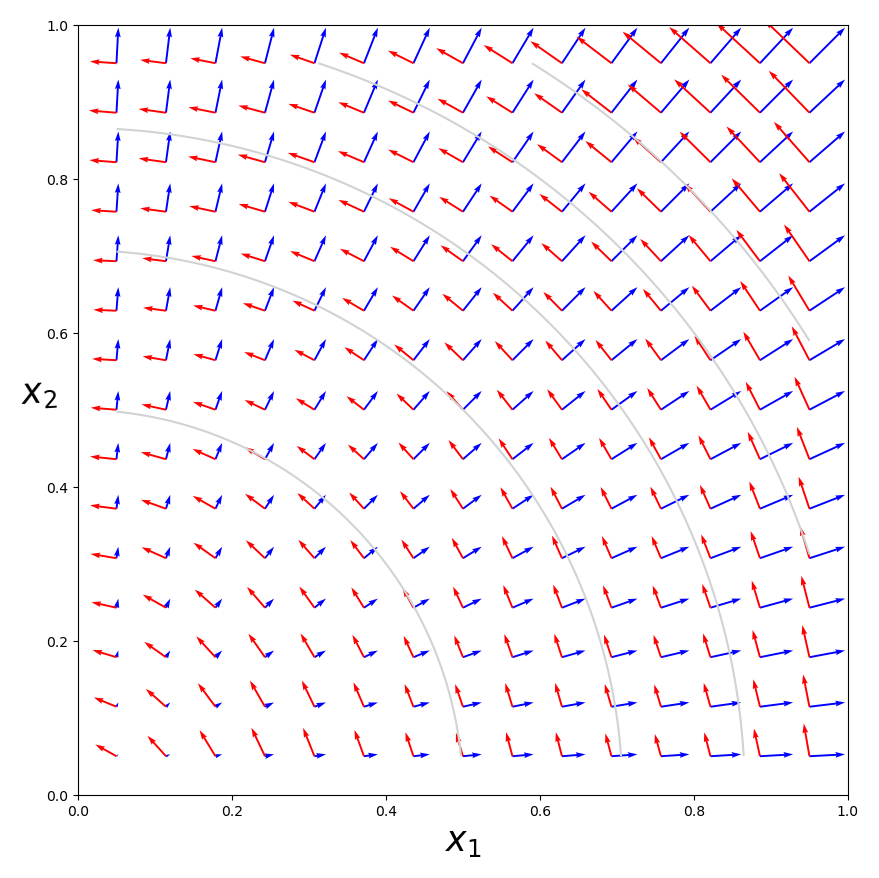}
\includegraphics[width=1.6in]{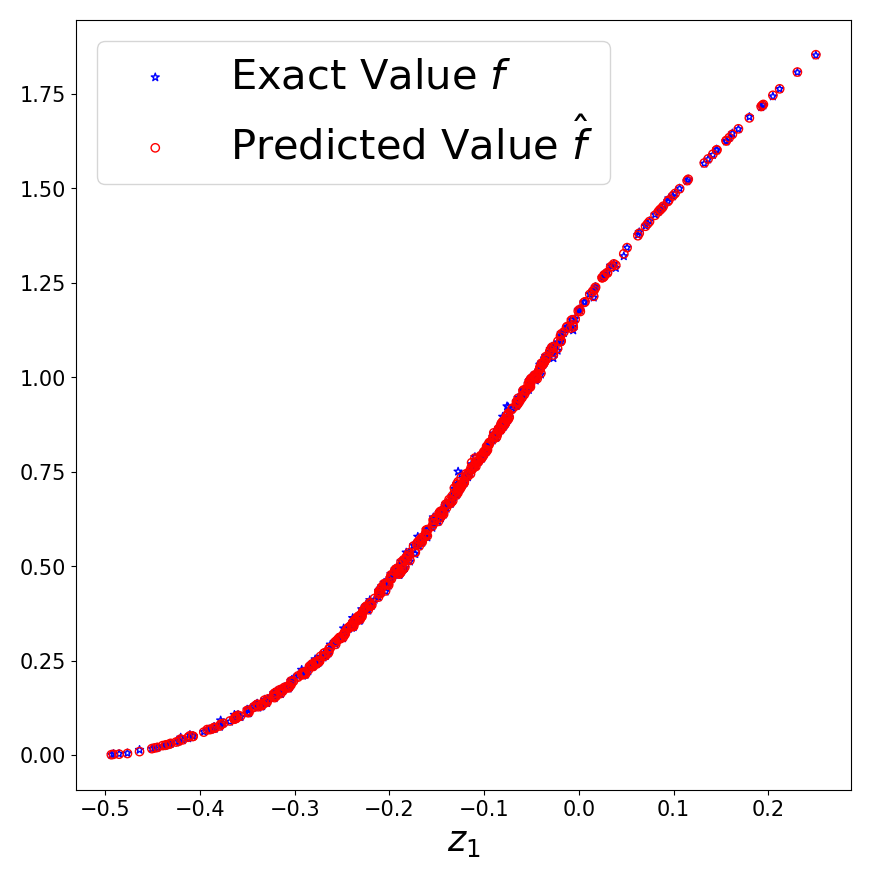}
%\caption{fig1}
\end{minipage}%
}%
    \vspace{-0.4cm}
\caption{Level set learning and function approximation results produced by the our DRiLLS method with PRNN (the quiver plot in Row 1 and the regression plot in Row 2) or the RevNet (the quiver plot in Row 3 and the regression plot in Row 4) for $f_1(\bm{x}) = x_1^2 + x_2^2$ in $\Omega^2_{A}=[0,1]^2$, at three different values of $\alpha$ = 0, 25, 50, respectively. There is no critical point in the interior of the domain $\Omega_A$ and both PRNN and RevNet successfully learn the level sets of the target function.
}
\label{f_1_01}
\end{figure}

\begin{figure}[!htbp]
\hspace{-0.3cm}
\subfigure{
\begin{minipage}[t]{0.33\linewidth}
\centering
  \texttt{PRNN} with $\alpha=0$
\includegraphics[width=1.65in]{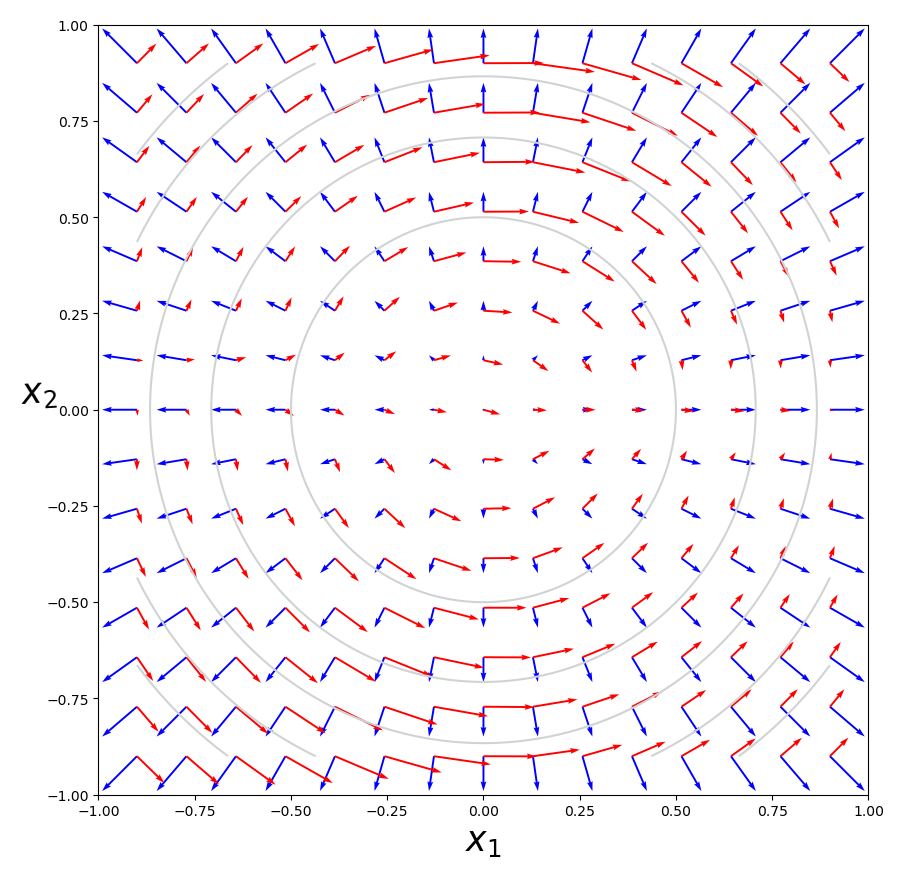}
\includegraphics[width=1.6in]{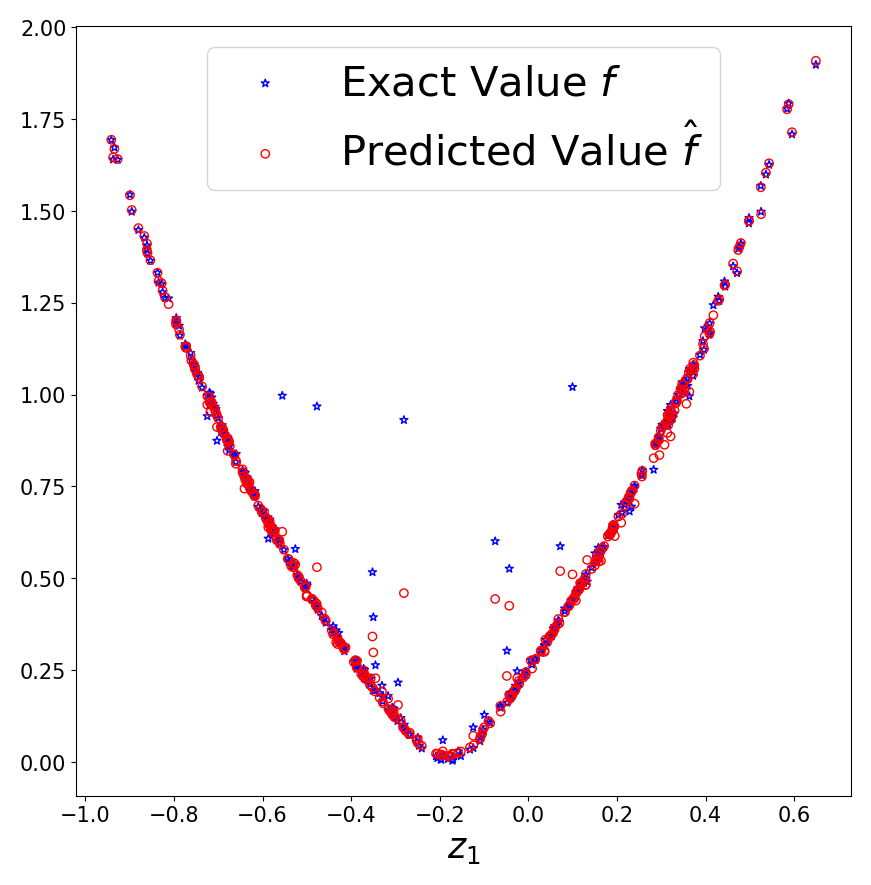}
  \texttt{RevNet} with $\alpha=0$
\includegraphics[width=1.65in]{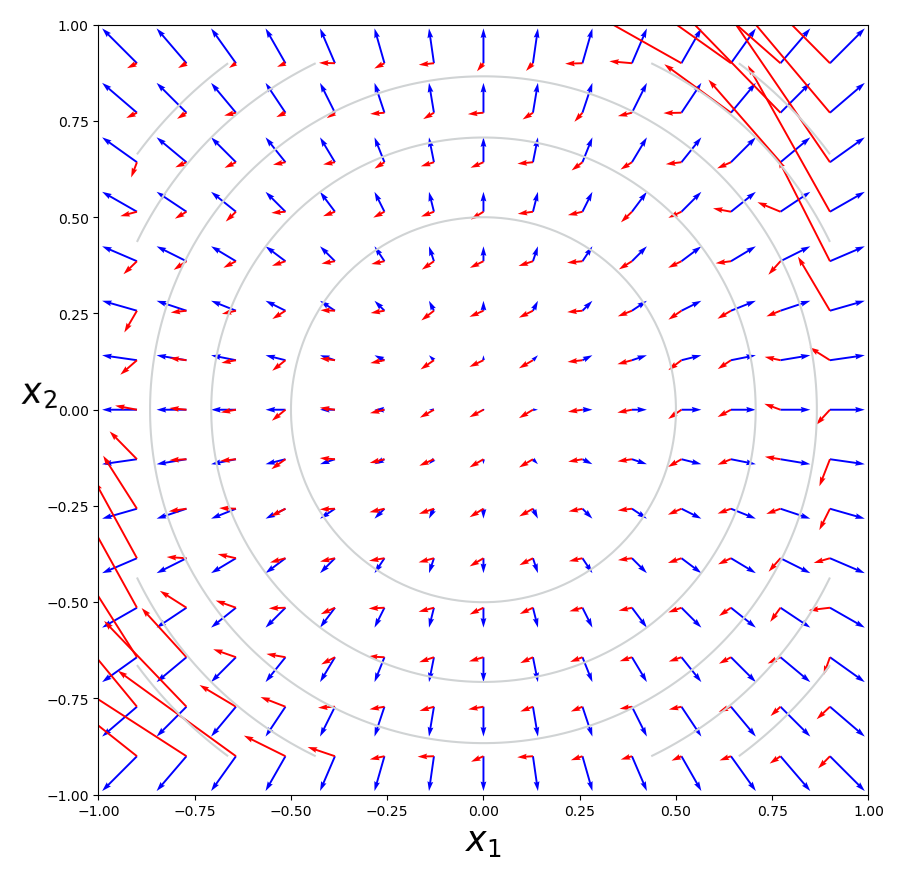}
\includegraphics[width=1.6in]{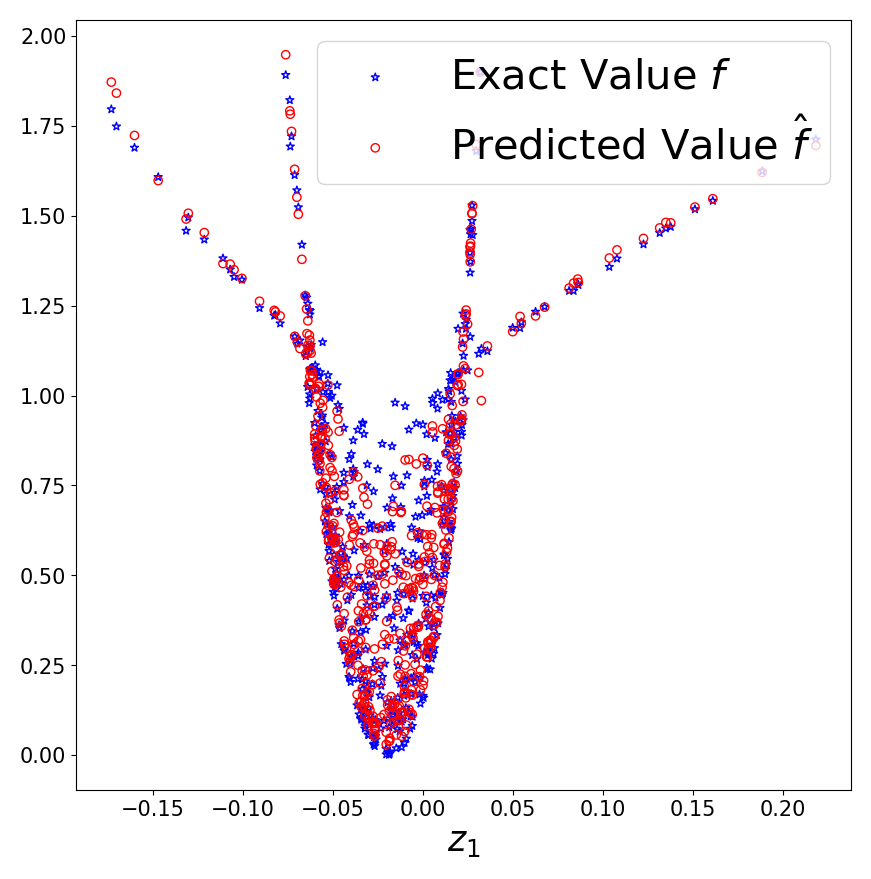}
%\caption{fig1}
\end{minipage}%
}%
\subfigure{
\begin{minipage}[t]{0.33\linewidth}
\centering
  \texttt{PRNN} with $\alpha=25$
\includegraphics[width=1.65in]{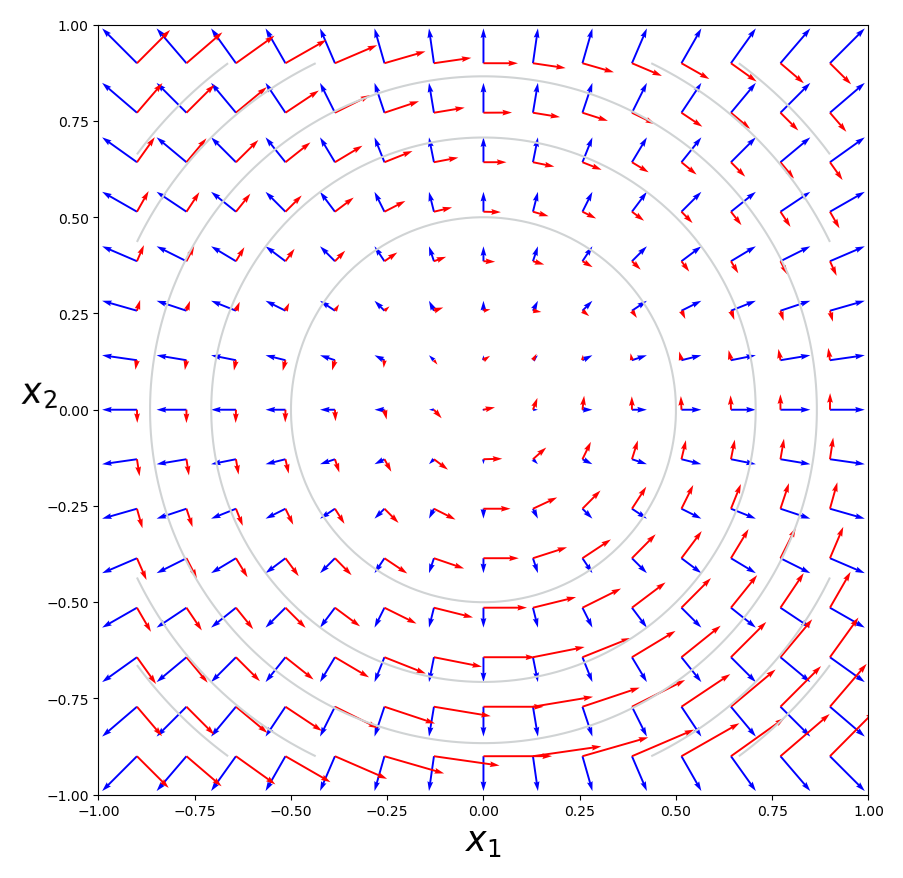}
\includegraphics[width=1.6in]{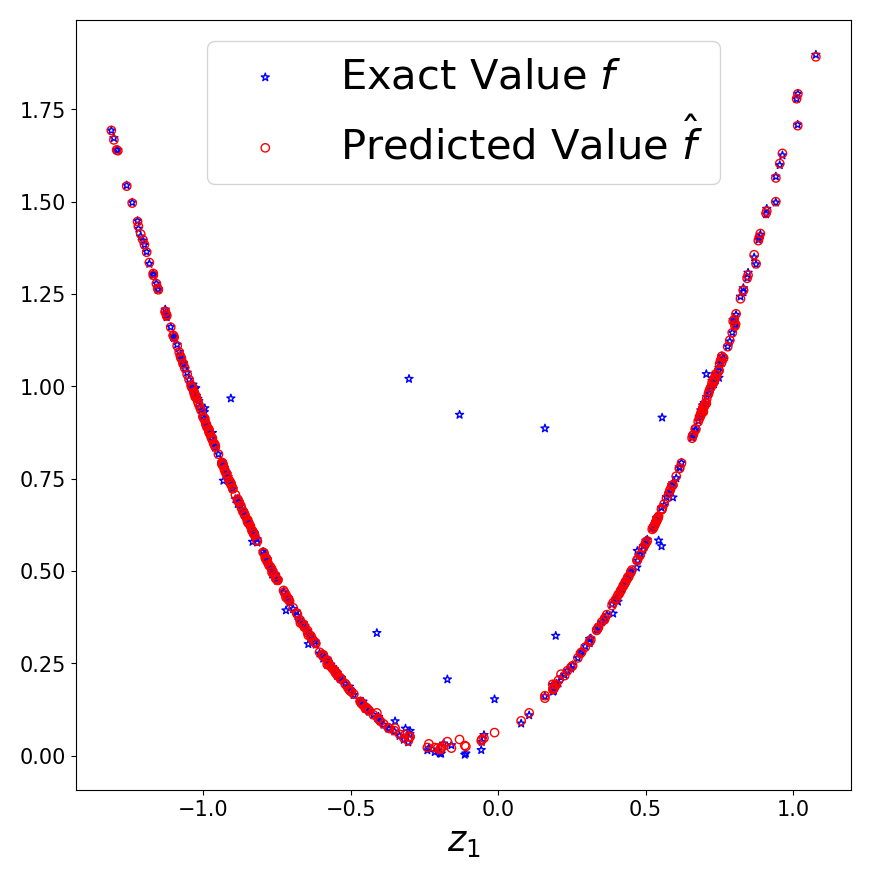}
  \texttt{RevNet} with $\alpha=25$
\includegraphics[width=1.65in]{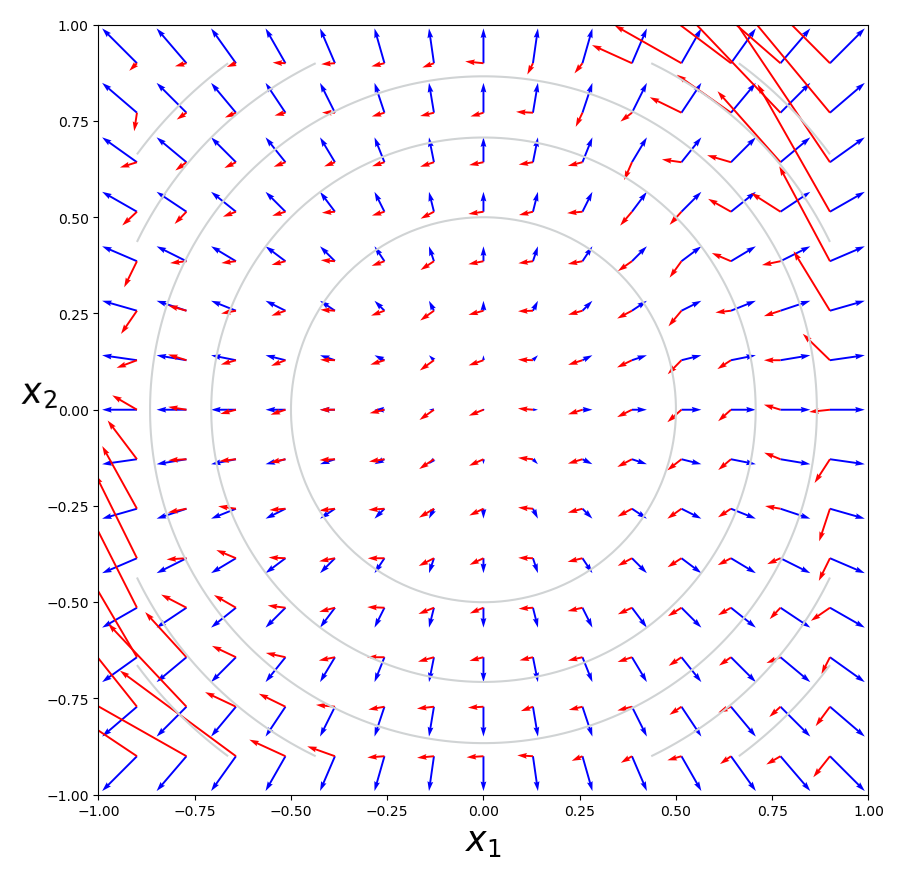}
\includegraphics[width=1.6in]{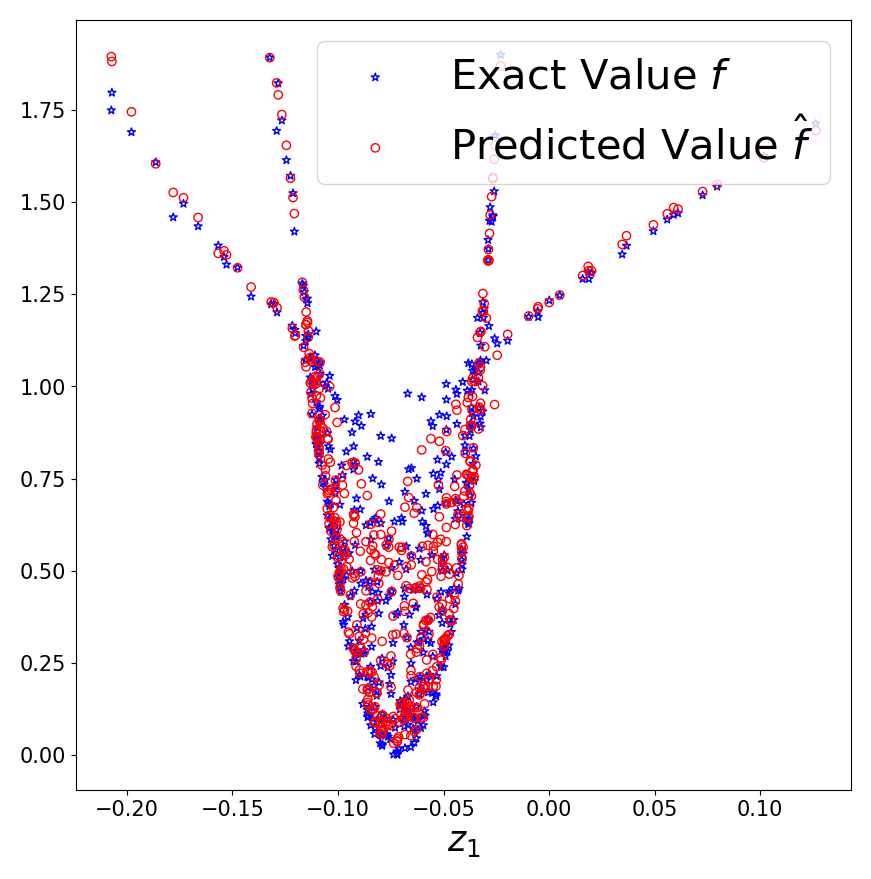}
%\caption{fig1}
\end{minipage}%
}%
\subfigure{
\begin{minipage}[t]{0.33\linewidth}
\centering
  \texttt{PRNN} with $\alpha=50$
\includegraphics[width=1.65in]{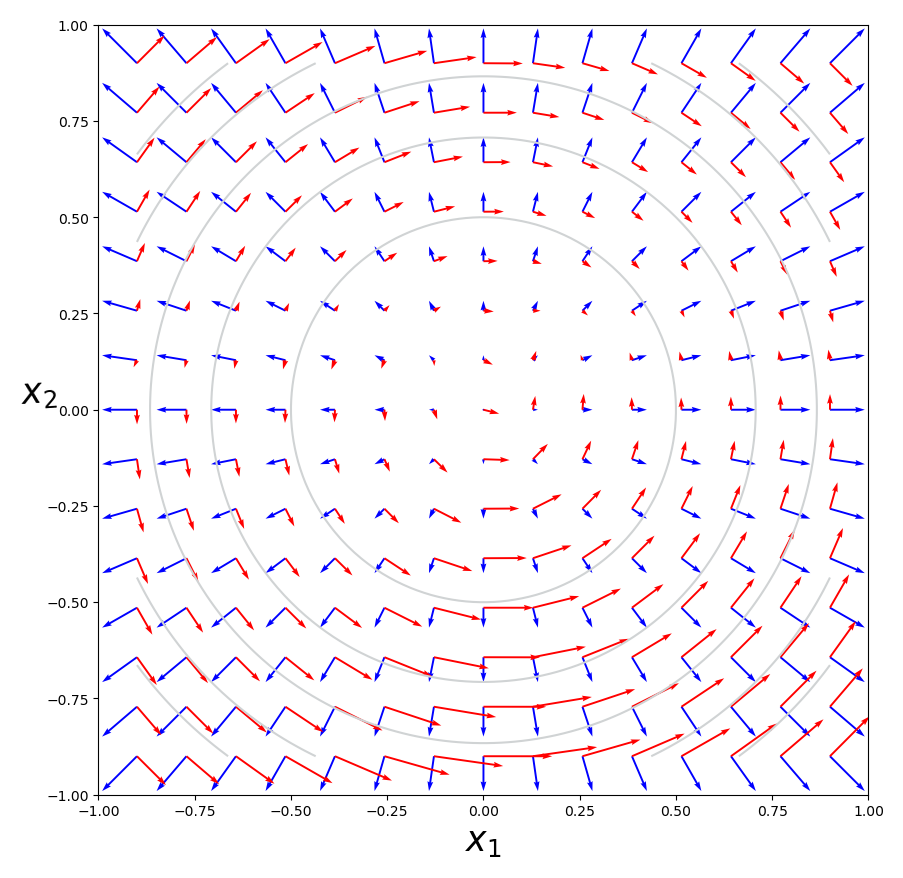}
\includegraphics[width=1.6in]{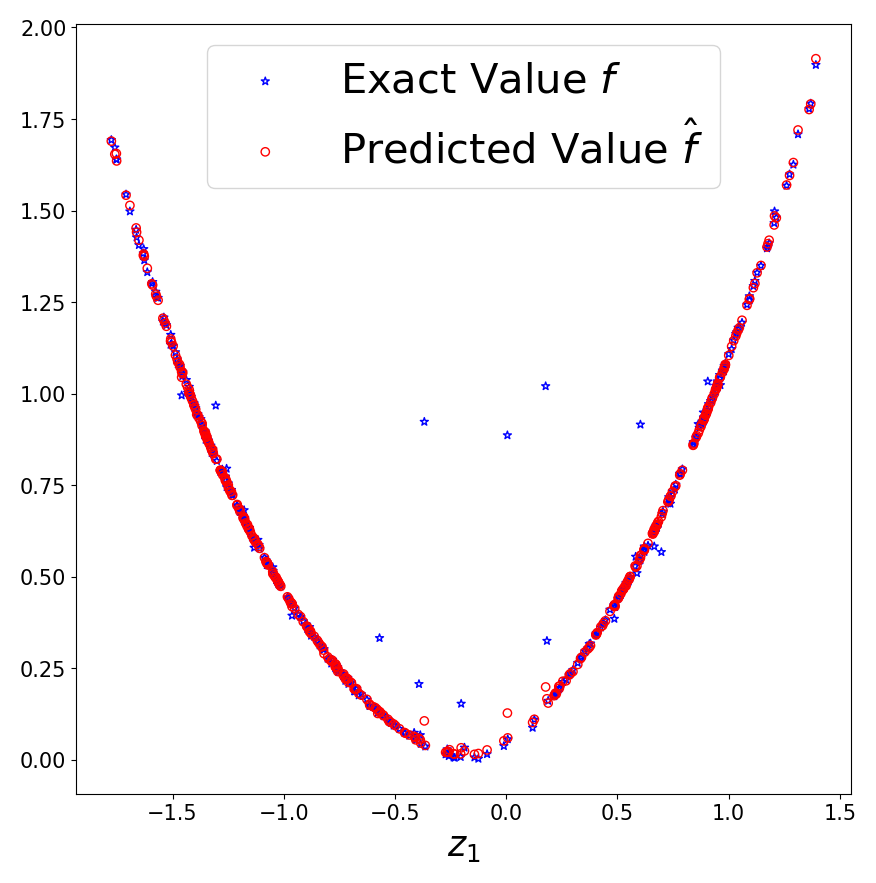}
  \texttt{RevNet} with $\alpha=50$
\includegraphics[width=1.65in]{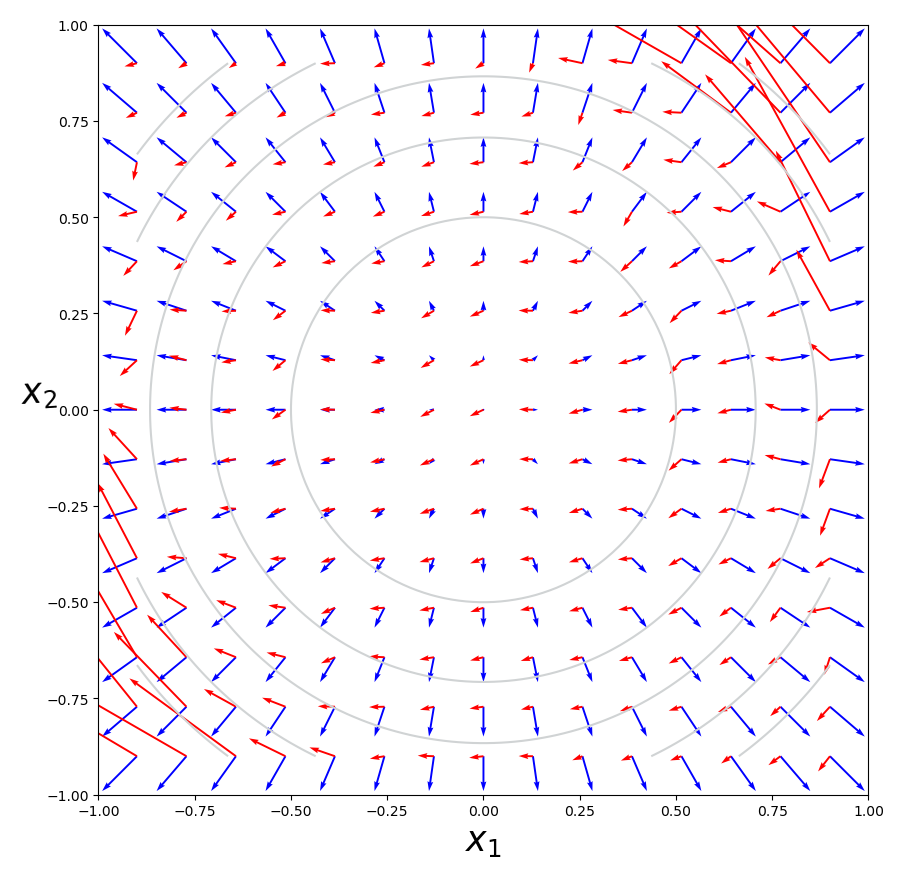}
\includegraphics[width=1.6in]{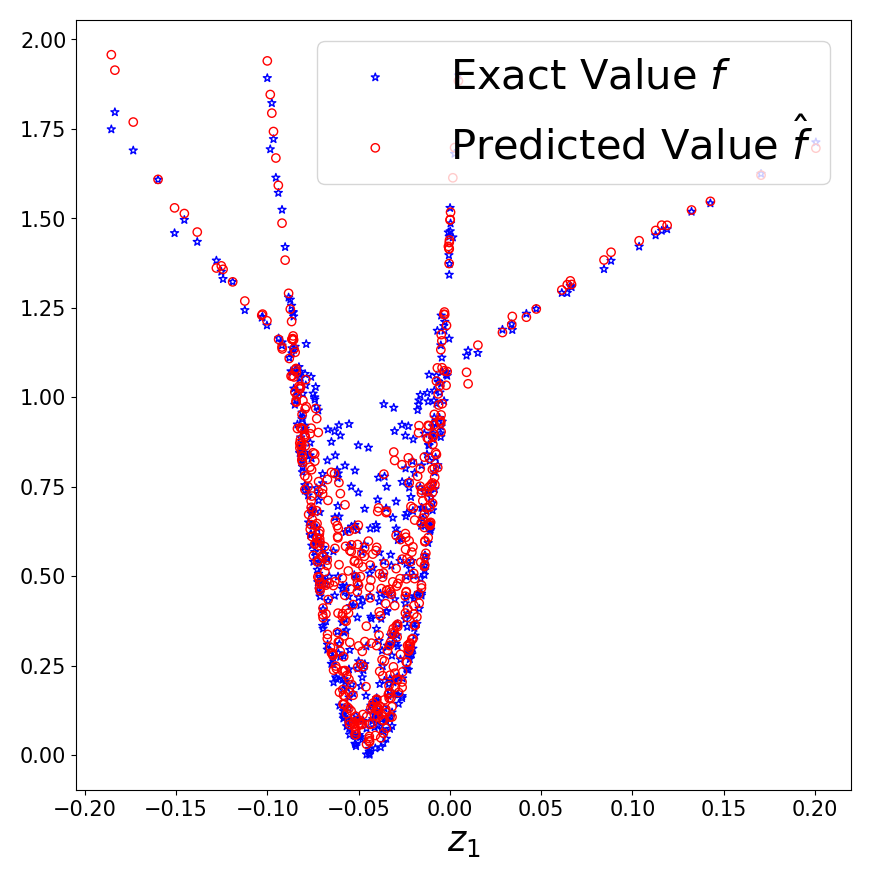}
%\caption{fig1}
\end{minipage}%
}%
    \vspace{-0.4cm}
\caption{Level set learning and function approximation results produced by our DRiLLS method with the PRNN (the quiver plot in Row 1 and the regression plot in Row 2) or the RevNet (the quiver plot in Row 3 and the regression plot in Row 4) for $f_1(\bm{x}) = x_1^2 + x_2^2$ in $\Omega^2_{B}=[-1,1]^2$, at three different values of $\alpha$ = 0, 25, 50, respectively. RevNet fails to learn the level sets of the target function because it cannot handle the interior critical point at the origin. In comparison, the PRNN successfully learns these level sets partly because it does not enforce the hard reversibility around the critical point.
}
\label{f_1_11}
\end{figure}

The testing results of $f_1(\bm{x})$ are presented in \Cref{f_1_01} for the case $\bm{x}\in \Omega^2_{A}$ and in \Cref{f_1_11} for the case $\bm{x}\in \Omega^2_{B}$, where several choices of $\alpha$ are considered, i.e., the first column for $\alpha=0$, the second column for $\alpha=25$, and the  third column for $\alpha=50$. It is observed that both network structures, the PRNN and the RevNet, work well for $f_1$ with the domain $\Omega^2_{A}$ as shown in \Cref{f_1_01}. It is worth noting that $\frac{\partial f_1}{\partial x_1}(\bm{x}) \geq 0$ and  $\frac{\partial f_1}{\partial x_2}(\bm{x})  \geq 0$ for any $\bm{x}\in \Omega^2_{A}$, i.e, the behavior of $f_1$ in $\Omega^2_{A}$ is somehow monotonic.
However, when the domain is changed to $\Omega^2_{B}$, the behavior of $f_1$ in $\Omega^2_{A}$ is not monotonic anymore and the RevNet encounters difficulties in finding the appropriate active variable. Indeed, as shown in the third row of \Cref{f_1_11}, the gradient is not orthogonal to $\mathbf{J}_2$ at many points no matter  the value of $\alpha$ is, which indicates the function value is still sensitive to the first inactivate variable $z_2$. This further leads larger errors in the regression process and function approximation, as seen in the fourth row of \Cref{f_1_11}. 

\begin{figure}[!htbp]
\hspace{-0.3cm}
\subfigure{
\begin{minipage}[t]{0.33\linewidth}
\centering
  \texttt{PRNN} with $\alpha=0$
\includegraphics[width=1.65in]{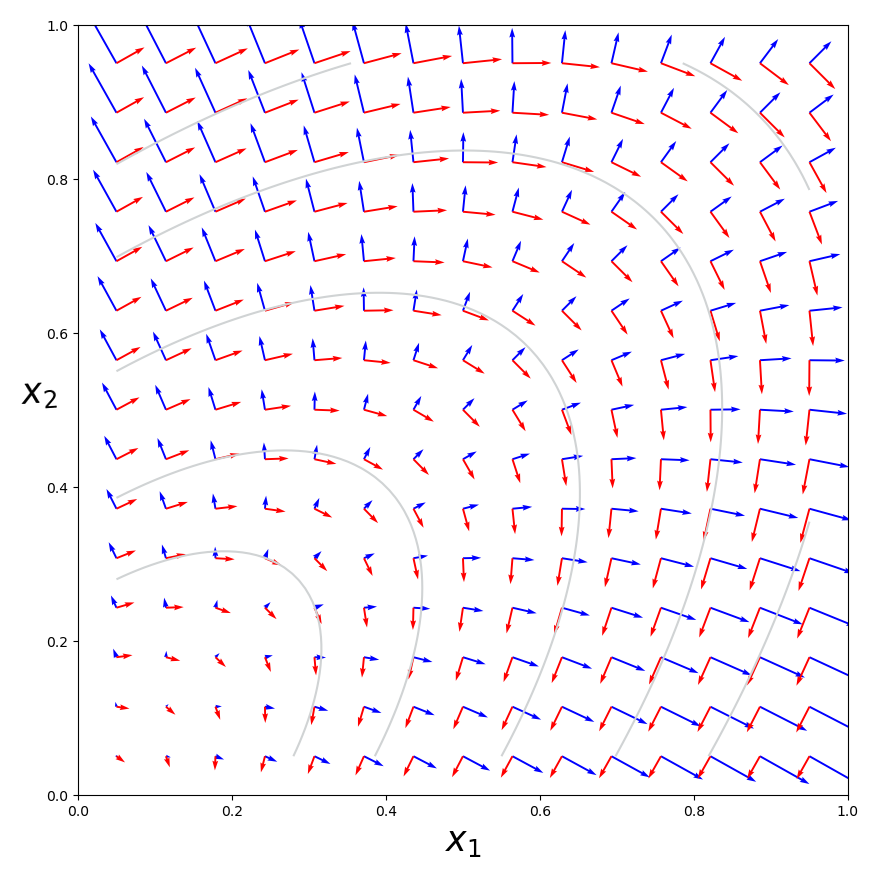}
\includegraphics[width=1.6in]{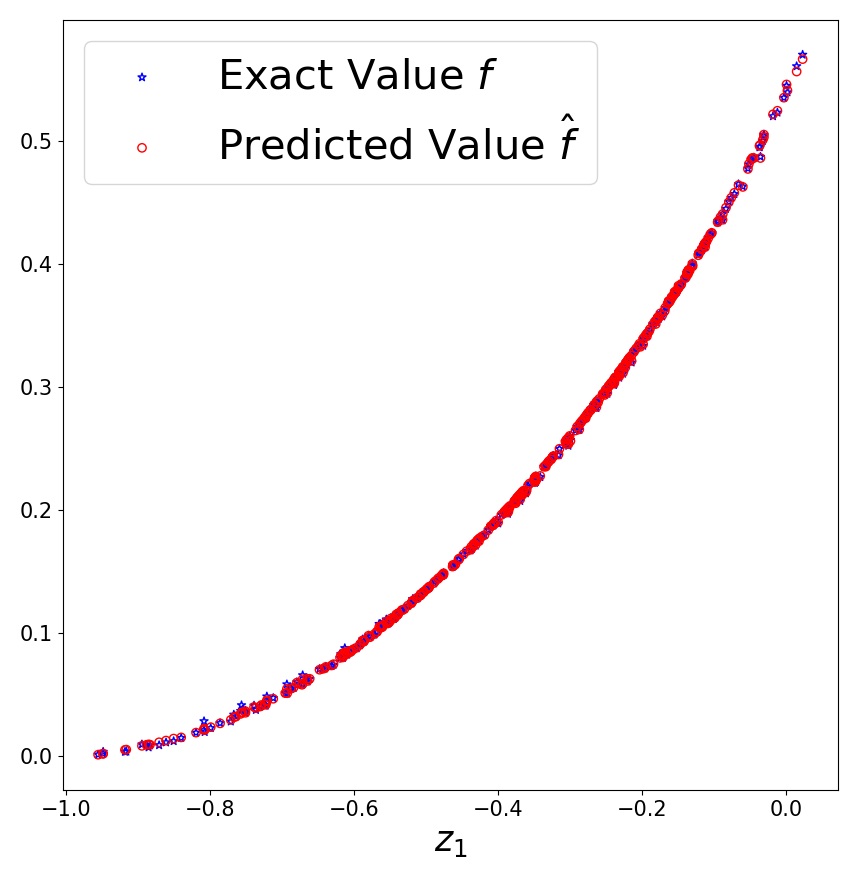}
  \texttt{RevNet} with $\alpha=0$
\includegraphics[width=1.65in]{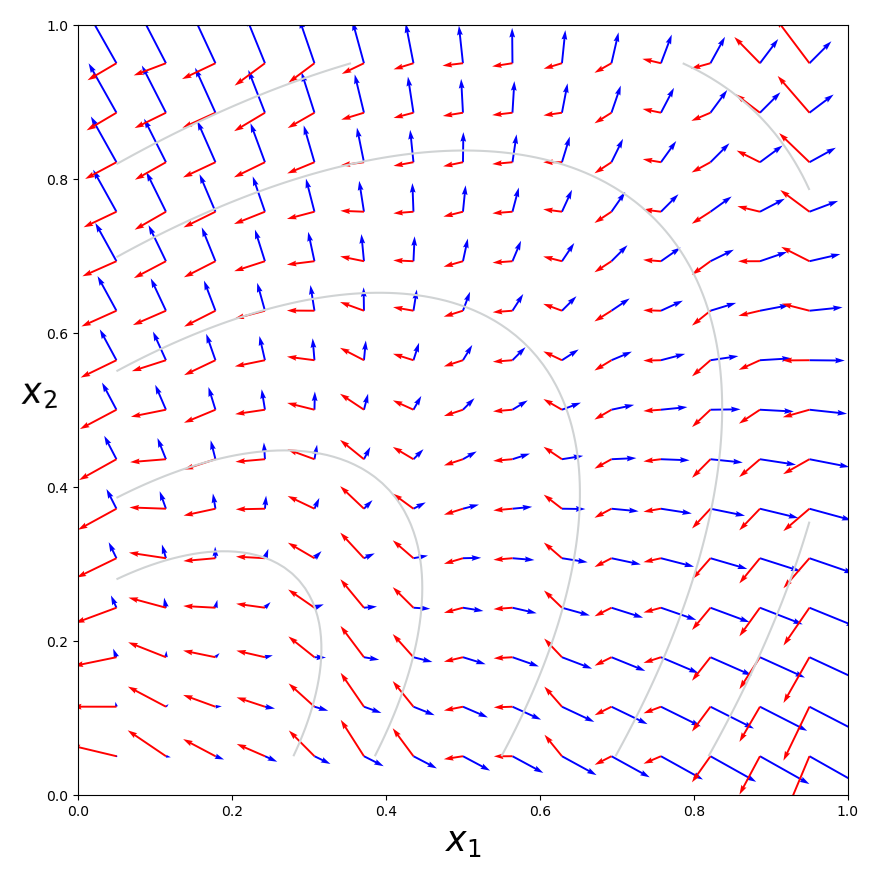}
\includegraphics[width=1.6in]{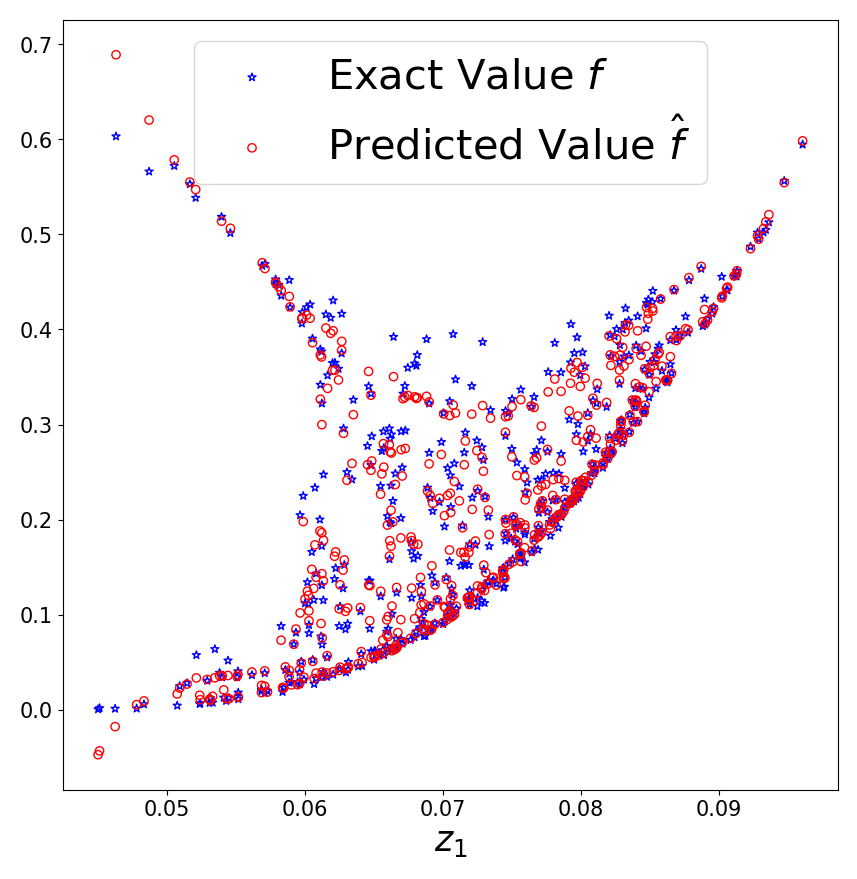}
%\caption{fig1}
\end{minipage}%
}%
\subfigure{
\begin{minipage}[t]{0.33\linewidth}
\centering
  \texttt{PRNN} with $\alpha=25$
\includegraphics[width=1.65in]{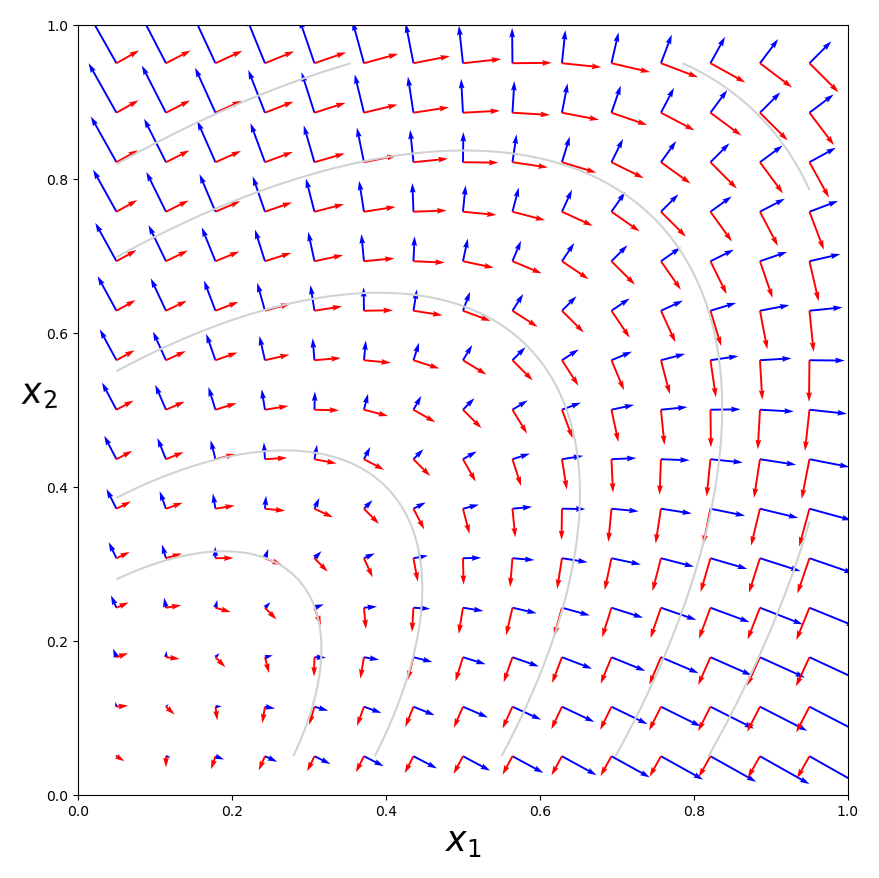}
\includegraphics[width=1.6in]{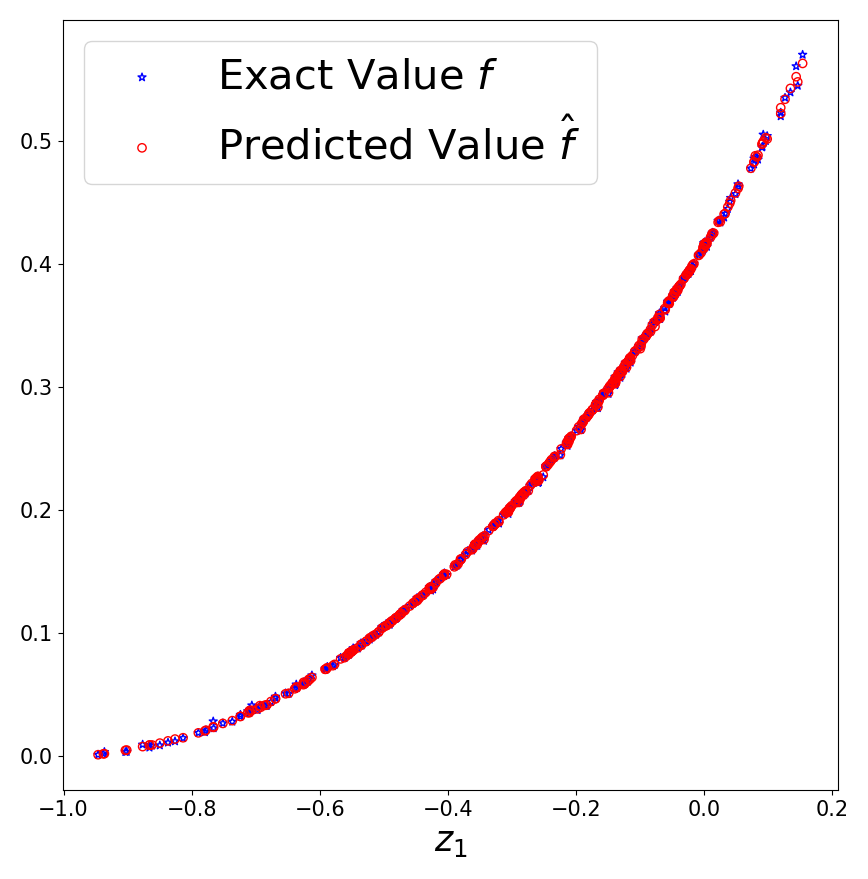}
  \texttt{RevNet} with $\alpha=25$
\includegraphics[width=1.65in]{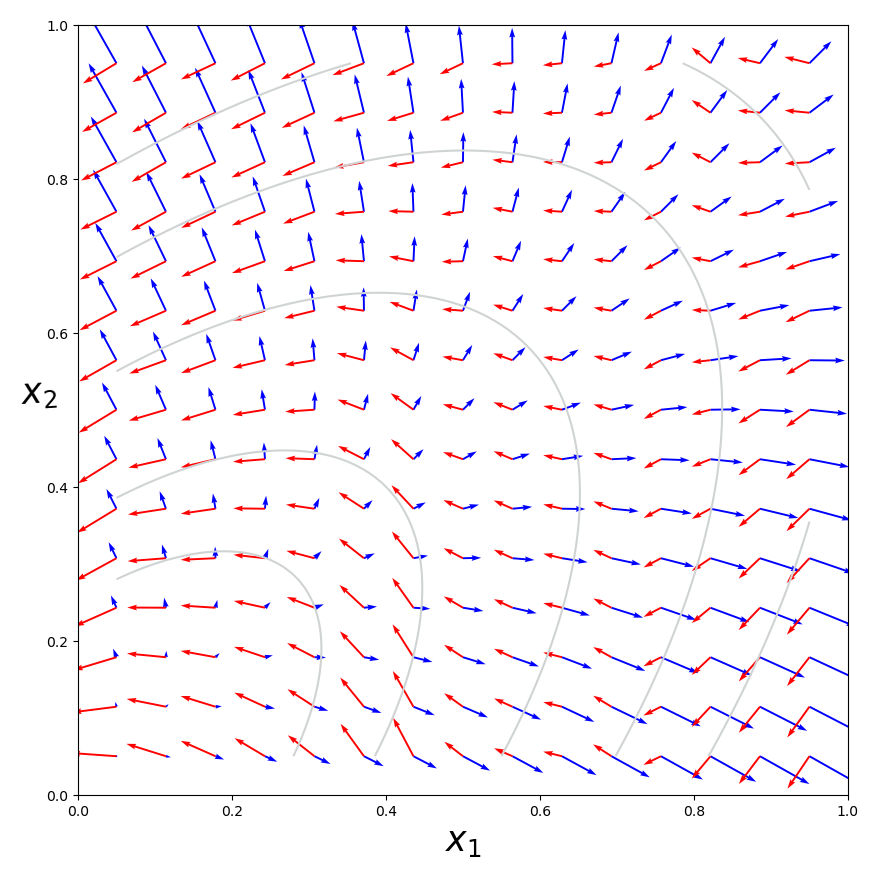}
\includegraphics[width=1.6in]{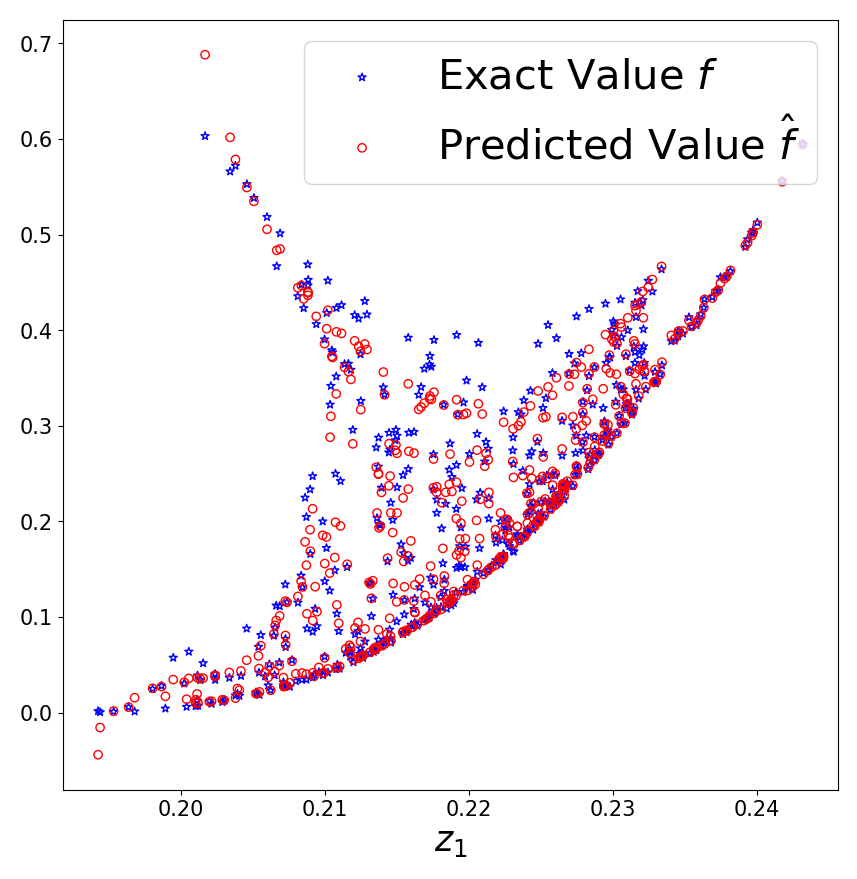}
%\caption{fig1}
\end{minipage}%
}%
\subfigure{
\begin{minipage}[t]{0.33\linewidth}
\centering
  \texttt{PRNN} with $\alpha=50$
\includegraphics[width=1.65in]{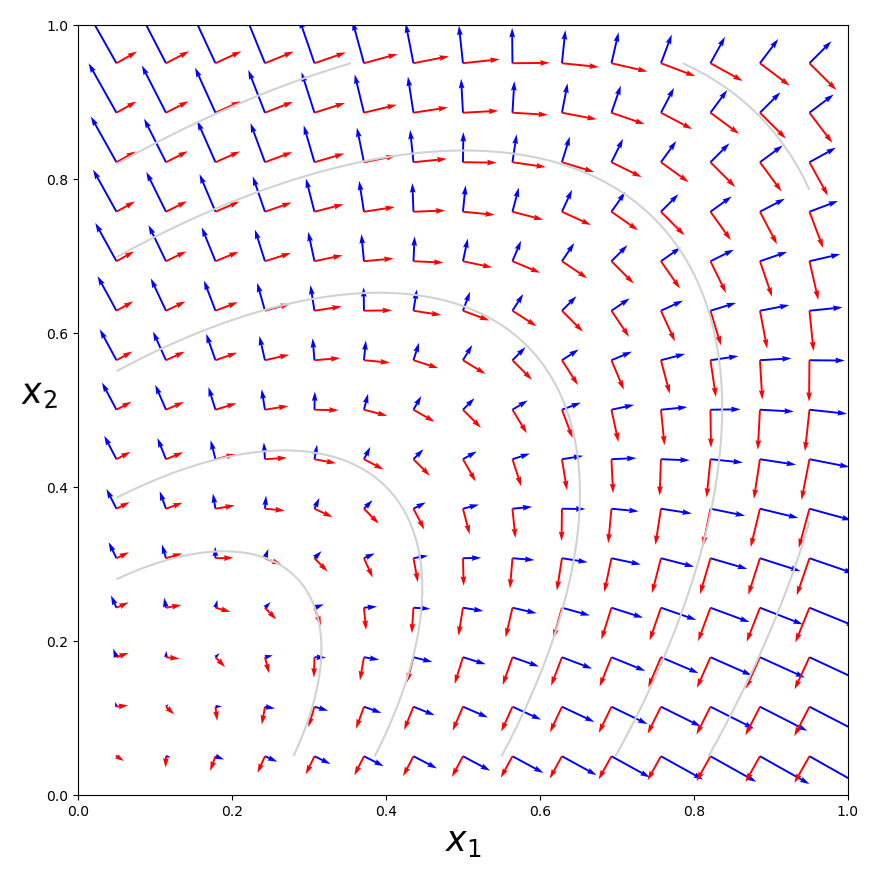}
\includegraphics[width=1.6in]{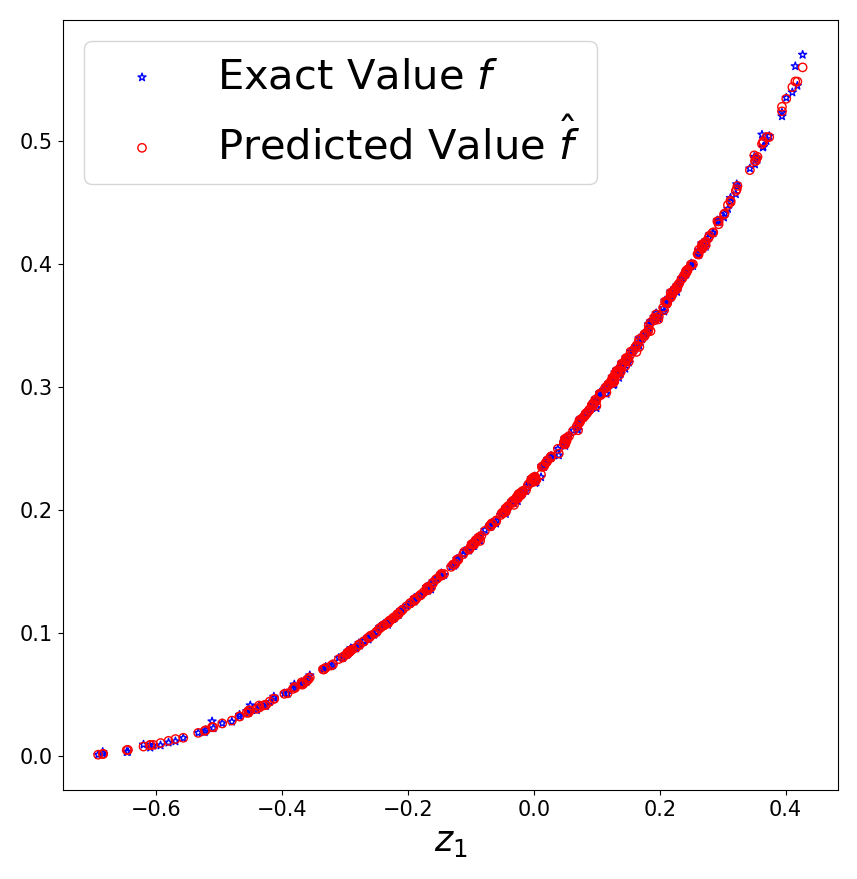}
  \texttt{RevNet} with $\alpha=50$
\includegraphics[width=1.65in]{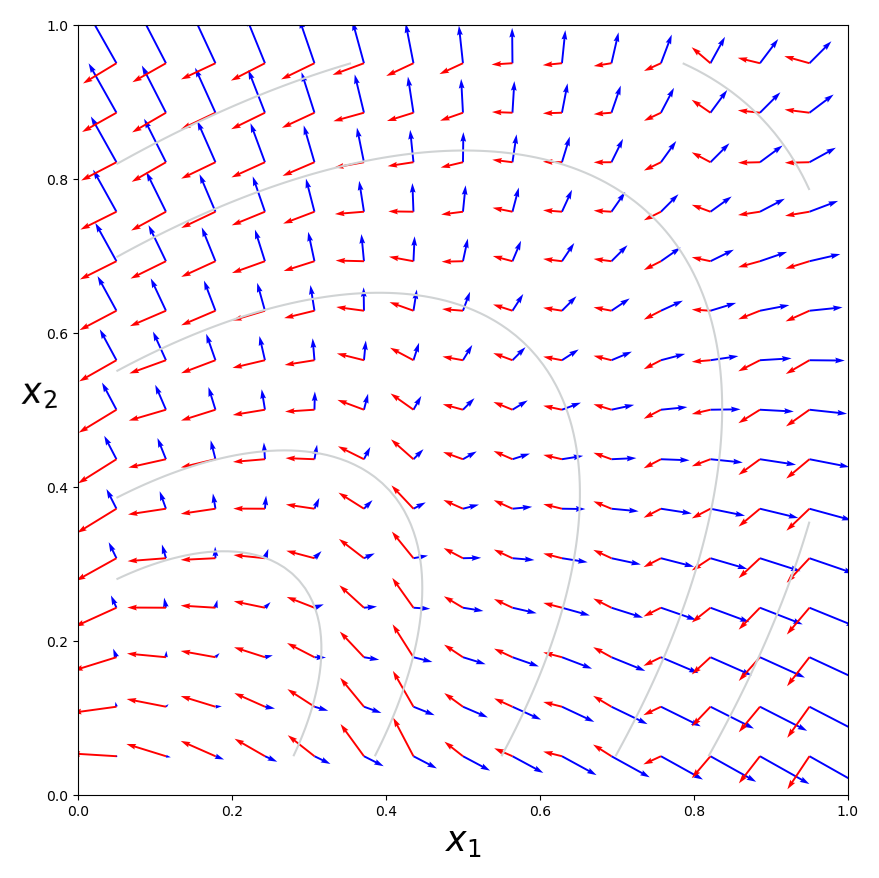}
\includegraphics[width=1.6in]{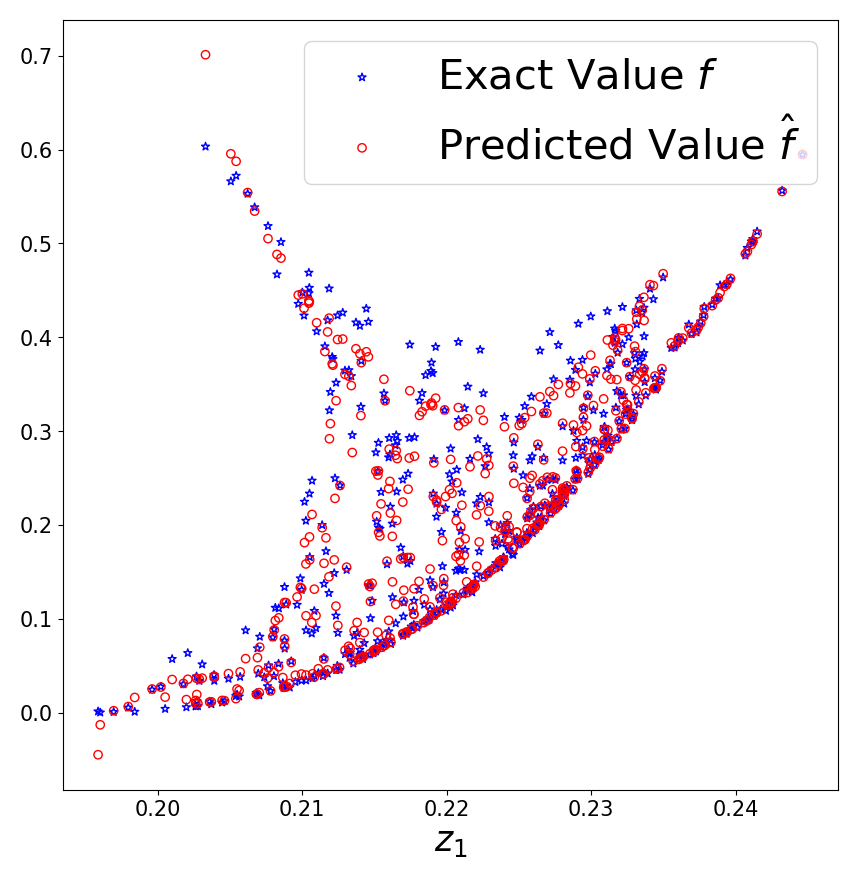}
%\caption{fig1}
\end{minipage}%
}%
    \vspace{-0.4cm}
\caption{Level set learning and function approximation results produced by our DRiLLS method with  the PRNN (the quiver plot in Row 1 and the regression plot in Row 2) or the RevNet (the quiver plot in Row 3 and the regression plot in Row 4) for $f_2(\bm{x}) = \frac{5}{8} x_1^2 + \frac{5}{8} x_2^2 - \frac{3}{4} x_1 x_2$ in $\Omega^2_{A}=[0,1]^2$, at three different values of $\alpha$ = 0, 25, 50, respectively.  PRNN successfully learns the level sets of the target function but RevNet is somehow  unable.
}
\label{f_2_01}
\end{figure}

\begin{figure}[!htbp]
\hspace{-0.3cm}
\subfigure{
\begin{minipage}[t]{0.33\linewidth}
\centering
  \texttt{PRNN} with $\alpha=0$
\includegraphics[width=1.65in]{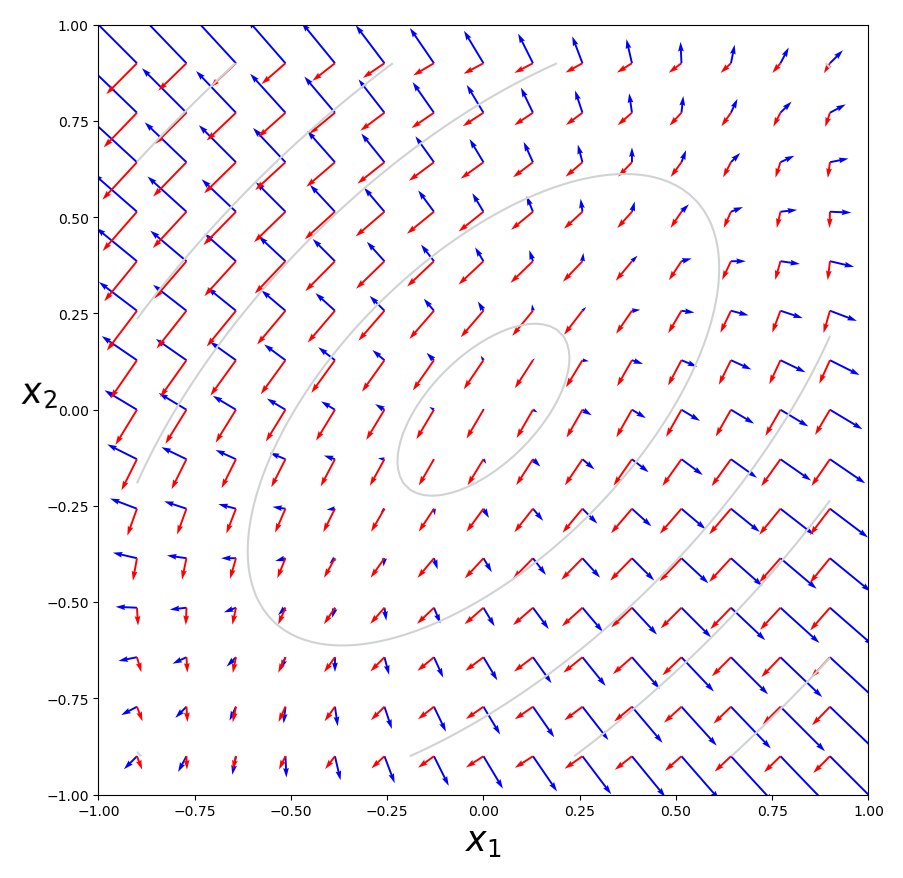}
\includegraphics[width=1.6in]{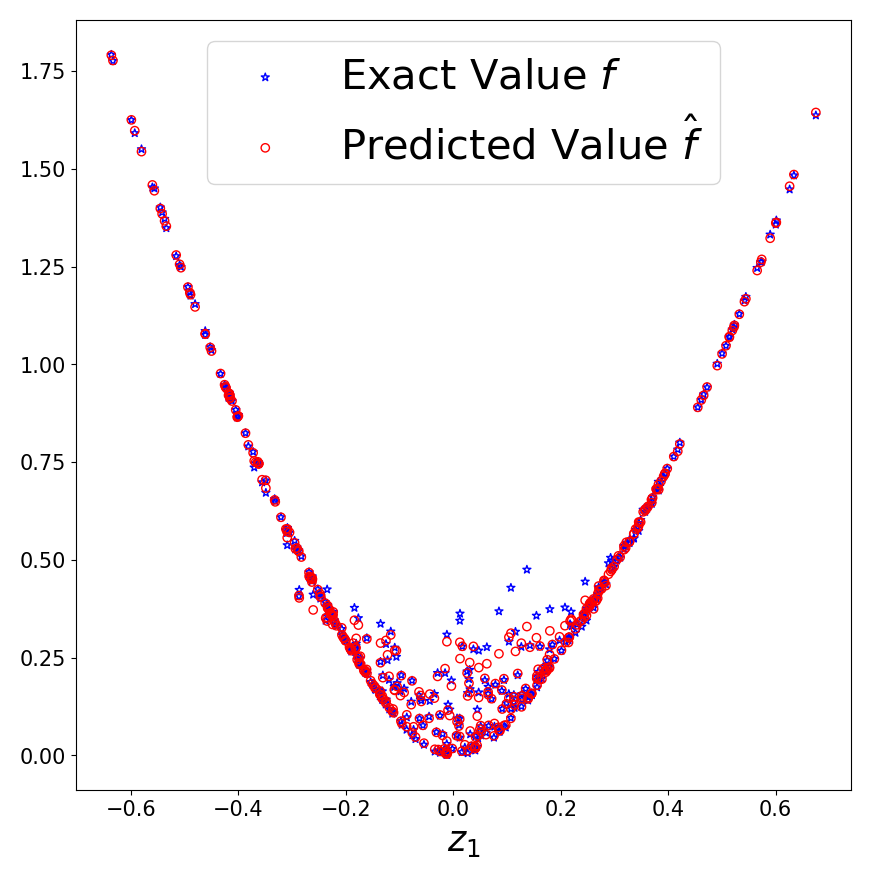}
  \texttt{RevNet} with $\alpha=0$
\includegraphics[width=1.65in]{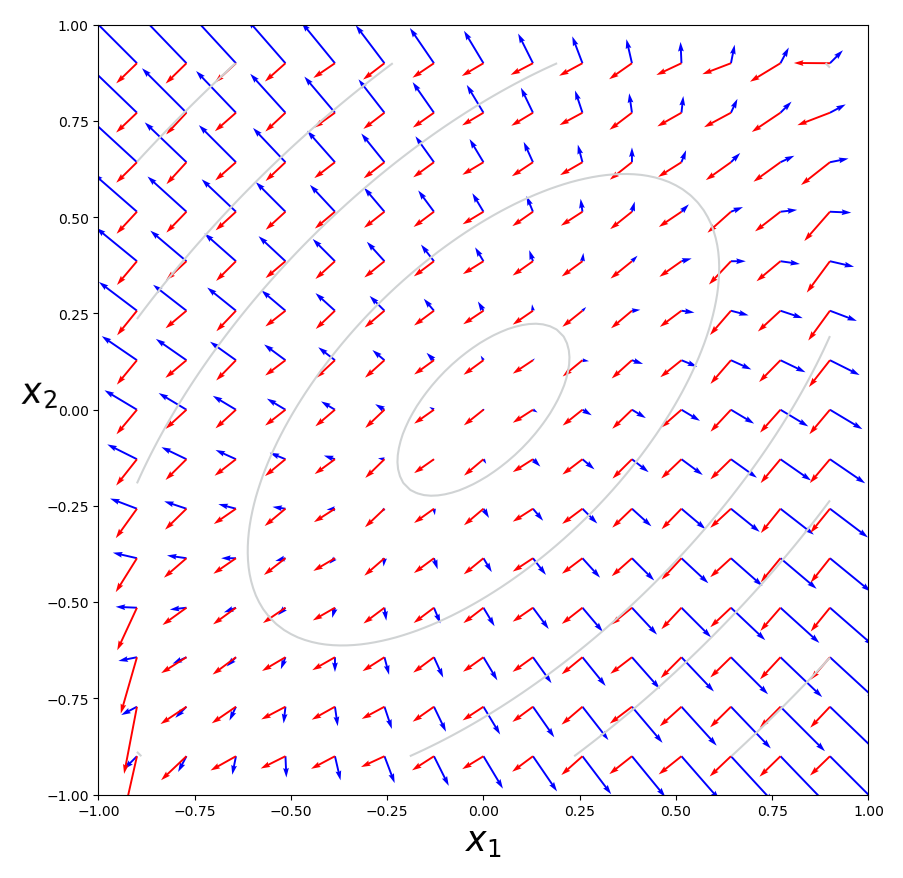}
\includegraphics[width=1.6in]{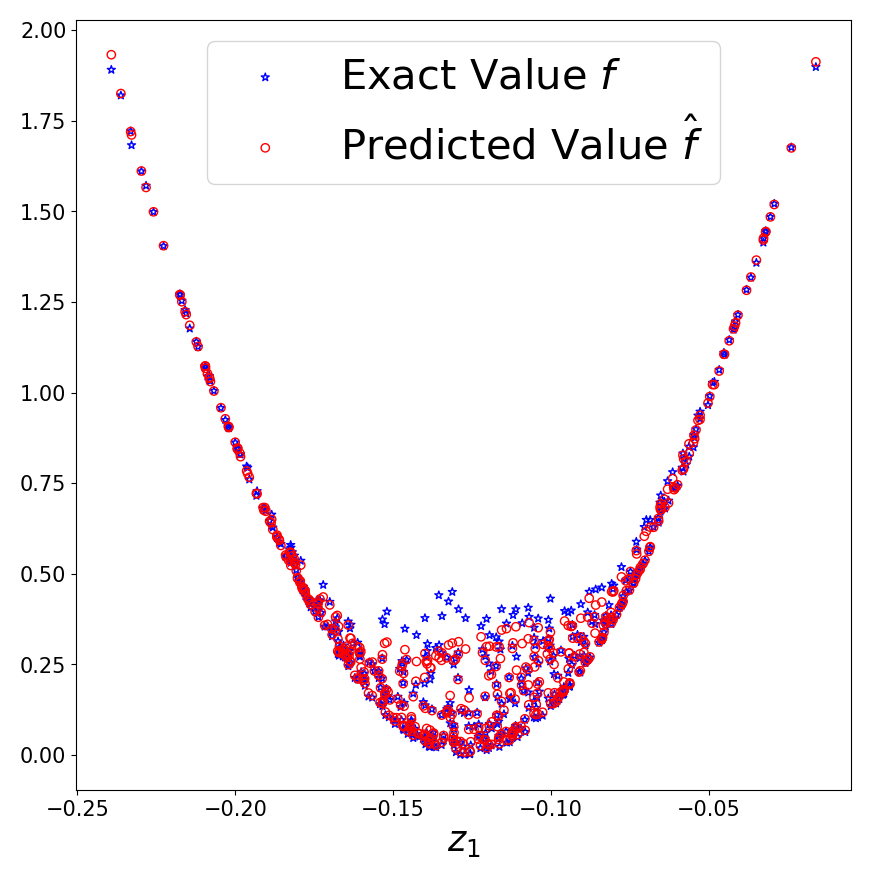}
%\caption{fig1}
\end{minipage}%
}%
\subfigure{
\begin{minipage}[t]{0.33\linewidth}
\centering
  \texttt{PRNN} with $\alpha=25$
\includegraphics[width=1.65in]{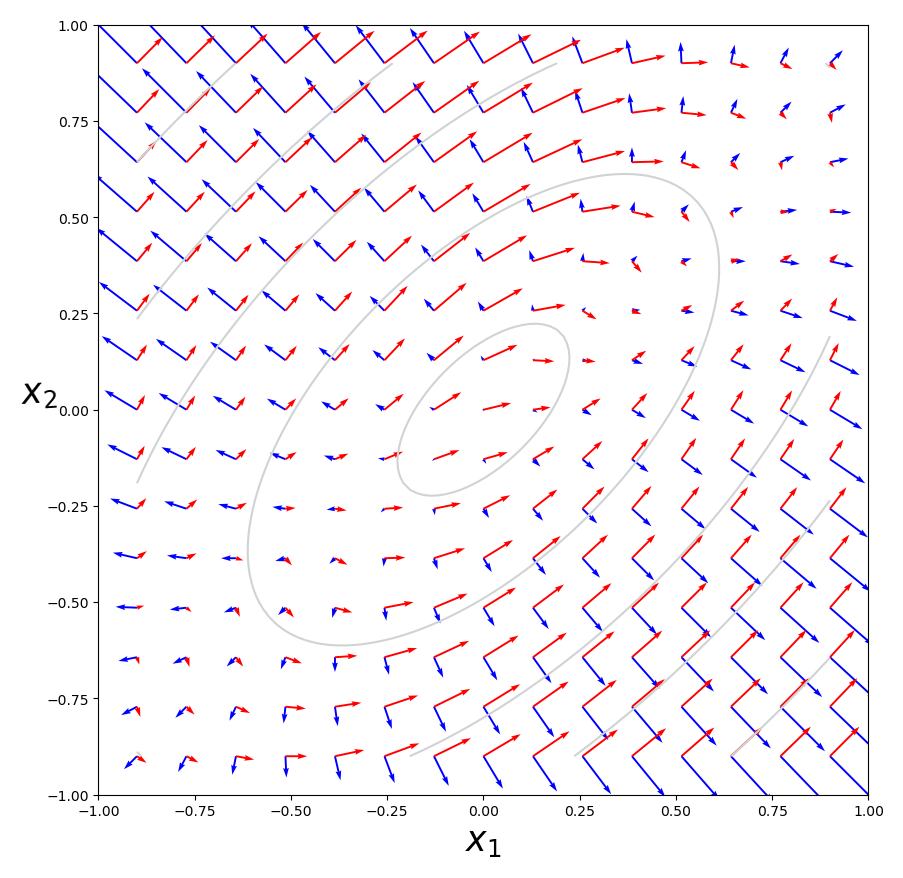}
\includegraphics[width=1.6in]{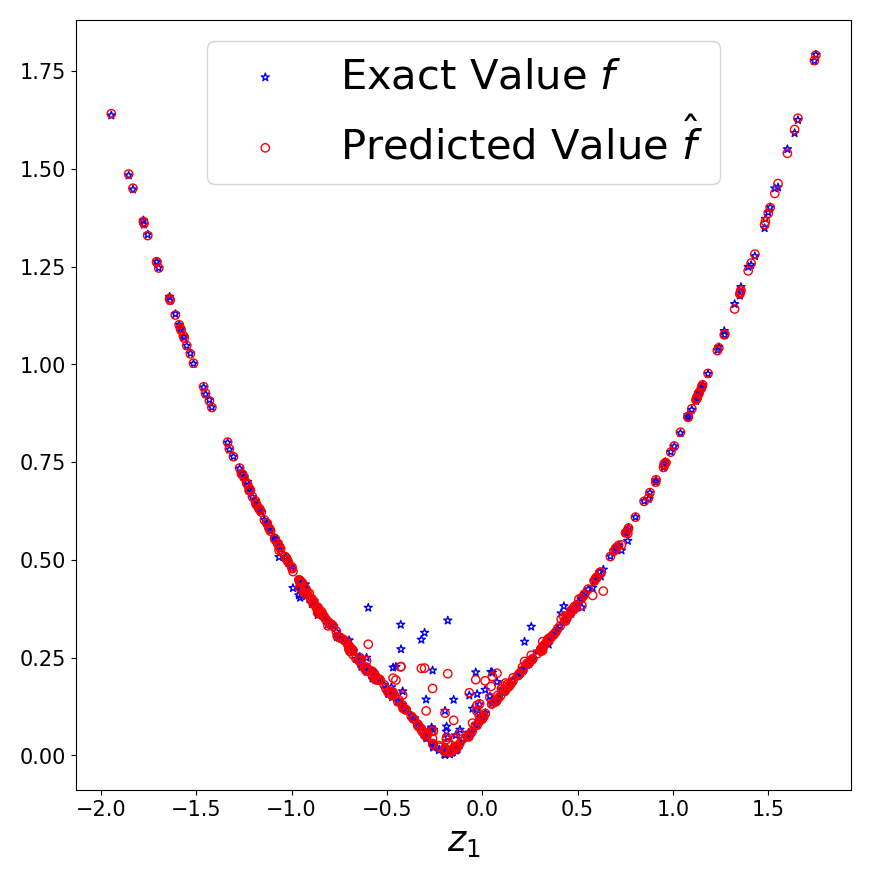}
  \texttt{RevNet} with $\alpha=25$
\includegraphics[width=1.65in]{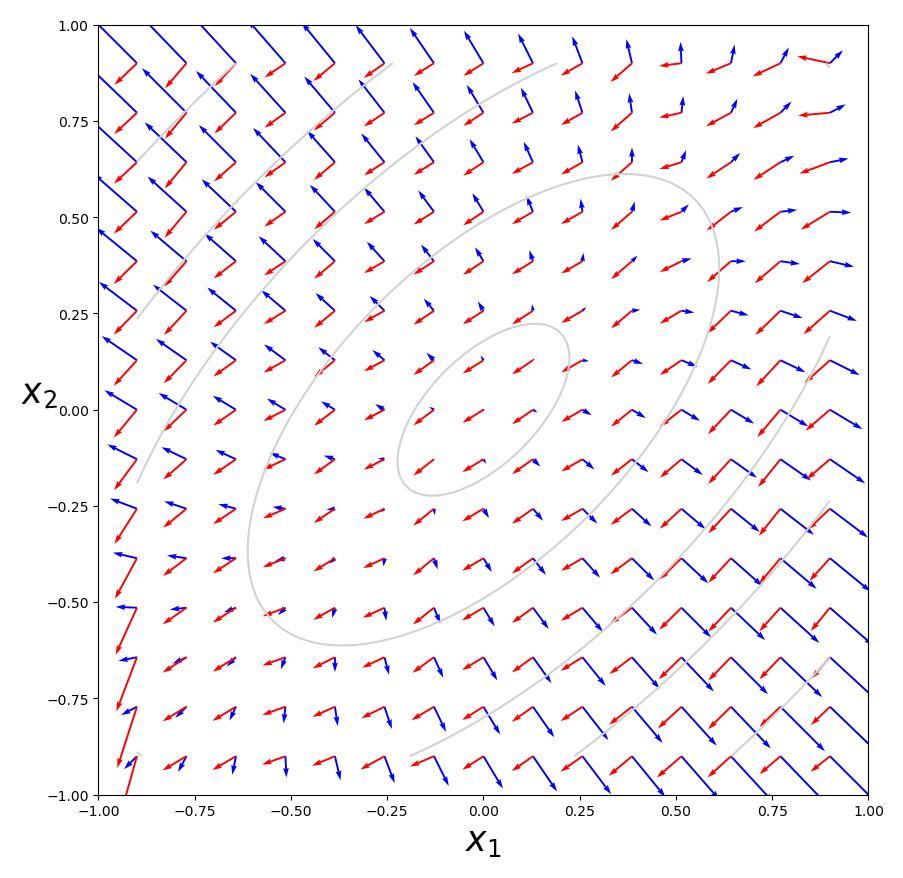}
\includegraphics[width=1.6in]{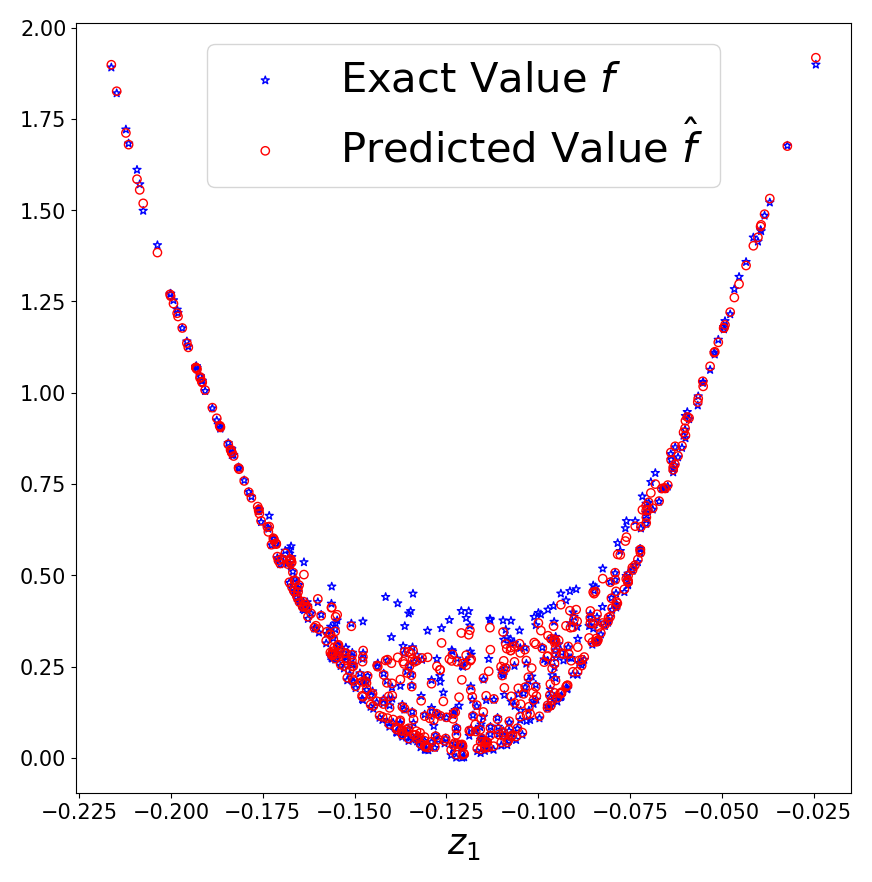}
%\caption{fig1}
\end{minipage}%
}%
\subfigure{
\begin{minipage}[t]{0.33\linewidth}
\centering
  \texttt{PRNN} with $\alpha=50$
\includegraphics[width=1.65in]{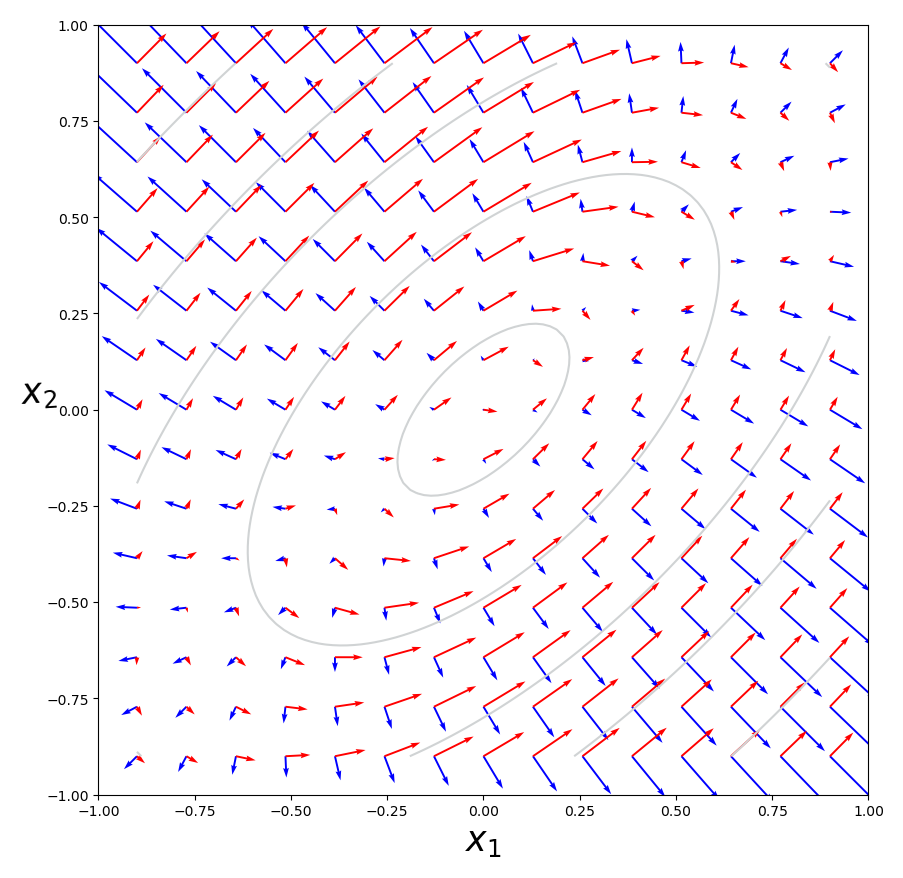}
\includegraphics[width=1.6in]{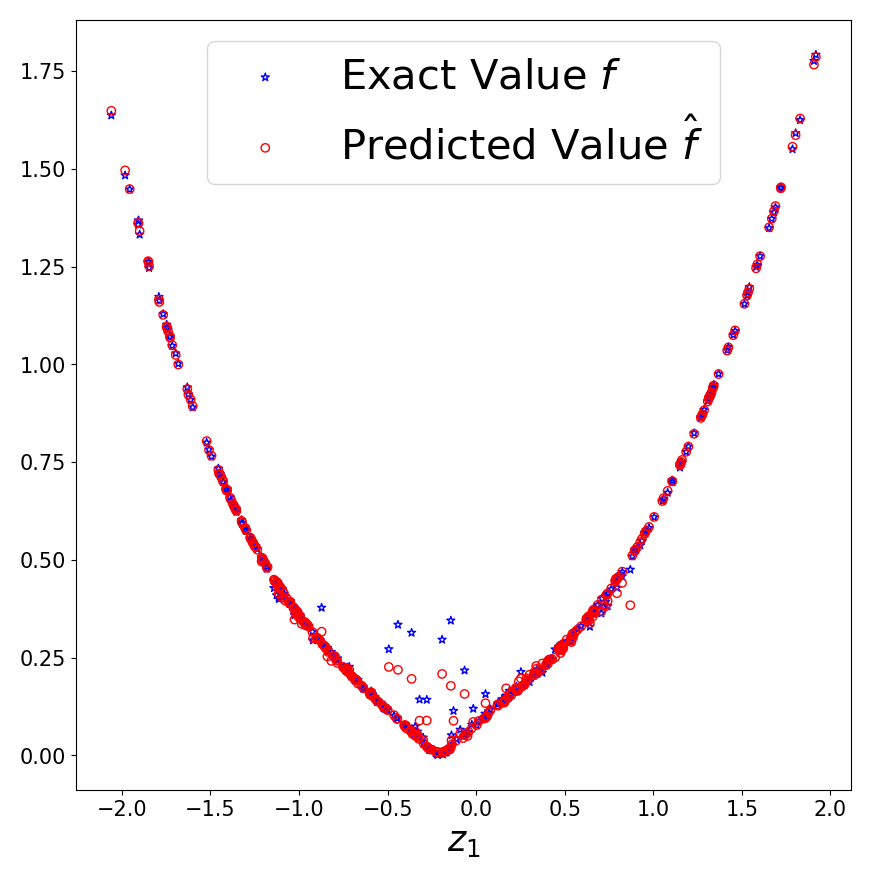}
  \texttt{RevNet} with $\alpha=50$
\includegraphics[width=1.65in]{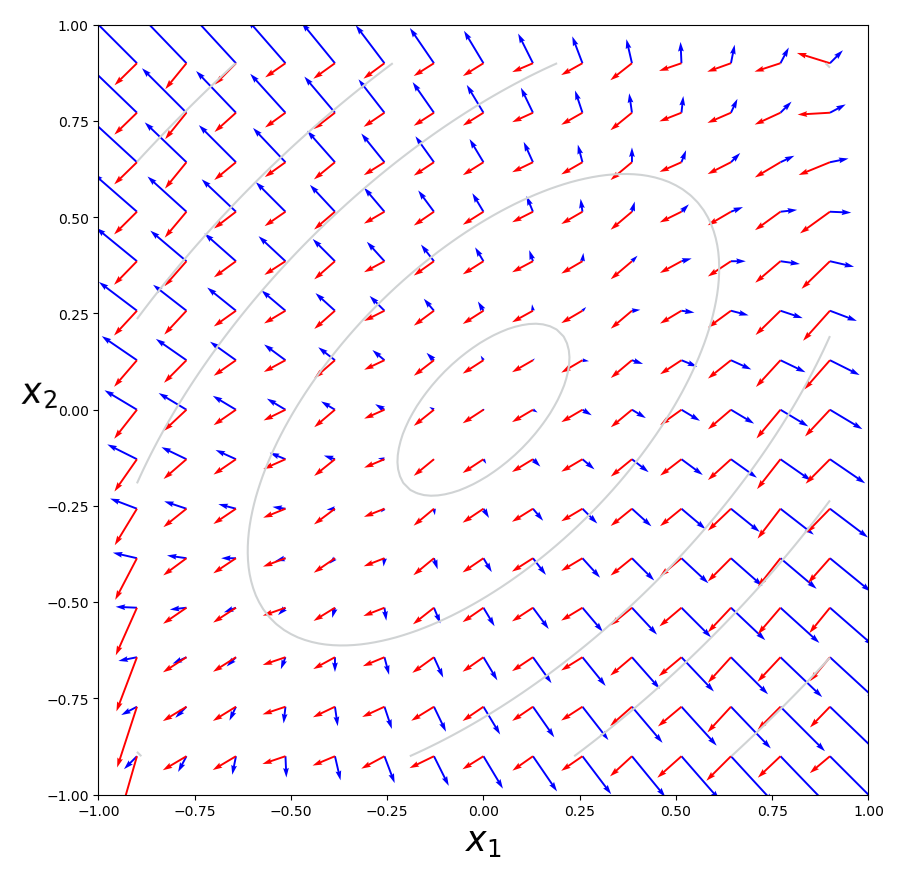}
\includegraphics[width=1.6in]{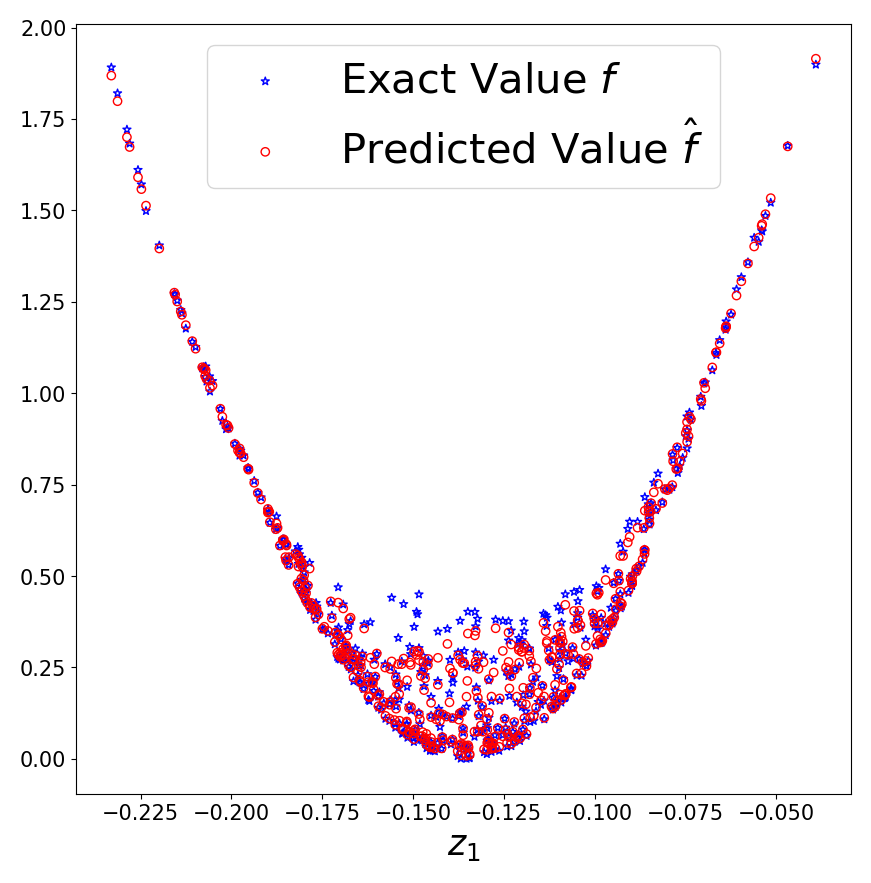}
%\caption{fig1}
\end{minipage}%
}%
    \vspace{-0.4cm}
\caption{Level set learning and function approximation results produced by our DRiLLS method with the PRNN (the quiver plot in Row 1 and the regression plot in Row 2) or the RevNet (the quiver plot in Row 3 and the regression plot in Row 4) for $f_2(\bm{x}) = \frac{5}{8} x_1^2 + \frac{5}{8} x_2^2 - \frac{3}{4} x_1 x_2$ in $\Omega^2_{B}=[-1, 1]^2$, at three different values of $\alpha$ = 0, 25, 50, respectively.
PRNN successfully learns the level sets of the target function but RevNet is somehow  unable.}
\label{f_2_11}
\end{figure}

The testing results of $f_2(\bm{x})$ are displayed in \Cref{f_2_01,f_2_11} for the function respectively defined in $\Omega^2_{A}$ and $\Omega^2_{B}$.
We remark that the behavior of $f_2$ in either $\Omega^2_{A}$  or in $\Omega^2_{B}$ is not monotonic at all.
It is observed from \Cref{f_2_01,f_2_11} that the PRNN achieves superior performances on both domains: the quiver plots indicate that the RevNet has difficulty in ensuring the function value to be insensitive to $z_2$ in both $\Omega^2_{A}$ and $\Omega^2_{B}$ cases, and the associated regression plots show that the RevNet produces a more erroneous function approximation. 
The PRNN, on the contrary, still works well on both domains, which further leads to more accurate function approximations.

Meanwhile, we also observe that the value of $\alpha$ does not have much impact on the performance of RevNet. 
For the PRNN, the effect of $\alpha$ on the performance  also seems  negligible for the case  $\bm{x}\in \Omega^2_{A}$, but 
  becomes significantly different for the case $\bm{x}\in \Omega^2_{B}$. As $\alpha$ increases from $0$ to $25$ and $50$, the learned level sets and
 the function approximations get more and more accurate, especially for $f_2$. As shown in the first rows of \Cref{f_1_11,f_2_11}, red arrows are well perpendicular to the blue arrows in the quiver plots for two larger values of $\alpha$, manifesting more effective dimension reductions.
Moreover, less blue dots are visible in the regression plots in the third column than those in the first two column, which indicates less discrepancy between the predicted values and the exact function values. 

\subsubsection{The effectiveness of the bounded derivative loss}\label{sec:bdl}
We use $f_2$ with the domain $\Omega^2_{B}$ to investigate the effect of the  bounded derivative loss $\mathcal{L}_2$ that is a new loss term compared to those used in the NLL method. 
The purpose of $\mathcal{L}_2$ is to reduce the oscillation in the function values after they are projected onto the active variable space, thus, it mainly can be regarded as a regularization term.

\begin{figure}[!htbp]
\hspace{-0.3cm}
\centering
\subfigure[Total Loss]{
\centering
\includegraphics[width=.32\textwidth]{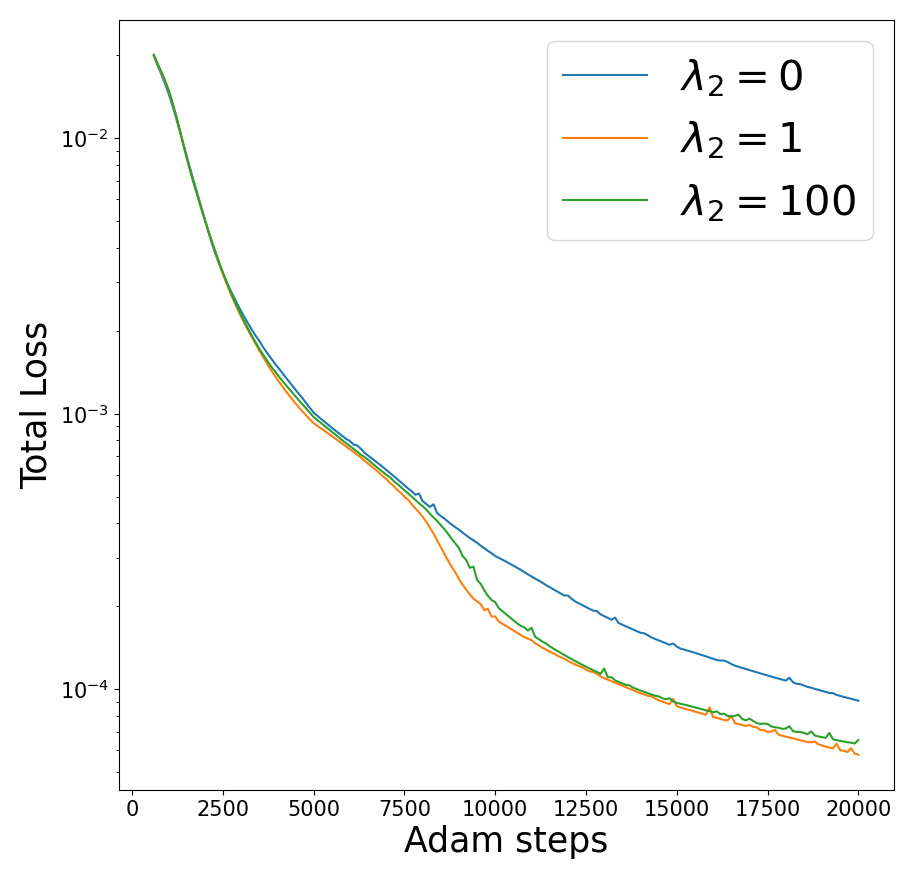}
%\caption{fig1}
}%
\subfigure[Pseudo-reversibility Loss]{
\centering
\includegraphics[width=.32\textwidth]{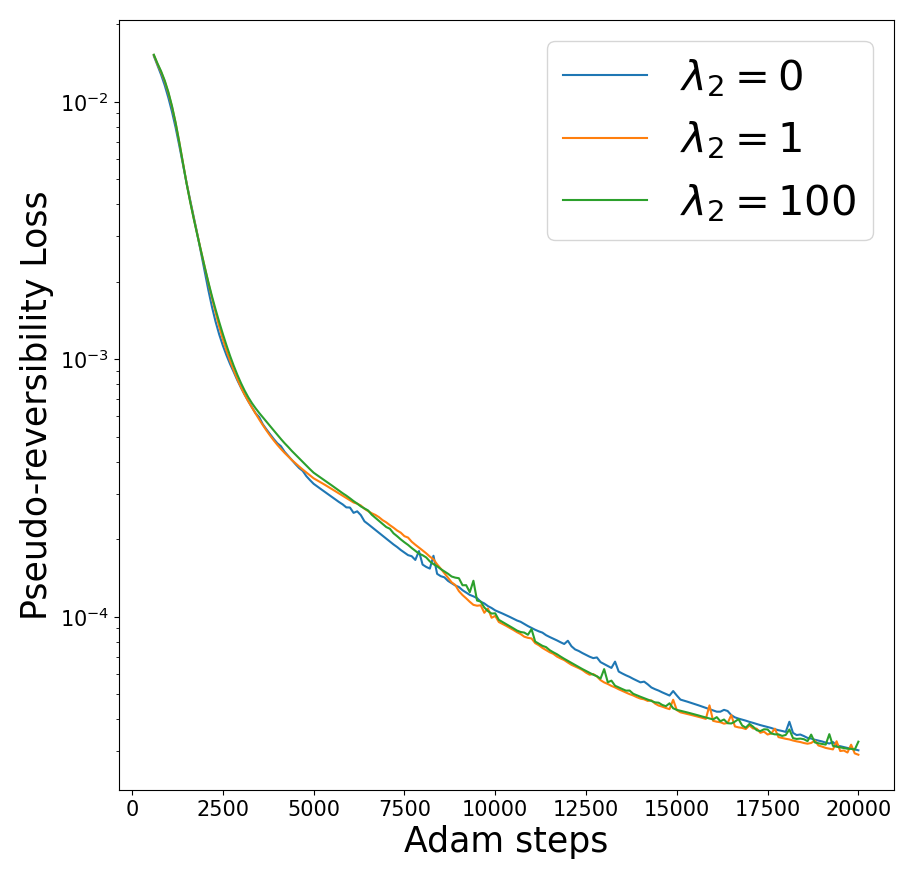}
%\caption{fig1}
}%
\subfigure[Active direction fitting Loss]{
\centering
\includegraphics[width=.32\textwidth]{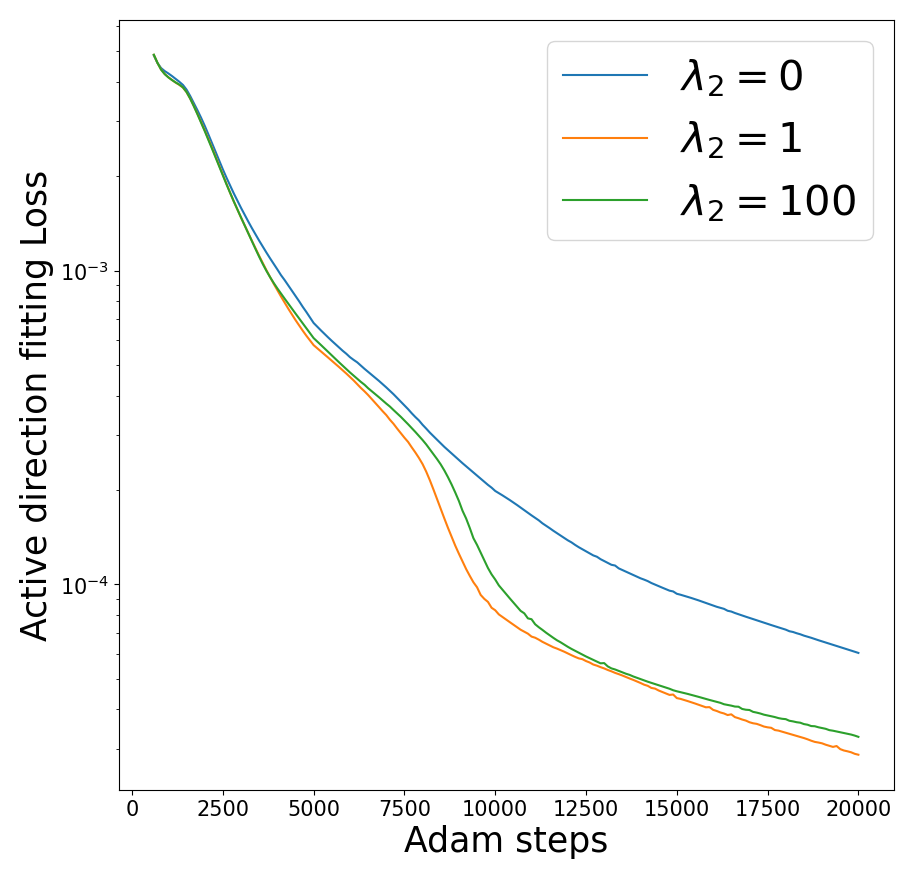}
%\caption{fig1}
}%
\centering
\caption{Evolutions of the total loss $\mathcal{L}$ (left), the pseudo-reversibility loss $\mathcal{L}_1$ (middle) and the active direction fitting loss $\mathcal{L}_2$ (right)  with three different values of $\lambda_2=0, 1, 100$ during the training process of the PRNN for  $f_2$ in $\Omega_{B}^2=[-1,1]^2$. }\label{fig:los}
\end{figure}

To check whether the proposed bounded derivative loss helps the training process of the proposed PRNN in our DRiLLS method, we vary the value of $\lambda_2$ from $0$ to $1$ and $100$
while fixing the other experimental settings.  The training dataset again has size 500.
The evolutions of the total loss $\mathcal{L}$, the pseudo-reversibility loss  $\mathcal{L}_1$, and the active direction fitting loss  $\mathcal{L}_2$ during the training process are presented  in Figure \ref{fig:los}. 
It is observed that the pseudo-reversibility loss $\mathcal{L}_1$ is not affected by the choice of $\lambda_2$, but
the total training loss  and the active direction fitting loss both decay  faster when $\lambda_2=1$ than when $\lambda_2=0$. Conversely, the even larger value $\lambda_2=100$ does not further  accelerate the training process. 

\subsubsection{The synthesized regression v.s. other regression methods}\label{sec:loc}
\begin{figure}[!htbp]
\centering
\begin{minipage}[t]{0.48\linewidth}
\includegraphics[width=2.2in]{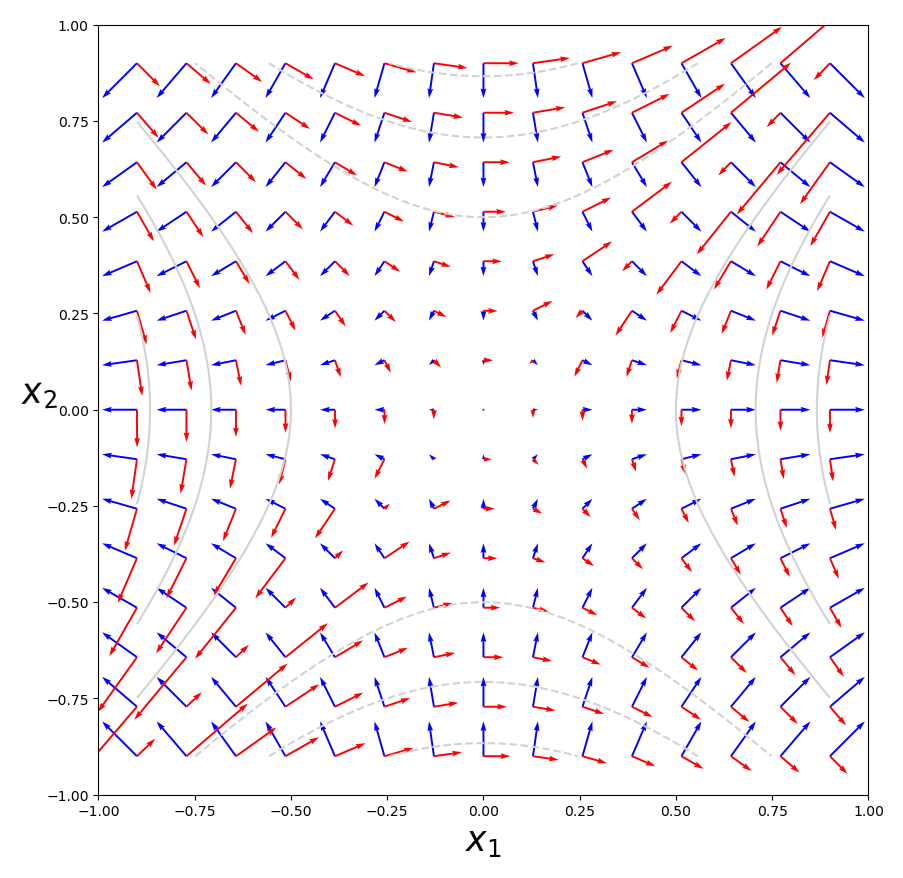}
\end{minipage}%
\begin{minipage}[t]{0.48\linewidth}
\includegraphics[width=2.15in]{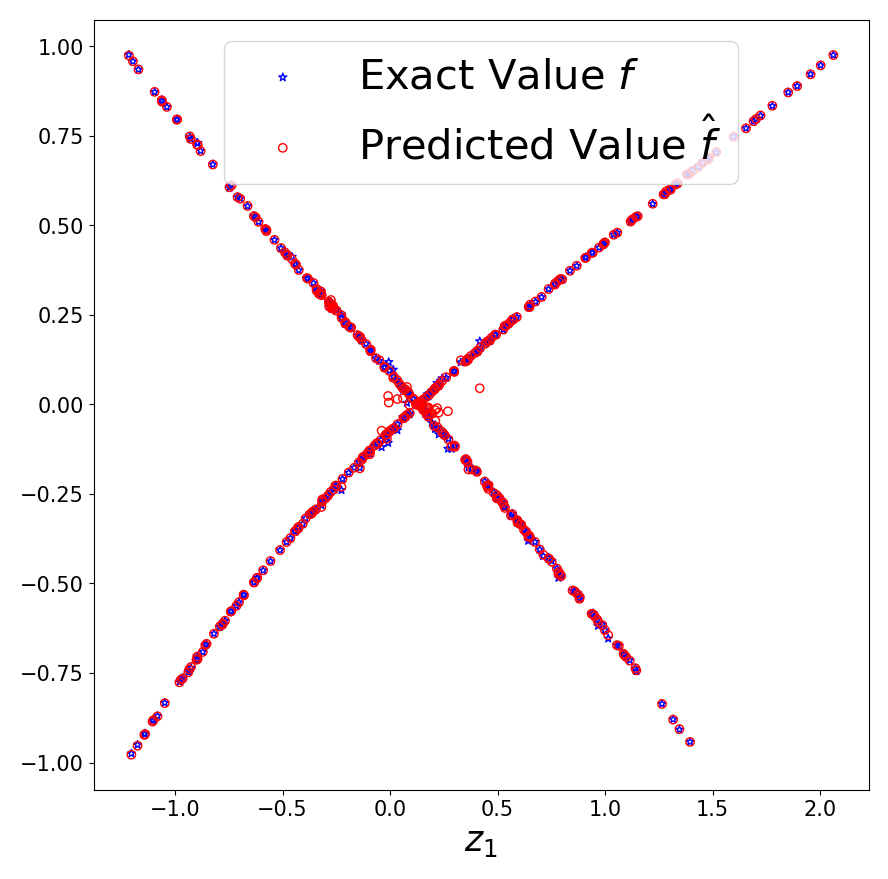}
\end{minipage}%
\vspace{-0.2cm}
\caption{Level set learning and function approximation results produced by our DRiLLS method for $f_3$ in $\Omega_{B}^2=[-1,1]^2$: the quiver plot (left)  and the regression plot (right). The synthesized regression approach successfully overcomes the numerical oscillation issue illustrated in the top row of \Cref{fig:reg}.}\label{fig:hy}
\end{figure}

Once the transformation to the active variable $\bm{z}_A$ is obtained through the PRNN, we apply the proposed synthesized regression for approximating the target function. 
To better demonstrate the advantage  of our synthesized regression, we consider the following example featured in Figure  \ref{fig:reg}:  
\begin{equation}
f_3(\bm{x}) = x_1^2-x_2^2 \quad  \text{for} \quad \bm{x} \in \Omega^2_{B}.
\end{equation}
Due to the complicated behavior of  the function $f_3$ in $[-1,1]^2$, we set the size of training dataset $N=2500$ in the PRNN and 
the associated quiver and regression plots produced by our method are presented in \Cref{fig:hy}. The former demonstrates the efficacy of PRNN dimension reduction as the derivative in the function with respect to $z_2$ is tangent to the level sets, and the latter indicates accurate  regressions have been obtained as almost all the blue stars and red circles coincide with each other, though the graph of $f$  has several branches. 

\begin{table}[!htbp]
\renewcommand{\arraystretch}{1.2}
{\footnotesize
\caption{Approximation errors by different regression methods on the same PRNN transformed data for $f_3$ in $\Omega_{B}^2=[-1,1]^2$.}\label{tab:hyper_11}
\centering
\begin{tabular}{|c|c|c|c|c|}
\hline
& Synthesized  Regression & Direct Local Fitting & Global Fitting & Neural Network \\
\hline
$\mathtt{NRMSE}$ & 0.86$\%$ & 20.49$\%$ & 20.16$\%$ & 19.93$\%$ \\
\hline
$\mathtt{RL_1}$  & 1.32$\%$ & 91.11$\%$ & 93.89$\%$ & 91.27$\%$ \\
\hline
\end{tabular}
}
\end{table}

The performance of the proposed synthesized regression  is also compared with some other popular regression methods based on the same PRNN transformed data, including the polynomial regression in local and global fashions and the nonlinear regression by neural networks. In particular,  cubic polynomial fitting is applied, and the neural network regression uses a FCNN of 3 hidden layers with 20 neurons in each layer.  The function approximation errors are summarized in \Cref{tab:hyper_11}, which shows that the direct local fitting, the global fitting, and the neural network regression all  fail to provide accurate predictions while the  synthesized regression  performs significantly  well. 

\subsection{High-dimensional function approximation with limited data}\label{sec:mor}

Here we compare our DRiLLS method with two popular dimension reduction methods, the AS and the NLL,  for function approximation with limited/sparse data. To ensure a fair comparison, the proposed synthesized regression will be applied to all compared methods for regression after active subspaces/variables are identified. In particular, 
 the dimension of the active variables  is set to $k^*=1$ and $2$  for DRiLLS and  NLL and similarly for AS,  which are often the typical choices in practice.
We consider the following four functions:
\begin{equation}
% \displaystyle
	\begin{array}{ll}
	f_4(\bm{x}) = \sum\limits_{i=1}^d x_i^2, \qquad &f_5(\bm{x}) = \sin \left(\sum\limits_{i=1}^d x_i^2\right),\vspace{2ex}\\ 
	f_6(\bm{x}) = \prod\limits_{i=1}^{d} \dfrac{1}{1 + x_i^2},\qquad  &f_7(\bm{x}) =- x_d^2+ \sum\limits_{i=1}^{d-1} x_i^2. 
	\end{array}
\end{equation}
For $f_4, f_5$ and $f_6$,  $\Omega_{A}^d=[0,1]^d$ with $d=10, 20$ and $\Omega_{B}^d=[-1,1]^d$ with $d=8, 12$ are respectively used  as the domain of the functions. For  $f_7$, we only take $\Omega_{B}^d=[-1,1]^d$ with $d=8, 12$ is used as the domain. Obviously, the behaviors of the these functions are much more complicated in $\Omega_B^d$ than in $\Omega_A^d$. 
{We observed that for all the tests reported in this section their training processes again terminated within 60000 Adam optimization steps.}

\begin{table}[!htbp]
\renewcommand{\arraystretch}{1.2}
{\footnotesize
  \caption{Numerical approximation errors produced by DRiLLS, NLL and AS for $f_4$ on various domains. }\label{tab:f_4}
\begin{center}
\begin{tabular}{|c|c|c|c|c|c|c|}
\hline
 & \multicolumn{6}{c|}{$f_4$ in $\Omega_{A}^{10}=[0,1]^{10}$}                                                           \\ \cline{2-7} 
                  & \multicolumn{2}{c|}{500}           & \multicolumn{2}{c|}{2500}          & \multicolumn{2}{c|}{10000}         \\ \cline{2-7} 
                  & $\mathtt{NRMSE}$ & $\mathtt{RL_1}$ & $\mathtt{NRMSE}$ & $\mathtt{RL_1}$ & $\mathtt{NRMSE}$ & $\mathtt{RL_1}$ \\ \hline
DRiLLS ($k^*=1$)             & 0.60$\%$         & 0.80$\%$        & 0.22$\%$         & 0.31$\%$        & 0.20$\%$         & 0.26$\%$        \\ \hline
DRiLLS ($k^*=2$)              & 0.61$\%$         & 0.82$\%$        & 0.22$\%$         & 0.31$\%$        & 0.20$\%$         & 0.26$\%$        \\ \hline
NLL ($k^*=1$)               & 0.32$\%$         & 0.43$\%$        & 0.30$\%$         & 0.43$\%$        & 0.32$\%$         & 0.45$\%$        \\ \hline
NLL ($k^*=2$)               & 0.37$\%$         & 0.50$\%$        & 0.40$\%$         & 0.56$\%$        & 0.37$\%$         & 0.51$\%$        \\ \hline
AS ($k^*=1$)                 & 3.52$\%$         & 5.51$\%$        & 3.20$\%$         & 4.88$\%$        & 2.68$\%$         & 4.19$\%$        \\ \hline
AS ($k^*=2$)                 & 3.81$\%$         & 5.84$\%$        & 3.49$\%$         & 5.22$\%$        & 2.92$\%$         & 4.39$\%$        \\ \hline\hline
 & \multicolumn{6}{c|}{$f_4$ in $\Omega_{A}^{20}=[0,1]^{20}$}                                                           \\ 
 \cline{2-7} 
                  & \multicolumn{2}{c|}{500}           & \multicolumn{2}{c|}{2500}          & \multicolumn{2}{c|}{10000}         \\ \cline{2-7} 
                  & $\mathtt{NRMSE}$ & $\mathtt{RL_1}$ & $\mathtt{NRMSE}$ & $\mathtt{RL_1}$ & $\mathtt{NRMSE}$ & $\mathtt{RL_1}$ \\ \hline\hline
DRiLLS ($k^*=1$)              & 9.74$\%$        & 11.18$\%$       & 0.57$\%$         & 0.62$\%$        & 0.39$\%$         & 0.40$\%$        \\ \hline
DRiLLS ($k^*=2$)              & 11.25$\%$         & 12.68$\%$        & 0.68$\%$         & 0.75$\%$        & 0.73$\%$         & 0.71$\%$        \\ \hline
NLL ($k^*=1$)              & 0.31$\%$         & 0.33$\%$        & 0.28$\%$         & 0.31$\%$        & 0.37$\%$         & 0.38$\%$        \\ \hline
NLL ($k^*=2$)               & 0.39$\%$         & 0.40$\%$        & 0.35$\%$         & 0.36$\%$        & 0.40$\%$         & 0.38$\%$        \\ \hline
AS ($k^*=1$)                & 3.83$\%$         & 4.52$\%$        & 3.74$\%$         & 4.35$\%$        & 3.68$\%$         & 4.24$\%$        \\ \hline
AS ($k^*=2$)                & 4.27$\%$         & 4.84$\%$        & 4.07$\%$         & 4.60$\%$        & 3.97$\%$         & 4.46$\%$        \\ \hline\hline
 & \multicolumn{6}{c|}{$f_4$ in $\Omega_{B}^{8}=[-1,1]^{8}$}                                             \\ \cline{2-7} 
                  & \multicolumn{2}{c|}{2500}           & \multicolumn{2}{c|}{10000}          & \multicolumn{2}{c|}{40000}         \\ \cline{2-7} 
                  & $\mathtt{NRMSE}$ & $\mathtt{RL_1}$ & $\mathtt{NRMSE}$ & $\mathtt{RL_1}$ & $\mathtt{NRMSE}$ & $\mathtt{RL_1}$ \\ \hline
DRiLLS ($k^*=1$)              & 4.26$\%$        & 3.19$\%$       & 2.25$\%$         & 1.51$\%$        & 1.53$\%$         & 1.17$\%$        \\ \hline
DRiLLS ($k^*=2$)              & 2.63$\%$         & 2.52$\%$        & 1.55$\%$         & 1.39$\%$        & 1.05$\%$         & 0.87$\%$        \\ \hline
AS ($k^*=1$)                & 11.42$\%$        & 19.14$\%$       & 9.73$\%$        & 15.49$\%$       & 7.53$\%$        & 12.14$\%$       \\ \hline
AS ($k^*=2$)                & 12.66$\%$             & 20.13$\%$            & 9.96$\%$             & 16.25$\%$            & 8.04$\%$             & 13.62$\%$            \\ \hline\hline
& \multicolumn{6}{c|}{$f_4$ in $\Omega_{B}^{12}=[-1,1]^{12}$}                                                     \\ \cline{2-7} 
                  & \multicolumn{2}{c|}{2500}           & \multicolumn{2}{c|}{10000}          & \multicolumn{2}{c|}{40000}         \\ \cline{2-7} 
                  & $\mathtt{NRMSE}$ & $\mathtt{RL_1}$ & $\mathtt{NRMSE}$ & $\mathtt{RL_1}$ & $\mathtt{NRMSE}$ & $\mathtt{RL_1}$ \\ \hline
DRiLLS ($k^*=1$)              & 13.71$\%$        & 19.66$\%$       & 3.59$\%$        & 2.40$\%$       & 2.22$\%$         & 1.59$\%$        \\ \hline
DRiLLS ($k^*=2$)              & 13.88$\%$         & 19.69$\%$        & 2.93$\%$         & 2.47$\%$        & 1.86$\%$         & 1.80$\%$        \\ \hline
AS ($k^*=1$)                & 12.96$\%$             & 19.38$\%$            & 12.17$\%$             & 17.45$\%$            & 10.98$\%$             & 15.22$\%$            \\ \hline
AS ($k^*=2$)                & 14.35$\%$             & 20.12$\%$            & 12.70$\%$             & 18.14$\%$            & 11.55$\%$             & 15.96$\%$            \\ \hline

\end{tabular}
\end{center}
}
\end{table}

\begin{table}[!htbp]
{\footnotesize
  \caption{Numerical approximation errors produced by DRiLLS, NLL and AS  for $f_5$ on various domains.}\label{tab:f_5}
\begin{center}
\renewcommand{\arraystretch}{1.2}
\begin{tabular}{|c|c|c|c|c|c|c|}
\hline
& \multicolumn{6}{c|}{$f_5$ in $\Omega_{A}^{10}=[0,1]^{10}$}                                                     \\ \cline{2-7} 
                  & \multicolumn{2}{c|}{500}           & \multicolumn{2}{c|}{2500}          & \multicolumn{2}{c|}{10000}         \\ \cline{2-7} 
                  & $\mathtt{NRMSE}$ & $\mathtt{RL_1}$ & $\mathtt{NRMSE}$ & $\mathtt{RL_1}$ & $\mathtt{NRMSE}$ & $\mathtt{RL_1}$ \\ \hline
DRiLLS ($k^*=1$)              & 1.54$\%$         & 3.56$\%$        & 0.71$\%$         & 1.63$\%$        & 0.51$\%$         & 1.18$\%$        \\ \hline
DRiLLS ($k^*=2$)              & 1.58$\%$         & 3.68$\%$        & 0.73$\%$         & 1.60$\%$        & 0.54$\%$         & 1.21$\%$        \\ \hline
NLL ($k^*=1$)               & 1.19$\%$         & 2.27$\%$        & 1.08$\%$         & 2.52$\%$        & 1.13$\%$         & 2.76$\%$        \\ \hline
NLL ($k^*=2$)               & 1.42$\%$         & 2.56$\%$        & 1.03$\%$         & 2.33$\%$        & 0.84$\%$         & 1.95$\%$        \\ \hline
AS ($k^*=1$)                & 8.95$\%$        & 22.95$\%$       & 8.07$\%$         & 20.66$\%$       & 6.91$\%$         & 17.55$\%$       \\ \hline
AS ($k^*=2$)                & 10.02$\%$        & 24.97$\%$       & 8.58$\%$         & 21.80$\%$       & 7.23$\%$         & 18.38$\%$       \\ \hline\hline
 & \multicolumn{6}{c|}{$f_5$ in $\Omega_{A}^{20}=[0,1]^{20}$}                                                     \\ \cline{2-7} 
                  & \multicolumn{2}{c|}{500}           & \multicolumn{2}{c|}{2500}          & \multicolumn{2}{c|}{10000}         \\ \cline{2-7} 
                  & $\mathtt{NRMSE}$ & $\mathtt{RL_1}$ & $\mathtt{NRMSE}$ & $\mathtt{RL_1}$ & $\mathtt{NRMSE}$ & $\mathtt{RL_1}$ \\ \hline
DRiLLS ($k^*=1$)              & 28.73$\%$        & 79.29$\%$       & 3.75$\%$         & 7.72$\%$        & 2.81$\%$         & 5.73$\%$        \\ \hline
DRiLLS ($k^*=2$)              & 33.28$\%$         & 87.72$\%$        & 4.06$\%$         & 7.72$\%$        & 3.08$\%$         & 5.82$\%$        \\ \hline
NLL ($k^*=1$)                & 5.09$\%$         & 7.46$\%$        & 3.48$\%$         & 5.78$\%$       & 2.58$\%$         & 4.92$\%$       \\ \hline
NLL ($k^*=2$)                & 5.92$\%$         & 8.13$\%$        & 4.17$\%$         & 6.22$\%$       &  3.07$\%$         & 5.02$\%$       \\ \hline
AS ($k^*=1$)                & 13.89$\%$        & 32.84$\%$       & 12.96$\%$        & 30.99$\%$       & 12.60$\%$        & 29.90$\%$       \\ \hline
AS ($k^*=2$)                & 15.54$\%$        & 35.73$\%$       & 14.04$\%$        & 32.87$\%$       & 13.42$\%$        & 31.44$\%$       \\ \hline\hline

 & \multicolumn{6}{c|}{$f_5$ in $\Omega_{B}^{8}=[-1,1]^{8}$}                                             \\ \cline{2-7} 
                  & \multicolumn{2}{c|}{2500}           & \multicolumn{2}{c|}{10000}          & \multicolumn{2}{c|}{40000}         \\ \cline{2-7} 
                  & $\mathtt{NRMSE}$ & $\mathtt{RL_1}$ & $\mathtt{NRMSE}$ & $\mathtt{RL_1}$ & $\mathtt{NRMSE}$ & $\mathtt{RL_1}$ \\ \hline
DRiLLS ($k^*=1$)              & 9.18$\%$        & 15.90$\%$       & 6.34$\%$         & 9.96$\%$        & 4.56$\%$         & 6.66$\%$        \\ \hline
DRiLLS ($k^*=2$)              & 8.06$\%$         & 13.56$\%$        & 5.10$\%$         & 8.19$\%$        & 3.11$\%$         & 4.90$\%$        \\ \hline
AS ($k^*=1$)                & 25.21$\%$        & 62.51$\%$       & 21.29$\%$        & 52.42$\%$       & 17.29$\%$        & 42.63$\%$       \\ \hline
AS ($k^*=2$)                & 26.61$\%$             & 64.48$\%$            & 22.42$\%$             & 54.13$\%$            & 18.31$\%$             & 44.15$\%$            \\ \hline\hline
 & \multicolumn{6}{c|}{$f_5$ in $\Omega_{B}^{12}=[-1,1]^{12}$}                                                    \\ \cline{2-7} 
                  & \multicolumn{2}{c|}{2500}           & \multicolumn{2}{c|}{10000}          & \multicolumn{2}{c|}{40000}         \\ \cline{2-7} 
                  & $\mathtt{NRMSE}$ & $\mathtt{RL_1}$ & $\mathtt{NRMSE}$ & $\mathtt{RL_1}$ & $\mathtt{NRMSE}$ & $\mathtt{RL_1}$ \\ \hline\hline
DRiLLS ($k^*=1$)              & 26.86$\%$        & 74.17$\%$       & 17.17$\%$        & 30.54$\%$       & 11.95$\%$        & 20.81$\%$       \\ \hline
DRiLLS ($k^*=2$)              & 22.54$\%$         & 49.60$\%$        & 15.55$\%$         & 27.14$\%$        & 9.62$\%$         & 15.90$\%$        \\ \hline
AS ($k^*=1$)                & 26.82$\%$             & 73.84$\%$            & 24.58$\%$             & 67.50$\%$            & 22.15$\%$             & 60.43$\%$            \\ \hline
AS ($k^*=2$)                & 28.66$\%$             & 76.43$\%$            & 26.76$\%$             & 70.45$\%$            & 23.78$\%$             & 62.81$\%$            \\ \hline
 \end{tabular}
\end{center}
}
\end{table}

\begin{table}[!htbp]
{\footnotesize
  \caption{Numerical approximation errors produced by DRiLLS, NLL and AS  for $f_6$ on various domains.}\label{tab:f_6}
\begin{center}
\renewcommand{\arraystretch}{1.2}
\begin{tabular}{|c|c|c|c|c|c|c|}
\hline
& \multicolumn{6}{c|}{$f_6$ in $\Omega_{A}^{10}=[0,1]^{10}$}                                                     \\ \cline{2-7} 
                  & \multicolumn{2}{c|}{500}           & \multicolumn{2}{c|}{2500}          & \multicolumn{2}{c|}{10000}         \\ \cline{2-7} 
                  & $\mathtt{NRMSE}$ & $\mathtt{RL_1}$ & $\mathtt{NRMSE}$ & $\mathtt{RL_1}$ & $\mathtt{NRMSE}$ & $\mathtt{RL_1}$ \\ \hline
DRiLLS ($k^*=1$)              & 0.32$\%$         & 1.44$\%$        & 0.12$\%$         & 0.39$\%$        & 0.09$\%$         & 0.27$\%$        \\ \hline
DRiLLS ($k^*=2$)              & 0.29$\%$         & 1.31$\%$        & 0.11$\%$         & 0.37$\%$        & 0.09$\%$         & 0.27$\%$        \\ \hline
NLL ($k^*=1$)               & 0.73$\%$         & 3.29$\%$        & 0.59$\%$         & 2.42$\%$        & 0.32$\%$         & 1.41$\%$        \\ \hline
NLL ($k^*=2$)               & 0.95$\%$         & 2.76$\%$        & 0.51$\%$         & 2.12$\%$        & 0.57$\%$         & 2.31$\%$        \\ \hline
AS ($k^*=1$)                & 2.19$\%$         & 9.91$\%$       & 2.03$\%$         & 8.73$\%$        & 1.76$\%$         & 7.58$\%$        \\ \hline
AS ($k^*=2$)                & 2.58$\%$         & 10.73$\%$       & 2.09$\%$         & 9.30$\%$        & 1.97$\%$         & 7.99$\%$        \\ \hline\hline
 & \multicolumn{6}{c|}{$f_6$ in $\Omega_{A}^{20}=[0,1]^{20}$}                                                     \\ \cline{2-7} 
                  & \multicolumn{2}{c|}{500}           & \multicolumn{2}{c|}{2500}          & \multicolumn{2}{c|}{10000}         \\ \cline{2-7} 
                  & $\mathtt{NRMSE}$ & $\mathtt{RL_1}$ & $\mathtt{NRMSE}$ & $\mathtt{RL_1}$ & $\mathtt{NRMSE}$ & $\mathtt{RL_1}$ \\ \hline
DRiLLS ($k^*=1$)              & 1.28$\%$         & 13.35$\%$       & 0.27$\%$         & 2.48$\%$        & 0.14$\%$         & 1.27$\%$        \\ \hline
DRiLLS ($k^*=2$)              & 1.34$\%$         & 13.63$\%$        & 0.23$\%$         & 1.97$\%$        & 0.14$\%$         & 1.23$\%$        \\ \hline
NLL ($k^*=1$)             & 0.67$\%$         & 5.11$\%$        & 1.76$\%$         & 16.86$\%$   & 0.96$\%$         & 8.23$\%$        \\ \hline
NLL ($k^*=2$)               & 0.85$\%$         & 6.88$\%$        & 0.46$\%$         & 4.20$\%$        & 0.54$\%$         & 4.83$\%$        \\ \hline
AS ($k^*=1$)               & 1.98$\%$         & 17.25$\%$       & 1.55$\%$         & 15.93$\%$       & 1.96$\%$         & 15.73$\%$       \\ \hline
AS ($k^*=2$)               & 2.37$\%$         & 19.84$\%$       & 1.90$\%$         & 17.48$\%$       & 1.99$\%$         & 16.93$\%$       \\ \hline\hline

 & \multicolumn{6}{c|}{$f_6$ in $\Omega_{B}^{8}=[-1,1]^{8}$}                                             \\ \cline{2-7} 
                  & \multicolumn{2}{c|}{2500}           & \multicolumn{2}{c|}{10000}          & \multicolumn{2}{c|}{40000}         \\ \cline{2-7} 
                  & $\mathtt{NRMSE}$ & $\mathtt{RL_1}$ & $\mathtt{NRMSE}$ & $\mathtt{RL_1}$ & $\mathtt{NRMSE}$ & $\mathtt{RL_1}$ \\ \hline
DRiLLS ($k^*=1$)              & 4.31$\%$        & 11.76$\%$       & 2.56$\%$         & 6.36$\%$        & 1.74$\%$         & 3.95$\%$        \\ \hline
DRiLLS ($k^*=2$)              & 3.38$\%$         & 8.15$\%$        & 2.09$\%$         & 5.05$\%$        & 1.06$\%$         & 2.66$\%$        \\ \hline
AS ($k^*=1$)               & 8.43$\%$        & 36.86$\%$       & 6.43$\%$        & 27.92$\%$       & 4.94$\%$        & 20.31$\%$       \\ \hline
AS ($k^*=2$)                & 8.96$\%$             & 38.55$\%$            & 6.90$\%$             & 29.48$\%$            & 5.17$\%$             & 21.98$\%$            \\ \hline\hline
& \multicolumn{6}{c|}{$f_6$ in $\Omega_{B}^{12}=[-1,1]^{12}$}                                                    \\ \cline{2-7} 
                  & \multicolumn{2}{c|}{2500}           & \multicolumn{2}{c|}{10000}          & \multicolumn{2}{c|}{40000}         \\ \cline{2-7} 
                  & $\mathtt{NRMSE}$ & $\mathtt{RL_1}$ & $\mathtt{NRMSE}$ & $\mathtt{RL_1}$ & $\mathtt{NRMSE}$ & $\mathtt{RL_1}$ \\ \hline
DRiLLS ($k^*=1$)              & 8.70$\%$        & 61.31$\%$      & 3.27$\%$        & 17.78$\%$       & 2.38$\%$        & 12.27$\%$       \\ \hline
DRiLLS ($k^*=2$)              & 9.82$\%$         & 64.64$\%$        & 2.80$\%$         & 14.60$\%$        & 2.02$\%$         & 10.16$\%$        \\ \hline
AS ($k^*=1$)               & 8.04$\%$             & 62.19$\%$            & 7.01$\%$             & 53.50$\%$            & 6.11$\%$             & 44.14$\%$            \\ \hline
AS ($k^*=2$)                & 10.13$\%$             & 67.74$\%$            & 8.34$\%$             & 56.40$\%$            & 6.61$\%$             & 46.68$\%$            \\ \hline

 \end{tabular}\end{center}
}
\end{table}

\begin{table}[!htbp]
{\footnotesize
  \caption{Numerical approximation errors produced by DRiLLS, NLL and AS  for $f_7$ on various domains.}\label{tab:f_7}
\begin{center}
\renewcommand{\arraystretch}{1.2}
\begin{tabular}{|c|c|c|c|c|c|c|}
\hline
 & \multicolumn{6}{c|}{$f_7$ in $\Omega_{B}^{8}=[-1,1]^{8}$}                                             \\ \cline{2-7} 
                  & \multicolumn{2}{c|}{2500}           & \multicolumn{2}{c|}{10000}          & \multicolumn{2}{c|}{40000}         \\ \cline{2-7} 
                  & $\mathtt{NRMSE}$ & $\mathtt{RL_1}$ & $\mathtt{NRMSE}$ & $\mathtt{RL_1}$ & $\mathtt{NRMSE}$ & $\mathtt{RL_1}$ \\ \hline
DRiLLS ($k^*=1$)              & 3.30$\%$        & 5.26$\%$       & 2.14$\%$         & 3.28$\%$        & 1.73$\%$         & 2.36$\%$        \\ \hline
DRiLLS ($k^*=2$)              & 3.25$\%$         & 3.87$\%$        & 1.63$\%$         & 1.94$\%$        & 1.13$\%$         & 1.34$\%$        \\ \hline
%NLL(10blocks)           & $\%$        & $\%$       & $\%$        & $\%$       & $\%$        & $\%$       \\ \hline
%NLL(20blocks)           & $\%$       & $\%$       & $\%$       & $\%$       & $\%$        & $\%$       \\ \hline
AS ($k^*=1$)                & 10.83$\%$        & 24.73$\%$       & 8.73$\%$        & 19.55$\%$       & 6.94$\%$        & 15.51$\%$       \\ \hline
AS ($k^*=2$)                & 11.84$\%$             & 26.12$\%$            & 9.61$\%$             & 21.26$\%$            & 7.52$\%$             & 16.64$\%$            \\ \hline\hline
& \multicolumn{6}{c|}{$f_7$ in $\Omega_{B}^{12}=[-1,1]^{12}$}                                                    \\ \cline{2-7} 
                  & \multicolumn{2}{c|}{2500}           & \multicolumn{2}{c|}{10000}          & \multicolumn{2}{c|}{40000}         \\ \cline{2-7} 
                  & $\mathtt{NRMSE}$ & $\mathtt{RL_1}$ & $\mathtt{NRMSE}$ & $\mathtt{RL_1}$ & $\mathtt{NRMSE}$ & $\mathtt{RL_1}$ \\ \hline
DRiLLS ($k^*=1$)              & 12.81$\%$        & 22.42$\%$      & 11.42$\%$        & 20.09$\%$       & 2.18$\%$        & 2.52$\%$       \\ \hline
DRiLLS ($k^*=2$)              & 9.49$\%$         & 16.02$\%$        & 2.91$\%$         & 2.98$\%$        & 1.74$\%$         & 1.67$\%$        \\ \hline
%NLL(10blocks)               & $\%$             & $\%$            & $\%$             & $\%$            & $\%$             & $\%$            \\ \hline
%NLL(20blocks)           & $\%$             & $\%$            & $\%$             & $\%$            & $\%$             & $\%$            \\ \hline
AS ($k^*=1$)                & 12.38$\%$             & 22.36$\%$            & 11.54$\%$             & 19.95$\%$            & 10.02$\%$             & 17.34$\%$            \\ \hline
AS ($k^*=2$)                & 13.65$\%$             & 23.71$\%$            & 12.22$\%$             & 21.14$\%$            & 10.77$\%$             & 18.41$\%$            \\ \hline

 \end{tabular}\end{center}
}
\end{table}

Numerical  approximation errors produced by our DRiLLS method as well as the NLL and  AS methods are reported for the above examples  in \Cref{tab:f_4,tab:f_5,tab:f_6,tab:f_7}, where the sizes of the training dataset are  selected as $N=500$, $2500$ and $10000$ for $\Omega_{A}^d$, and $N=2500$, $10000$ and $40000$ for $\Omega_B^d$. 
Note that the training samples in all these cases are very sparse due to the high dimension of input space. Some observations from these tables are summarized below.

For the functions $f_4, f_5, f_6$ with the domain $\Omega_{A}^{10}$, both DRiLLS  and NLL perform better than  AS, as their approximation errors are several times smaller than in that of AS  with all tested sizes of training samples 500, 2500 and 10000. When 500 samples are used
for training, DRiLLS performs similarly to NLL, but gradually outperforms NLL when the size of the training dataset increases to 2500 and 10000. For the functions $f_4, f_5, f_6$ with the domain $\Omega_{A}^{20}$,  NLL performs better than the AS. When 500 training samples are used, DRiLLS achieves worse results than  NLL. For $f_4$ and $f_5$, it even yields errors bigger than  AS. However, once the size of the training dataset size increases to $2500$ and  $10000$, the performance of  DRiLLS improves significantly: its errors are close to that of NLL for  $f_4$ and $f_5$, and better than  that of NLL for  $f_6$. 

For the functions $f_4, f_5, f_6, f_7$ with the domain $\Omega_{B}^{8}$, DRiLLS  achieves the best performance among all three methods with all tested sizes of training samples $2500$, $10000$ and $40000$. Particularly, NLL does not work at all partially due to  the existence of interior critical points in the domain.  For the functions $f_4, f_5, f_6, f_7$ with the domain $\Omega_{B}^{12}$, NLL still does not work at all as in the  case of $\Omega_B^8$. For the training dataset of the size $2500$, both DRiLLS  and AS cannot achieve good approximations. However, once the sample size increases to $10000$ and $40000$,  DRiLLS yields much a much more accurate function approximation whose errors are several times smaller than that of  AS with one or two active coordinates. 

For the functions $f_4$, $f_5$, $f_6$ with the domains $\Omega_{A}^{10}$ and $\Omega_{A}^{20}$, DRiLLS  with $k^*=2$ achieves approximation errors very close to that with $k^*=1$ case. For $f_4, f_5, f_6, f_7$ with the domains $\Omega_{B}^{8}$ and  $\Omega_{B}^{12}$, DRiLLS  with $k^*=2$ usually achieves slightly smaller errors than that with $k^*=1$, but these values are generally on the same order. This makes sense since $k^*=1$ is the ideal and natural choice when the level sets of the target function are well captured.  On the other hand, AS with two active variables performs almost  the same as that with only one active variable,  partly because the input dimension can not be effectively reduced by linear transformations when the level sets are nonlinear \cite{zhang2019learning}.

Overall, the approximation error by our DRiLLS method quickly  decreases as the size of training dataset increases, and  our method significantly outperforms the NLL and  AS methods when the dataset size becomes relatively large.

\subsection{A PDE-related application}\label{pdeapp}
The PDE-constrained optimization and optimal control problems in engineering applications often lead to multi-query computing scenarios where multiple numerical PDE solves are required as parameters of the problems change, which results in huge or even prohibitive computational cost. On the other hand, the goal of multi-query numerical simulations is often to determine the response of certain quantities of interest (QoI) to the varying parameters and/or the sensitivity of the QoI with respect to its parameters. Therefore, a  function approximation method  can be used to model the QoI offline as a function of system parameters, which can then be  be applied online to provide real-time responses. In the following, we demonstrate the performance of our DRiLLS method through a thermal block engineering problem, which is a popular test case for model order reduction algorithms \cite{haasdonk2017reduced}. 
	
\begin{figure}[!htbp]
\begin{minipage}[b]{0.325\linewidth}
\centering
\includegraphics[width=1\textwidth]{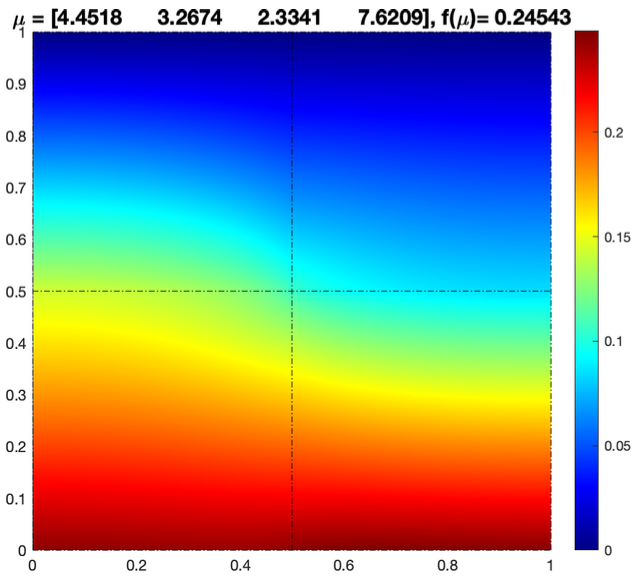}
\end{minipage}
\begin{minipage}[b]{0.325\linewidth}
\centering
\includegraphics[width=1\textwidth]{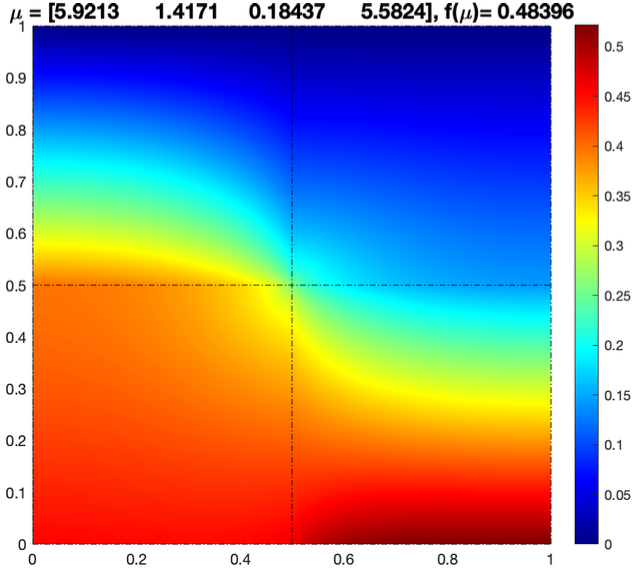}
\end{minipage}
\begin{minipage}[b]{0.325\linewidth}
\centering
\includegraphics[width=1\textwidth]{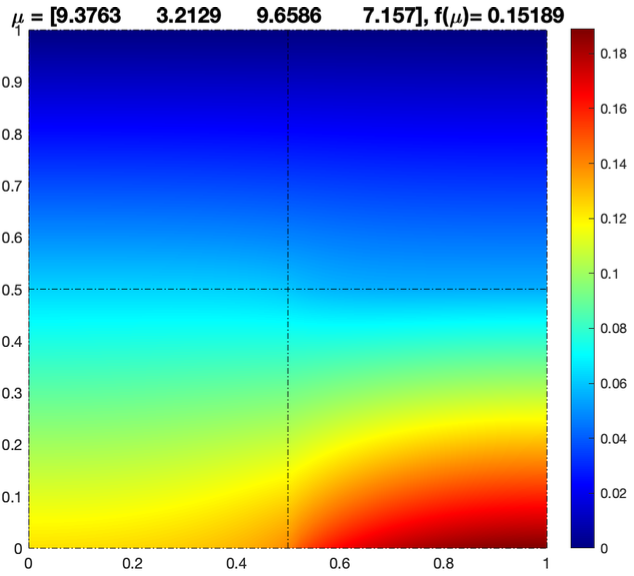}
\end{minipage}
\caption{Sample solutions to the thermal block problem for the  case  $p=4$.}
\label{fig:tb_soln}
\end{figure}

	Consider the heat diffusion on the domain $\Omega=[0,1]^2$ as follows:
	\begin{equation}\label{tb:eq}
	\begin{aligned}
	-\nabla \cdot (\kappa(x,y;  \bm{\mu}) \nabla u(x,y; \bm{\mu})) &= 0,\quad\text{in } \Omega,\\
	                                    u(x,y;  \bm{\mu}) &= 0, \quad \text{on } \Gamma_D, \\
	(\kappa(x,y; \bm{\mu})\nabla u(x,y;  \bm{\mu}))\cdot {\bf n}(x,y)    &= i,\,\quad\text{on } \Gamma_{N,i}, \text{ for } i= 0, 1,
	\end{aligned}
	\end{equation}
	where the zero Dirichlet boundary condition is imposed on the upper boundary denoted by $\Gamma_D$, the left and right edges of the domain are insulated, denoted by $\Gamma_{N,0}$, and unit flux in is assumed on the lower boundary denoted by $\Gamma_{N,1}$. Suppose that the domain is uniformly divided into $p$ sub-blocks $\{\Omega_i\}_{i=1}^p$ and a piecewise constant diffusion coefficient is assumed in each sub-block whose magnitude could vary in a prescribed interval. That is, $$\kappa(x,y;  \bm{\mu}) \coloneqq \sum_{i=1}^{p} \mu_i \chi_{\Omega_i}(x,y)$$ with $ \bm{\mu}= (\mu_1,\mu_2,\cdots,u_p) \in \mathcal{P} \coloneqq [0.1, 10]^p$. The QoI is the average temperature at the lower boundary 
	\begin{equation}\label{tb:f}
	f( \bm{\mu}) \coloneqq \int_{\Gamma_{N,1}} u(x,y;  \bm{\mu})\,ds,
	\end{equation}
which can be regarded as a function of $p$ input variables. 
To learn this function, $N$ samples are randomly selected in the parameter space $\mathcal{P}$, and the finite element method is taken to solve \Cref{tb:eq} and evaluate $f( \bm{\mu})$ \cite{brenner2008mathematical}. The derivative information, $\nabla_{ \bm{\mu}} f$,  is then calculated using an adjoint approach. As an example, several sample solutions of the thermal block problem are plotted in \Cref{fig:tb_soln} for the  case  $p=4$. 

\begin{figure}[!htbp]
\hspace{-0.3cm}
\centering
    \label{fig:rs}
\subfigure[4D]{
\centering
\includegraphics[width=1.63in]{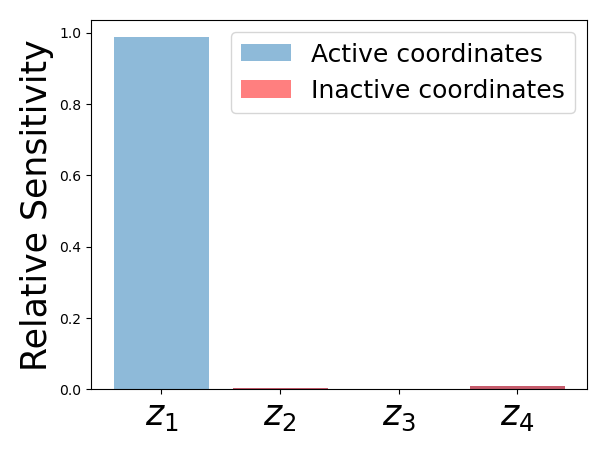}
%\caption{fig1}
}%
\subfigure[9D]{
\centering
\includegraphics[width=1.63in]{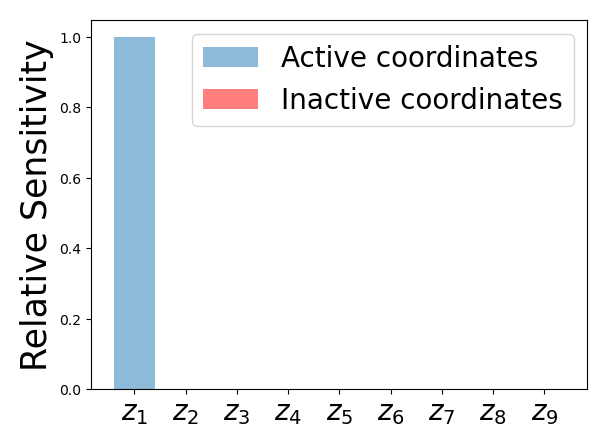}
%\caption{fig1}
}%
\subfigure[16D]{
\centering
\includegraphics[width=1.63in]{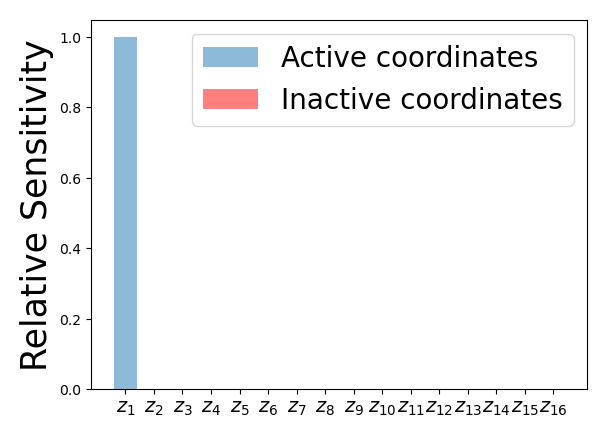}
%\caption{fig1}
}%
 \vspace{-0.2cm}
\caption{Visual illustration of the relative sensitivities of the QoI \eqref{tb:f} to the transformed variables $\bm{z}$ produced by our DRiLLS method with $k^*=1$.}
\end{figure}

\begin{figure}[!htbp]
\hspace{-0.3cm}
\centering
    \label{fig:tbs_low}
\subfigure[4D]{
\centering
\includegraphics[width=1.63in]{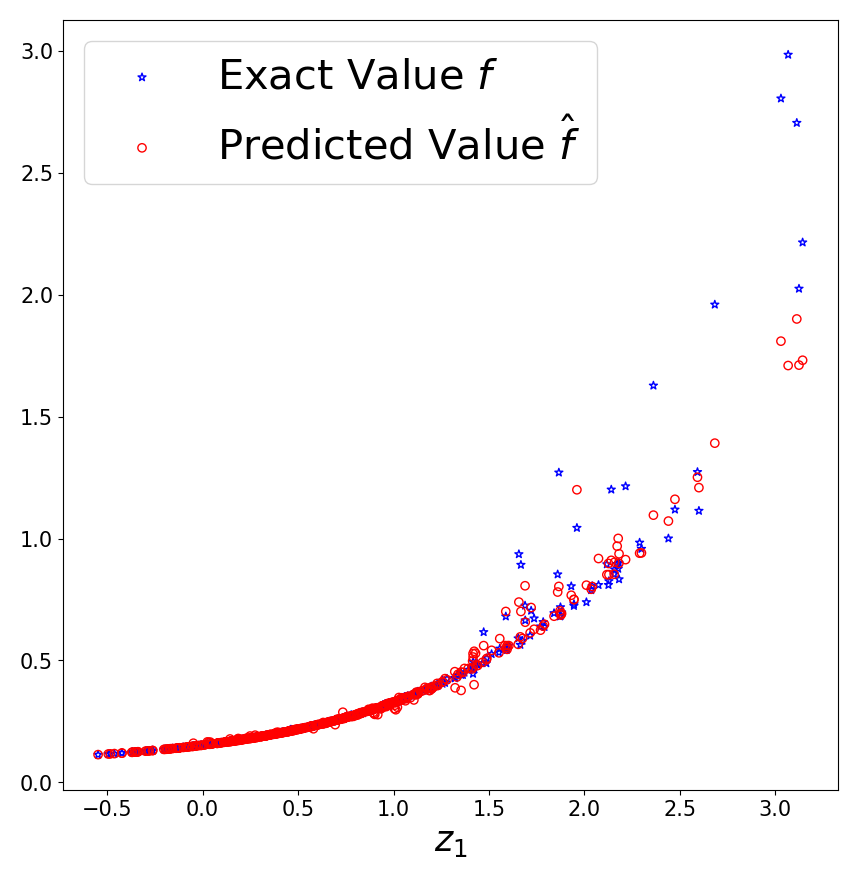}
%\caption{fig1}
}%
\subfigure[9D]{
\centering
\includegraphics[width=1.63in]{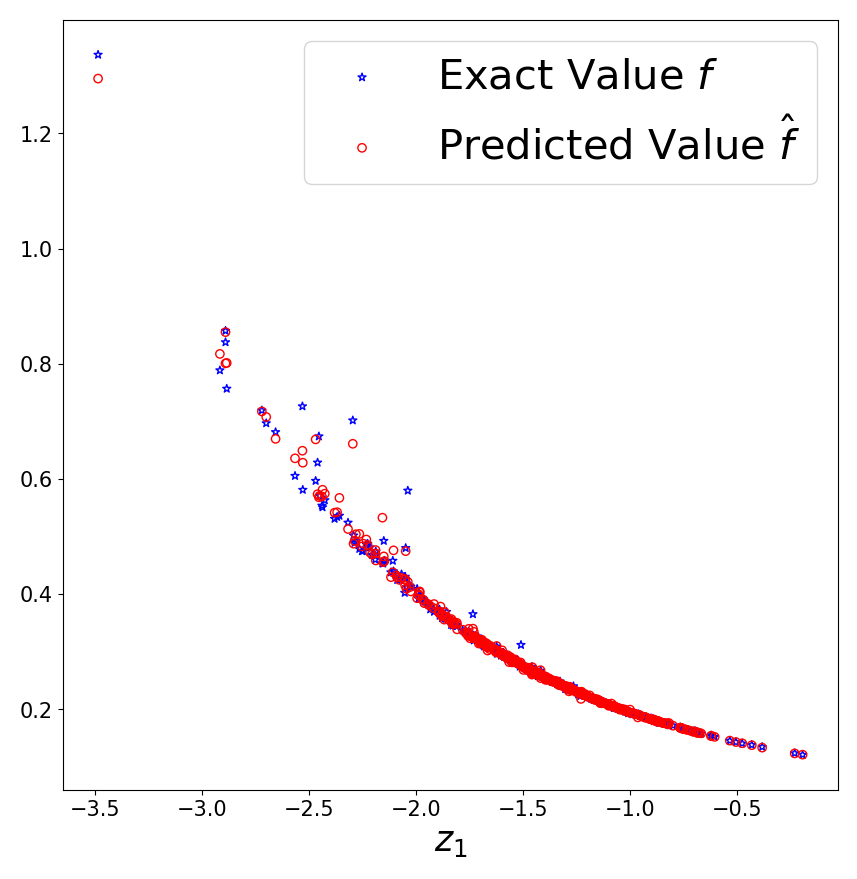}
%\caption{fig1}
}%
\subfigure[16D]{
\centering
\includegraphics[width=1.63in]{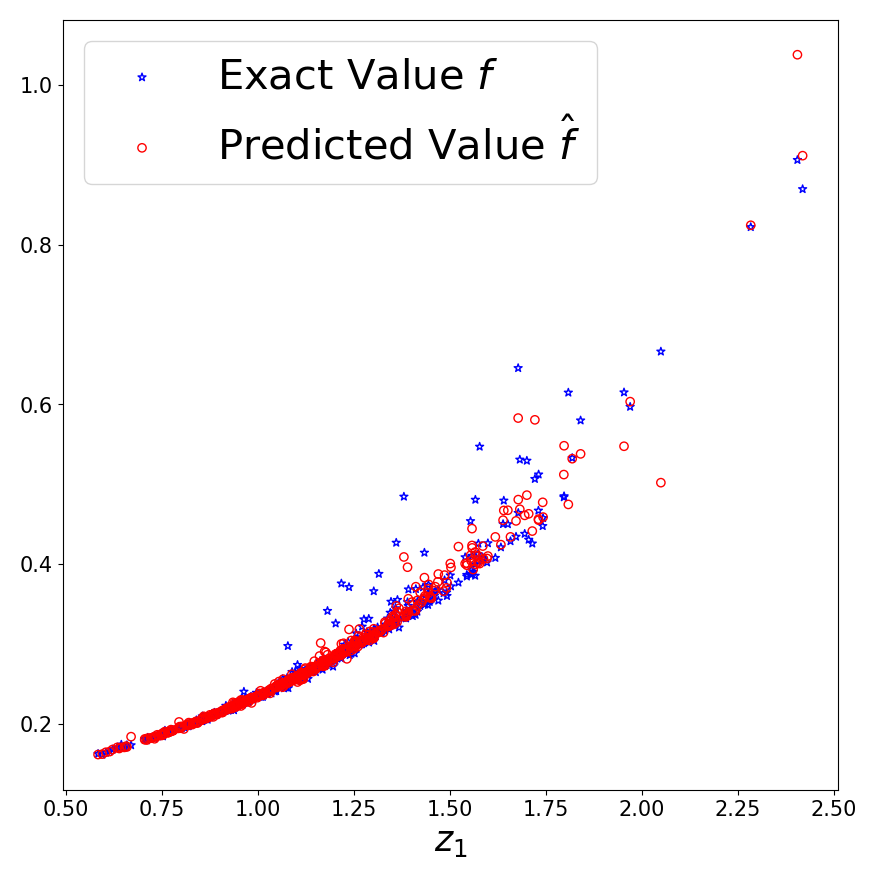}
%\caption{fig1}
}%
 \vspace{-0.2cm}
\caption{The QoI \eqref{tb:f} approximated by our DRiLLS method with $k^*=1$ for the thermal block problem.}
\end{figure}

Assume the dimension of the parametric space to be $p=4$, $9$ and $16$ respectively.  The DRiLLS method is used to learn and predict the QoI \eqref{tb:f}.
For the 4D ($p=4$) case,  $\bm{z}=\bm{g}({\bm \mu})$ and $\hat{\bm \mu}=\bm{h}({\bm z})$ in the  PRNN  are modeled by two FCNNs with 2 hidden layers and  the size of training dataset is $N=500$, but for the 9D ($p=9$) and 16D ($p=16$) cases, $\bm{g}$ and $\bm{h}$ are modeled by two FCNN with 4 hidden layers and the size of training dataset is $N=10000$.These experiments use the ideal choice $\bm{\omega}=(0,1,\cdots,1)$, because the results in \Cref{sec:mor} show that our DRiLLS method can performs well even with $k^*=1$.
Moreover, $\lambda_1=1$, $\lambda_2=100$, $\alpha=50$ are set in the loss function. 
We observe that this example benefits from some L-BFGS optimization steps in addition to the 60000 Adam optimization steps used by default.
To evaluate the sensitivity of the QoI with respect to transformed variables $\bm{z}$, we calculate the following relative sensitivity with respect to $z_i$: for $i=1, \ldots, p$, 
\begin{equation}
\mathtt{RS_i}\coloneqq \frac{|\frac{1}{N}\sum_{n=1}^N \nabla f(\bm \mu_{test}^{(n)}) \cdot\frac{\partial \hat{\bm \mu}_{test}^{(n)}}{\partial z_i}|}{\sum_{m=1}^{p} |\frac{1}{N}\sum_{n=1}^N \nabla f(\bm \mu_{test}^{(n)}) \cdot\frac{\partial \hat{\bm \mu}_{test}^{(n)}}{\partial z_m}|},
\end{equation}
where $\hat{\bm \mu}_{test}^{(n)} = {\bm h} \circ {\bm g} (\bm \mu_{test}^{(n)})$.
The numerical results on relative sensitivities of the QoI produced by our DRiLLS method are reported in \Cref{tab:rs} and also visually illustrated in \Cref{fig:rs}. In all cases, it is seen that the target function is sensitive only to the active variable $z_1$, as desired.
The regression plots  are presented in \Cref{fig:tbs_low}, which again show good agreement between the exact values and the predicted values.
For the purpose of comparison, the approximation errors produced by our DRiLLS method
 are presented  in \Cref{tb:tbs} together with those by the NLL  and  AS methods (with $k^*=1$ and $2$). It is easy to see that our DRiLLS method significantly outperforms the other two on this problem.   

\begin{table}[!htbp]
{\footnotesize
  \caption{Results on the relative sensitivities of the QoI \eqref{tb:f} to the transformed variables $\bm{z}$ produced by our DRiLLS method with $k^*=1$.}
  \label{tab:rs}
\begin{center}
\renewcommand{\arraystretch}{1.2}
\begin{tabular}{|c|c|c|c|c|c|c|c|c|}
\hline
       & $z_1$    & $z_2$    & $z_3$    & $z_4$    & $z_5$    & $z_6$    & $z_7$    & $z_8$     \\ \hline
4D  & 9.87e-01 & 4.06e-3  & 1.21e-3  & 7.88e-3  &          &          &          &           \\ \hline
9D  & 9.99e-01 & 1.52e-04 & 4.86e-05 & 6.86e-05 & 1.26e-04 & 1.08e-04 & 6.80e-05 & 9.39e-05  \\ \hline
16D & 9.99e-01 & 6.18e-05 & 7.29e-05 & 6.86e-05 & 4.69e-05 & 3.63e-05 & 6.72e-05 & 4.69e-05  \\ \hline\hline
       & $z_9$      & $z_{10}$   & $z_{11}$   & $z_{12}$   & $z_{13}$   & $z_{14}$   & $z_{15}$   & $z_{16}$   \\ \hline
4D  &            &            &            &            &            &            &            &            \\ \hline
9D  & 6.18e-05 &            &            &            &            &            &            &            \\ \hline
16D & 3.32e-05 & 1.09e-04 & 8.29e-05 & 1.92e-04 & 9.05e-05 & 5.91e-05 & 3.06e-05 & 6.10e-05 \\ \hline
\end{tabular}
\end{center}
}
\end{table}

\begin{table}[!htbp]
{\footnotesize
  \caption{Numerical approximation errors produced by DRiLLS, NLL and AS  for the QoI \eqref{tb:f} in the thermal block problem when the number of input parameters is $4$, $9$, or $16$.}
    \label{tb:tbs}
    \renewcommand{\arraystretch}{1.2}
\begin{center}
\begin{tabular}{|c|c|c|c|c|c|c|}\hline
& \multicolumn{2}{c|}{4D} & \multicolumn{2}{c|}{9D} & \multicolumn{2}{c|}{16D}\\ \cline{2-7} 
& $\mathtt{NRMSE}$    & $\mathtt{RL_1}$   & $\mathtt{NRMSE}$    & $\mathtt{RL_1}$   & $\mathtt{NRMSE}$    & $\mathtt{RL_1}$     \\ \hline
DRiLLS ($k^*=1$) &          1.87$\%$&       4.38$\%$&           1.33$\%$&       1.94$\%$&         2.19$\%$&        2.72$\%$ \\ \hline
NLL ($k^*=1$) &          4.45$\%$&       19.03$\%$&          5.10$\%$&       13.15$\%$&         6.26$\%$&        12.71$\%$ \\ \hline
NLL ($k^*=2$) &          4.00$\%$&       14.25$\%$&          4.70$\%$&       12.64$\%$&         6.66$\%$&        13.45$\%$ \\ \hline
AS ($k^*=1$)  &          4.69$\%$&       20.48$\%$&         5.20$\%$&       13.94$\%$&          6.13$\%$&        12.69$\%$ \\ \hline
AS ($k^*=2$)                & 3.79$\%$             & 16.37$\%$            & 4.46$\%$             & 12.71$\%$            & 5.95$\%$             & 12.83$\%$            \\ \hline
 \end{tabular}
\end{center}
}
\end{table}

\section{Conclusions}\label{sec:conclusion}

In this paper,  a  neural network-based method ``DRiLLS''  has been proposed for dimension reduction  via learning level sets in function approximation, which performs very well on application involving high-dimensional limited/sparse data.  This model consists of a PRNN module and a synthesized regression module: the former aims to find a handful of active variables that reduce the dimension of original input space and  to capture the level set information of the target function, while the latter is designed for effectively approximating function values based on these active variables and neighborhood relationships in the input space. Particularly, the PRNN employs two feed-forward fully connected neural networks to model the nonlinear transformations $\bm{z} = \bm{g}(\bm{x}; \Theta_g)$ and $\hat{\bm x} = \bm h(\bm z; \Theta_h)$, which transform the original data $\mathbf{x}$ but do not change its dimension. A new total loss function has been introduced which involves a pseudo-reversibility term enforcing $\bm{h}\circ\bm{g}$ to be close to the identity mapping, an active direction fitting loss compelling changes in  target function value caused by small perturbations of the inactive variables to be tangent to the function's level sets, and a bounded derivative loss regularizing the graph of the target function as the input variables change from $\bm{x}$ to $\bm{z}_A$. Once the PRNN is trained, function values are predicted using a synthesized regression that selects neighboring training samples based on the distance information from the original input space and  projects them to the space of the active variables to perform local least-squares polynomial fitting.

Some ablation studies are carried out to show the effect of the major components and hyper-parameters of our DRiLLS method. 
Several high-dimensional function approximation examples and a PDE-related application problem are also used to investigate and compare the performance of the proposed method with the popular nonlinear and linear dimension reduction methods (NLL and AS), and experimental results demonstrate that the DRiLLS method is superior in many examples of high-dimensional function approximation with limited/sparse data. Note that our method requires gradient information of the target function just as the NLL and  AS methods. As a next step, we will explore gradient-free approaches to level set learning and function approximation. 
We will further combine our method with model order reduction methods to develop efficient intrusive computational surrogates for complex systems with high-dimensional parameters.

\section*{Acknowledgments}
This work is partially supported by U.S. Department of Energy, Office of Science, Office of Advanced Scientific Computing Research, Office of Biological and Environmental Research through Applied Mathematics, Earth and Environmental System Modeling, and Scientific
Discovery through Advanced Computing programs under university grants DE-SC0022254 (L. Ju)  and DE-SC0020270 (L. Ju and Z. Wang) and contract ERKJ387 (G. Zhang) at Oak Ridge National Laboratory, and  by U.S.  National Science Foundation under grant DMS-2012469 (Z. Wang). 
 
%Z. Wang's work is partially supported by U.S. Department of Energy Office of Science under grant DE-SC0020270 and National Science Foundation under grant DMS-2012469.  L. Ju's work is partially supported by U.S. Department of Energy Office of Science under grants DE-SC0020270 and DE-SC0022254. G.~Zhang's work is supported by U.S. Department of Energy, Office of Science, Office of Advanced Scientific Computing Research, Applied Mathematics program under contract ERKJ387, at Oak Ridge National Laboratory (ORNL). ORNL is operated by UT-Battelle, LLC., for  U.S. Department of Energy under Contract DE-AC05-00OR22725.  

\bibliographystyle{siamplain}
\bibliography{references}
\end{document}